%% file: main.tex
\documentclass{article}
\usepackage[margin=1in]{geometry}

\usepackage{float}
\usepackage{graphicx}
\usepackage{subfig}   
\usepackage{caption}
\usepackage{xcolor}
\usepackage[super,sort&compress]{natbib}  
\usepackage{amsmath,amssymb}
\usepackage{tabularx}
\usepackage{booktabs}
\usepackage{siunitx}
\usepackage{longtable}
\usepackage{authblk}
\usepackage{ltbf_macros}
\usepackage{doi}  
\usepackage{url}  
\usepackage{hyperref}  

\usepackage{longtable}
\usepackage{array}
\usepackage{booktabs}
\usepackage{makecell}
\usepackage{adjustbox}

\usepackage{ragged2e}
\usepackage{fancyvrb}

\DefineVerbatimEnvironment{Prompt}{Verbatim}{
    frame=single,
    rulecolor=\color{gray},
    fontsize=\small,
    xleftmargin=5mm,
    xrightmargin=5mm,
    commandchars=\\\{\}
}

\renewcommand{\arraystretch}{1.2}
\newcolumntype{L}[1]{>{\raggedright\arraybackslash}p{#1}}
\newcolumntype{C}[1]{>{\centering\arraybackslash}p{#1}}
\newcommand{\na}{\multicolumn{1}{c}{--}}

\newcommand{\iterSummary}[3]{%
    \vspace{0.2cm}
    \noindent\fcolorbox{black}{gray!10}{%
        \parbox{\dimexpr\textwidth-2\fboxsep-2\fboxrule}{%
            \small
            \textbf{Decision:} #1 \\
            \textbf{Updated Instructions:} #2 \\
            \textbf{Rationale:} #3
        }%
    }%
    \vspace{0.5cm}
}

\begin{document}

\renewcommand\Affilfont{\footnotesize}
\renewcommand\Authfont{\normalsize}

\title{Scaling Clinician-Grade Feature Generation from Clinical Notes with Multi-Agent Language Models}

\author[1,*]{Jiayi (Joyee) Wang}
\author[1]{Jacqueline Jil Vallon}
\author[2]{Nikhil V. Kotha}
\author[2]{Neil Panjwani}
\author[2]{Xi Ling}
\author[9]{Margaret Redfield}
\author[3]{Sushmita Vij}
\author[4]{Sandy Srinivas}
\author[5,6,7]{John Leppert}
\author[2,$\dagger$]{Mark K. Buyyounouski}
\author[2,8,9,$\dagger$]{Mohsen Bayati}

\affil[1]{Department of Management Science and Engineering, Stanford University School of Engineering, Stanford, CA}
\affil[2]{Department of Radiation Oncology, Stanford University School of Medicine, Stanford, CA}
\affil[3]{Graduate Business School Research Hub, Stanford University Graduate Business School, Stanford, CA}
\affil[4]{Department of Medicine (Oncology), Stanford University School of Medicine, Stanford, CA}
\affil[5]{Department of Medicine, Stanford University School of Medicine, Stanford, CA}
\affil[6]{Department of Urology, Stanford University School of Medicine, Stanford, CA}
\affil[7]{Veterans Affairs Palo Alto Health Care System, Palo Alto, CA}
\affil[8]{Department of Electrical Engineering, Stanford University School of Engineering, Stanford, CA}
\affil[9]{Operations, Information and Technology, Stanford University Graduate Business School, Stanford, CA}

\renewcommand\Authsep{, }
\renewcommand\Authand{, }
\renewcommand\Authands{, }

\date{}
\maketitle

\vspace{-0.5cm}
\noindent $^{*}$Corresponding author. Email: jyw@stanford.edu\\
$^{\dagger}$Served as equally contributing co-senior authors

\begin{abstract}
Developing accurate clinical prediction models is often bottlenecked by the difficulty of generating meaningful predictive features from unstructured data. While electronic health records (EHRs) contain rich narrative information, extracting a comprehensive list of structured features from them requires extensive domain knowledge and granular clinical judgment, a process that is historically manual, unscalable, and impractical for large cohorts. In this study, we first established a rigorous patient-level Clinician Feature Generation (CFG) protocol, in which domain experts manually reviewed notes to define and extract nuanced features for a cohort of 147 patients with prostate cancer. 
As a high-fidelity ground truth, this labor-intensive process provided the blueprint for SNOW (Scalable Note-to-Outcome Workflow), a transparent multi-agent large language model (LLM) system designed to autonomously mimic the iterative reasoning and validation workflow of clinical experts. In the prostate cancer cohort, SNOW achieved a predictive performance for 5-year recurrence (AUC-ROC $0.767 \pm 0.041$) that was indistinguishable from the gold-standard manual CFG ($0.762 \pm 0.026$) and superior to structured baselines, clinician-guided LLM extraction, and six representational feature generation (RFG) approaches. Manual CFG required prolonged expert review and per‑patient abstraction; in contrast, once configured, SNOW generated the full patient‑level feature table in 12 hours with 5 hours of clinician oversight, reducing human expert effort by approximately 48-fold.
To assess scalability in a setting where manual CFG is infeasible, we deployed SNOW on an external population of 2,084 patients with heart failure with preserved ejection fraction (HFpEF) from the MIMIC-IV database.
Without task-specific tuning, SNOW generated prognostic features that outperformed baseline and RFG methods for 30-day (SNOW: $0.851 \pm 0.008$) and 1-year (SNOW: $0.763 \pm 0.003$) mortality prediction. These results demonstrate that a modular LLM agent-based system can scale expert-level feature generation from clinical notes, while enabling interpretable use of unstructured EHR text in outcome prediction and preserving generalizability across a variety of settings and conditions.
\end{abstract}


\begin{figure}[htbp]
\centering
\subfloat[]{
    \includegraphics[width=.9\linewidth]{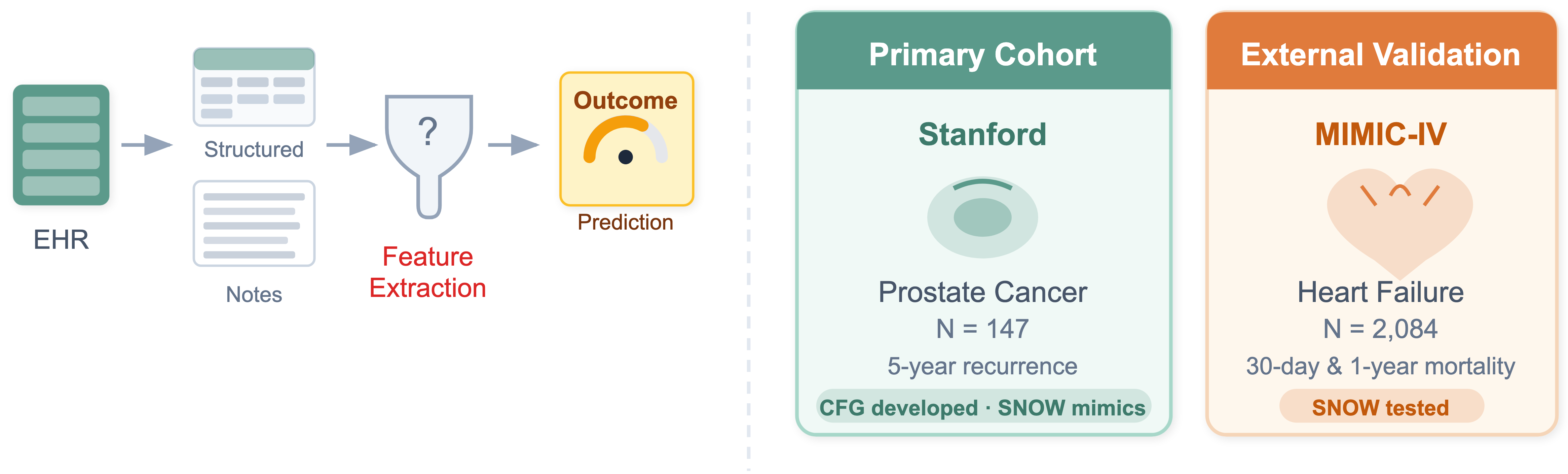}
    \label{fig:1a}
}

\subfloat[]{
    \includegraphics[width=0.95\linewidth]{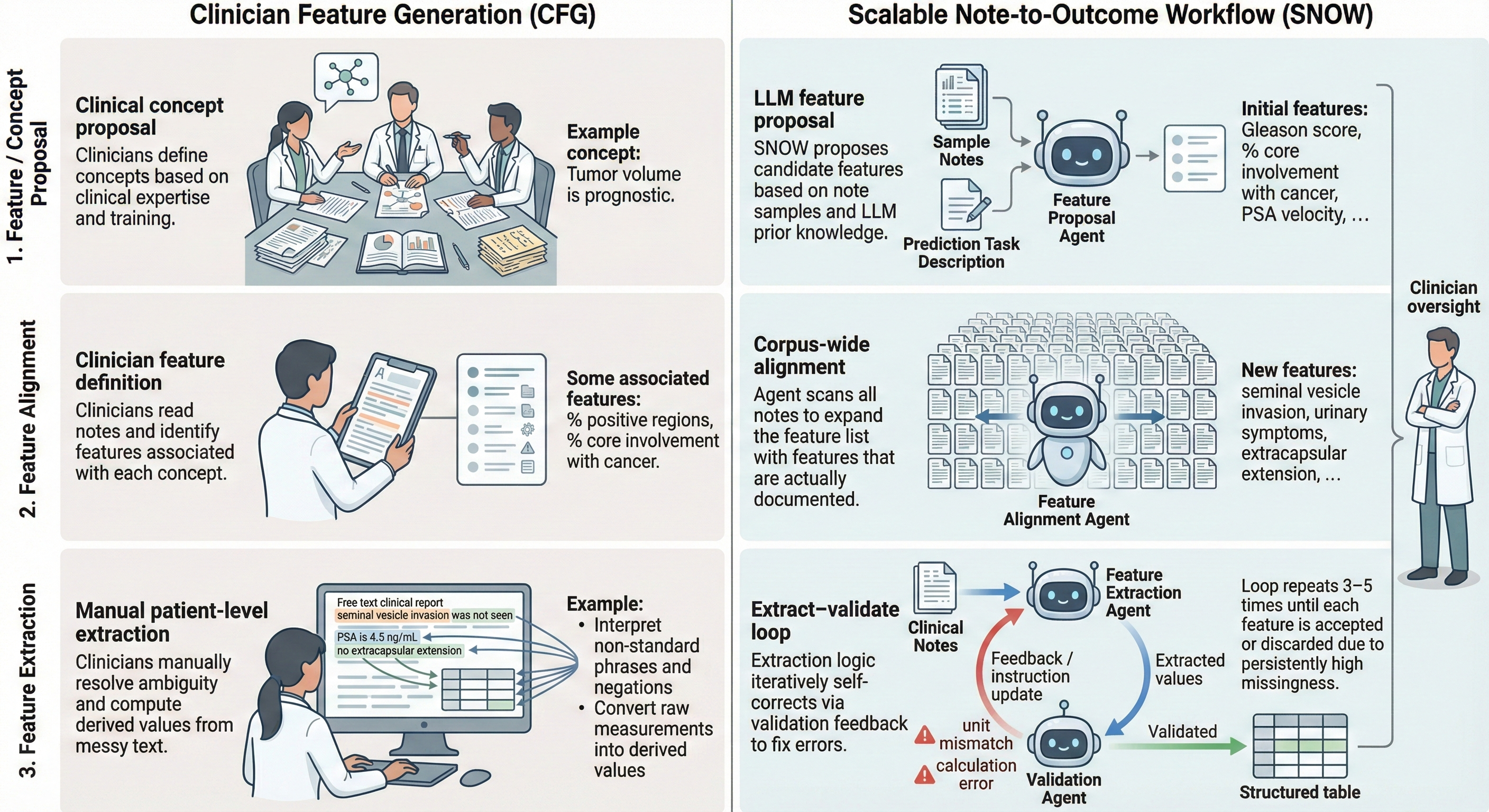}
    \label{fig:1b}
}
\caption{\textbf{Systematic framework for scalable, expert-level clinical feature generation.} 
\textbf{(a)}, Overview of the clinical prediction pipeline highlighting the bottleneck of unstructured text (left),
Summary of the dual-cohort evaluation strategy: the primary cohort (N=147) serves as a high-fidelity `calibration testbed' to benchmark agentic reasoning against a high-resolution human ground truth, while the external cohort (N=2,084) serves as a `deployment testbed' to validate scalability in a real-world setting where manual ground truth generation is intractable (right).
\textbf{(b)}, Comparison of the manual Clinician Feature Generation (CFG) protocol versus SNOW system. 
}
\label{fig:main}
\end{figure}

\input{patient_level_main.tex}

\newpage
\bibliographystyle{naturemag}
\bibliography{mybib}

\newpage
\appendix
\counterwithin{figure}{section}
\counterwithin{table}{section}
\renewcommand{\thefigure}{\thesection.\arabic{figure}}
\renewcommand{\thetable}{\thesection.\arabic{table}}
\setcounter{figure}{0}
\setcounter{table}{0}

\begin{center}
{\Large \textbf{Supplementary Information:}}\\[0.3cm]
{\large Scaling Clinician-Grade Feature Generation from Clinical Notes with Multi-Agent Language Models}\\[0.5cm]

{\normalsize
Jiayi (Joyee) Wang$^{1,*}$,
Jacqueline Jil Vallon$^{1}$,
Nikhil V. Kotha$^{2}$,
Neil Panjwani$^{2}$,
Xi Ling$^{2}$,
Margaret Redfield$^{9}$,
Sushmita Vij$^{3}$,
Sandy Srinivas$^{4}$,
John Leppert$^{5,6,7}$,
Mark K. Buyyounouski$^{2,\dagger}$,
Mohsen Bayati$^{2,8,9,\dagger}$
}\\[0.3cm]

{\footnotesize
$^{1}$Department of Management Science and Engineering, Stanford University School of Engineering\\
$^{2}$Department of Radiation Oncology, Stanford University School of Medicine\\
$^{3}$Graduate Business School Research Hub, Stanford University Graduate Business School\\
$^{4}$Department of Medicine (Oncology), Stanford University School of Medicine\\
$^{5}$Department of Medicine, Stanford University School of Medicine\\
$^{6}$Department of Urology, Stanford University School of Medicine\\
$^{7}$Veterans Affairs Palo Alto Health Care System\\
$^{8}$Department of Electrical Engineering, Stanford University School of Engineering\\
$^{9}$Operations, Information and Technology, Stanford University Graduate Business School\\[0.2cm]
$^{*}$Corresponding author. $^{\dagger}$Co-senior authors
}
\end{center}

\vspace{0.5cm}

\newpage

\section{CFG Concept Abstraction}
\label{appendix:cfg-features}
\input{appendix/cfg_table}

\section{SNOW Feature Table}
\label{appendix:afg-features}
\input{appendix/feature_table}

\section{HFpEF Baseline Features}
\label{appendix:hfpef-features}
\input{appendix/hfpef_summary_stats}

\section{Performance Comparison of RFG Methods}
\label{appendix:rfg_comparison}
\input{appendix/rfg_comparison}

\section{Feature Importance Analysis}
\label{appendix:feature_analysis}
\input{appendix/feature_analysis}

\section{Sensitivity Analysis Including Post-Prostatectomy Radiotherapy Patients}
\label{appendix:168_analysis}
\input{appendix/168_patients}

\section{
Sensitivity Analysis on Alternative Definition of Post-Prostatectomy Biochemical Failure
}
\label{appendix/alternative-BF}
\input{appendix/alternative_BF}

\section{Sensitivity Analysis on Excluding Sociodemographic Features (Race, Ethnicity, Language)}
\label{appendix:sociodemographic_analysis}
\input{appendix/sociodemographic_analysis}

\section{Additional Performance Metrics (AUPR and F1 Score)}
\label{appendix:aupr-f1}
\input{appendix/aupr_f1}

\section{Sensitivity Analysis: Imputation Methods}
\label{appendix:imputation-analysis}
\input{appendix/imputation_analysis}

\section{LLM Specifications and Prompts}
\label{appendix:llm-specs}
\input{appendix/LLM_specs}

\section{Iterative Extract-Validate Loop Examples}
\label{appendix:SNOW-examples}
\input{appendix/SNOW_examples}

\section{Supplementary Tables}
\input{appendix/hfpef_missing_rate}
\end{document}

%% file: patient_level_main.tex
\section{Introduction}
\label{sec:patient-level-paper}
\input{new_intro}

\section{Methods}
\label{sec:patient-level-methods}

This study was reviewed and approved by the Institutional Review Board (IRB) of Stanford University (IRB-49456), and informed consent was obtained from all participants. All Large Language Model (LLM) operations in this study were performed via the Secure API provided by Stanford Health Care and Stanford School of Medicine. This API is approved for use with sensitive data, including Protected Health Information (PHI) and Personally Identifiable Information (PII), and adheres to institutional data security and privacy standards. Using this secure API ensured that all LLM-based feature extraction was conducted in compliance with HIPAA regulations and appropriate ethical guidelines for handling clinical data.

We organize our methodology by first describing the experimental framework within the context of our primary dataset: a cohort of patients with prostate cancer treated at Stanford. In the subsections that follow, we define the primary patient cohort and prediction outcome, and subsequently detail the development and implementation of different feature generation approaches: the manual Clinician Feature Generation (CFG) protocol, the Scalable Note-to-Outcome Workflow (SNOW), and baseline Representational Feature Generation (RFG) methods. Following this primary analysis, we describe the external validation of our agentic framework on a different population of patients with heart failure using the MIMIC-IV database, for which access was granted under a standard Data Use Agreement (DUA) to assess system generalizability across independent institutions and clinical domains.

\subsection{Primary Cohort: Prostate Cancer}
\label{sec:patient-level-methods-patient-cohort}

We established a primary retrospective cohort of patients treated for prostate cancer at the Stanford Cancer Institute between 2005 and 2015. This inclusion window was specifically selected to ensure a minimum of five years of post-treatment follow-up for all individuals, a necessary duration to reliably obtain the primary endpoint of biochemical failure. This dataset serves as a fixed, high-fidelity benchmark, having undergone the extensive manual curation process described below to establish ground-truth clinical features.

Data extraction was performed using the institutional Electronic Health Record (EHR). The dataset integrates multimodal patient information, comprising structured clinical fields (e.g., demographics, laboratory values, diagnosis codes) and unstructured narrative text. The latter includes physician progress notes, operative reports, and pathology biopsy reports, which serve as the primary inputs for our feature generation pipelines.

Inclusion criteria for the patient cohort ($n=147$) are: individuals who have a prostate-specific antigen (PSA) lab value available $>$ 5 years after treatment; have a pre-treatment biopsy report with the 12 systematic cores recorded ($[$right/left$]$ $[$apex/mid/base$]$ $[$lateral/medial$]$); and received a prostatectomy or radiation therapy as their first treatment with no subsequent treatment within five years. The first criterion ensures we do not misclassify patients as recurrence-free due to loss to follow-up, as the outcome is computed strictly from longitudinal PSA values. The second criterion constrains the study to patients who have sufficient information in a pre-treatment biopsy report to capture a representative description of their disease state. Finally, the third criterion excludes an additional 21 patients who received both prostatectomy and radiation therapy within a five-year span. To address potential selection bias, specifically the concern that excluding these patients might omit instances of biochemical failure treated via salvage radiation, we performed a detailed sensitivity analysis including these 21 patients (see Appendix \ref{appendix:168_analysis}). Table \ref{tab:patient_cohort-chapter2} shows key summary statistics for our final patient cohort.

\begin{table}[ht!]
\centering
\caption{Summary statistics of patient cohort.}
\label{tab:patient_cohort-chapter2}
\begin{tabular}{l c c}
    \toprule
    \textbf{Continuous Features} & \textbf{Mean (SD)} & \textbf{Median} \\
    \midrule
    Age at Treatment & 68.5 (8.4) & 68.7\\
    Charlson Comorbidity Index & 5.2 (2.0) & 5.0\\
    Maximum Pre-treatment PSA (ng/mL) & 10.8 (21.5) & 6.2\\
    Percent Positive Regions & 40.8 (26.5) & 33.3\\
    Radiation Dose (cGy)$^{*}$ & 7688.0 (620.4) & 7800.0\\
    PSA Nadir (ng/mL) & 0.12 (0.18) & 0.05\\
    PSA at Failure (ng/mL)$^{\dagger}$ & 3.81 (3.19) & 2.97\\
    \midrule
    \textbf{Binary/Categorical Features} & \multicolumn{2}{c}{\textbf{Number (\%)}} \\
    \midrule
    Patients with biological failure & \multicolumn{2}{c}{11 (7.5)} \\
    Treatment & & \\
    \hspace{1em}Radiation only & \multicolumn{2}{c}{85 (57.8)} \\
    \hspace{1em}Prostatectomy only & \multicolumn{2}{c}{62 (42.2)} \\
    Radiation Technique$^{*}$ & & \\
    \hspace{1em}Non-HDR & \multicolumn{2}{c}{85 (57.8)} \\
    Race = White & \multicolumn{2}{c}{110 (74.8)} \\
    Staging & & \\
    \hspace{1em}t1 & \multicolumn{2}{c}{75 (51.0)} \\
    \hspace{1em}t2 & \multicolumn{2}{c}{47 (32.0)} \\
    \hspace{1em}t3 & \multicolumn{2}{c}{12 (8.2)} \\
    Grade Group of Max Gleason Score & & \\
    \hspace{1em}1 & \multicolumn{2}{c}{37 (25.2)} \\
    \hspace{1em}2 & \multicolumn{2}{c}{36 (24.5)} \\
    \hspace{1em}3 & \multicolumn{2}{c}{31 (21.1)} \\
    \hspace{1em}4 & \multicolumn{2}{c}{17 (11.6)} \\
    \hspace{1em}5 & \multicolumn{2}{c}{26 (17.7)} \\
    \midrule
    \textbf{Total number of patients} & \multicolumn{2}{c}{\textbf{147}} \\
    \bottomrule
\end{tabular}
\\[4pt]
\footnotesize{$^{*}$Stats among patients receiving radiation (n=85); all non-HDR. \\$^{\dagger}$PSA at failure available for patients with biological failure (n=11).}
\end{table}

\subsection{Outcome}
\label{sec:patient-level-methods-outcome}
All models in prostate cancer analyses predict the probability of biological failure (BF) within five years after the end of treatment signaling cancer recurrence. Biochemical failure is the gold-standard for defining recurrent prostate cancer and an important endpoint because it prompts an evaluation for local recurrence and metastatic disease, and in some cases additional therapy. For patients who receive radiation therapy, a patient is classified as BF if they have a PSA level post-treatment that is at least 2 ng/mL greater than their post-treatment nadir~\citep{roach_defining_2006}. For patients who undergo a prostatectomy, they are classified as BF if they have a PSA level of 0.4 ng/mL or above with the subsequent PSA level increasing after the end of treatment~\citep{stephenson_defining_2006}. Given there is variability in the latter definition, in Appendix~\ref{appendix/alternative-BF}, we complete a sensitivity analysis on the results in which we classify a patient who got a prostatectomy as BF if they have a PSA level post-prostatectomy of 0.2 ng/mL or above with the subsequent PSA level also being 0.2 ng/mL or above~\citep{american_urological_association_psa_2013}.

\subsection{Baseline Features}
\label{sec:patient-level-methods-baseline-features}
Baseline features include features from structured data sources that require minimal data analysis to compute with the output from the EHR. These include demographic and socioeconomic features, the maximum pre-treatment PSA level, and the Charlson Comorbidity Index~\citep{glasheen_charlson_2019}. We include the features of race, ethnicity, and language in our study, but perform a sensitivity analysis (see Appendix~\ref{appendix:sociodemographic_analysis}) on the results excluding these three features to ensure that the outcome prediction machine learning models are not using these features in any biased fashion. For the purpose of our study, the baseline features require minimal clinical expertise to identify, either because they are readily available in the EHR (e.g., PSA values) or because there has been extensive research already conducted on how to compute them providing us with a template to follow (e.g., Charlson Comorbidity Index).

\subsection{Patient-Level Clinician Feature Generation (CFG)}
\label{sec:patient-level-methods-cfg}

To establish a high-fidelity ground truth for prostate cancer recurrence prediction, we developed and executed a rigorous \textit{Clinician Feature Generation} (CFG) protocol. Unlike standard data curation, which typically applies pre-determined rules to extract fixed variables, CFG was designed as an adaptive, expert-driven process capable of resolving the extensive ambiguity present in real-world narrative oncology notes.

The development of this protocol involved a year-long preparatory phase involving bi-weekly multidisciplinary meetings between oncologists, pathologists, and data scientists. This phase focused on translating domain knowledge regarding the disease's natural history and treatment trajectories into a structured feature generation pipeline. Through iterative reading of physician progress notes, the team reinforced a well-recognized challenge in real-world EHR NLP: important prognostic information (e.g., tumor volume proxies like percentage core-length involvement) was often documented implicitly or heterogeneously, rendering fixed population-level extraction rules insufficient.

\begin{figure}[htbp]
\centering
\includegraphics[width=0.9\linewidth]{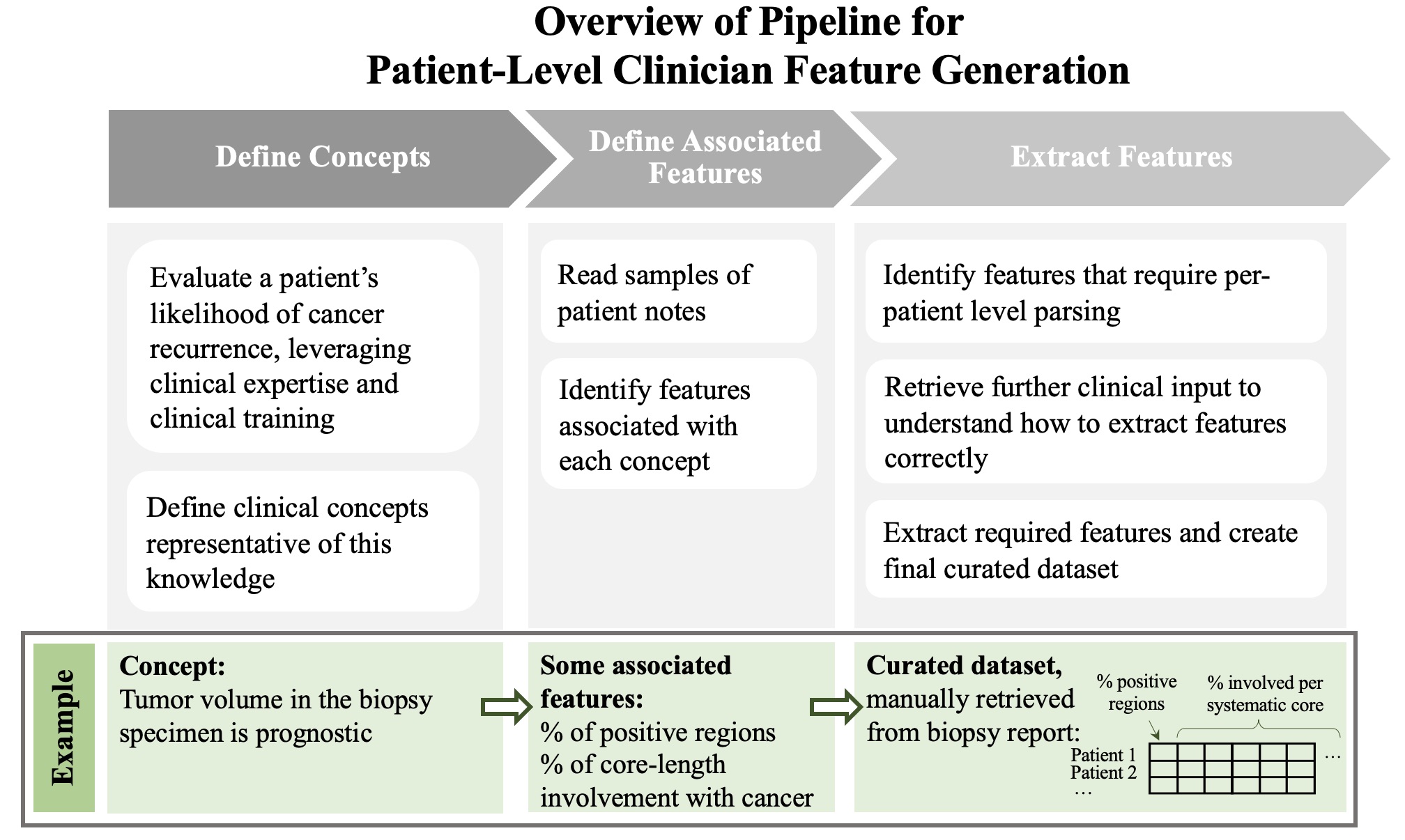}
\caption{\textbf{Systematic workflow for patient-level clinician feature generation (CFG).} 
Stepwise pipeline used by oncologists and data scientists to translate domain expertise into structured features, including defining clinical concepts based on expertise (left), defining associated features by reviewing patient note samples (center), and manually extracting features requiring per-patient parsing and clinical input (right) to create the final curated dataset.}
\label{fig:overview-pipeline}
\end{figure}

To address this, we formalized CFG as a three-stage workflow (Figure~\ref{fig:overview-pipeline}) that mirrors the clinical reasoning process:

\begin{enumerate}
    \item \textbf{Concept Definition (Clinical Translation):} We first translated abstract clinical knowledge into concrete domain-specific concepts required to prognosticate the outcome (e.g., determining that ``tumor volume in the biopsy specimen'' is a critical prognostic concept).
    \item \textbf{Feature Operationalization:} We mapped these concepts to specific, computable features available in the unstructured text. For instance, the concept of tumor volume was operationalized into specific variables such as ``percentage of positive regions'' and ``percentage core-length involvement per systematic core'' (see Appendix \ref{appendix:cfg-features} for the full feature list).
    \item \textbf{Adaptive Patient-Level Extraction:} Crucially, the extraction phase was not a static retrieval task but a dynamic clinical adjudication process. For every patient, clinicians manually reviewed notes to extract basic data points. When standard extraction was impeded by ambiguity, the protocol allowed for real-time collaboration between oncologists and pathologists to revise the interpretation logic for that specific case. This adaptive phase enabled the resolution of complex edge cases that would otherwise be treated as missing or noisy data.
\end{enumerate}

\begin{figure}[htbp]
\centering
\includegraphics[width=\linewidth]{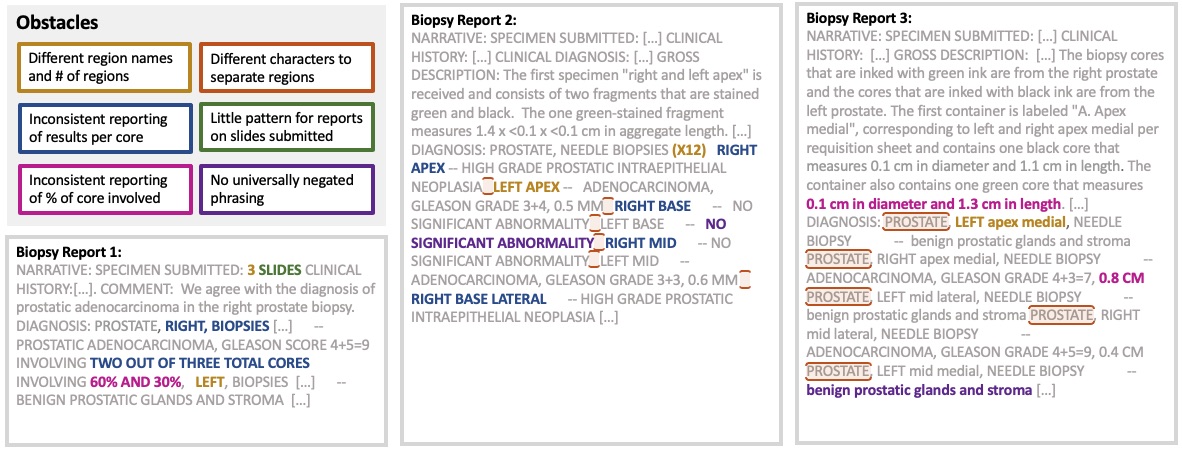}
\caption{\textbf{Obstacles in extraction of patient-level CFG.} 
Representative excerpts from biopsy and pathology notes demonstrating challenges for rule-based extraction, such as varying region names, inconsistent reporting of results per core, diverse formatting, and irregular negation phrasing. 
These examples motivate the need for flexible, context-aware methods such as SNOW to reliably recover clinically meaningful features from unstructured text.}
\label{fig:data-obstacles}
\end{figure}

The necessity of this adaptive, per-patient approach was driven by significant heterogeneity in the source documentation. As illustrated in Figure \ref{fig:data-obstacles}, we identified six primary obstacles that precluded reliable rule-based extraction: (1) inconsistent naming conventions for anatomical regions; (2) variable delimiters for regional data; (3) diverse reporting formats (e.g., grouped vs. separate core reporting); (4) structural variability in outside institution slides; (5) inconsistent quantitative notation (e.g., reporting raw lengths requiring normalization vs. percentages); and (6) irregular negation patterns.

For example, accurately calculating the ``percent core involvement'' often required parsing narrative descriptions where only the tumor length was stated, necessitating a manual search for the corresponding core length elsewhere in the gross description to compute the derived value. This process was repeated for the full cohort of 147 patients treated between 2005 and 2015, with all curation performed on pre-treatment notes to maintain blindness to the recurrence outcome. 

The resulting CFG dataset provides a granular, expert-verified depiction of each patient’s disease state, including derived features often absent from structured EHR data (e.g., percent Gleason pattern 4/5). This manually curated dataset serves as the ``Oracle'' ground truth against which other methods are evaluated in this study.

\subsection{\AFG, an agent-based Scalable Note-to-Outcome Workflow}
\input{agentic_method.tex}

\subsection{Representational Feature Generation (RFG)}
\label{sec:rfg-method}
\input{new_rfg.tex}

\subsection{Clinician-Guided LLM Feature Generation (CLFG)}
\input{llm_method.tex}

\subsection{Outcome Prediction Machine Learning Models}
\label{sec:patient-level-methods-ml-models}
\input{new_outcome_model.tex}

\subsection{External Validation Cohort}
\input{mimic.tex}

\section{Results}
\label{sec:patient-level-results}
\input{new_results.tex}

\section{Discussion}
\label{sec:patient-level-discussion}
\input{new_discussion.tex}

\section{Code Availability}
\label{sec:code-availability}
\input{code_availability}

%% file: new_intro.tex
The widespread adoption of electronic health records (EHRs) has created an unprecedented opportunity to leverage artificial intelligence (AI) for precision medicine. A primary application of clinically-oriented AI is risk stratification, where models utilize patient data to predict trajectories and guide treatment allocation\citep{see_postoperative_2007, bayati_data-driven_2014, henry_targeted_2015, obermeyer_predicting_2016, wiens_patient_2016, chen_machine_2017, komorowski_artificial_2018,rajkomar_scalable_2018,tomasev_clinically_2019}. However, the predictive power of these models is fundamentally limited by the quality of their inputs. While structured data (e.g., past diagnosis codes, laboratory values) are easily accessible, they often lack the granularity required for nuanced clinical decision-making. The richest phenotypic information, symptom severity, disease progression logic, and response to therapy, remains locked within unstructured physician progress notes\citep{xu_feature_2012,ford_extracting_2016,alice_zheng_feature_2018,si_enhancing_2019,hsu_characterizing_2020}. Consequently, the methodology used to distill these complex narratives into structured features (covariates), and the degree to which clinical expertise permeates this process, determines both the performance and the trustworthiness of the resulting models\citep{verduijn_temporal_2007, delisle_combining_2010, zhao_combining_2011, singh_incorporating_2015, moskovitch_procedure_2017, roe_feature_2020}.

Current approaches to feature generation generally fall into two opposed paradigms. At one end of the spectrum is \emph{Representational Feature Generation (RFG)}, which prioritizes scalability by minimizing human input. These fully automated methods range from sparse bag-of-words features to dense representations learned by pre-trained clinical language models (e.g., ClinicalBERT) and newer instruction-tuned foundation models, mapping raw notes to vectors or predictions end-to-end\citep{devlin_bert_2019,alsentzer_publicly_2019, liu_deep_2018, gehrmann_comparing_2018,hahn_medical_2020,huang_clinicalbert_2020, dhrangadhariya_classification_2021}. While RFG effectively utilizes vast corpora without annotation costs, the resulting features are often opaque ``black boxes.'' This lack of interpretability raises significant concerns in high-stakes clinical settings, regarding susceptibility to spurious correlations and the potential amplification of systemic biases\citep{obermeyer_predicting_2016, gianfrancesco_potential_2018,obermeyer_dissecting_2019,seyyed-kalantari_underdiagnosis_2021,ghassemi_medicine_2022}.

At the opposite end of the spectrum lies what we term \emph{Clinician Feature Generation (CFG)}. This approach represents the historical and current ``gold standard'' for high-quality clinical evidence generation, akin to the rigorous manual abstraction protocols employed by national tumor registries (e.g., SEER, NCDB) and adjudicated clinical trials\citep{chapman2012workload,tangka2016cost,casalino_us_2016,nguyen2020generating, fries_ontology-driven_2021}. In CFG, domain experts apply clinical judgment to define meaningful concepts (e.g., ``symptomatic progression'') and manually parse messy texts to extract accurate values on a per-patient basis\citep{alzubi2021ehrabstraction, senders2020automating}. While this process yields highly interpretable and granular data, it imposes a prohibitive administrative burden\citep{saraswathula2023volume,agatstein_chart_2023}. In contrast to scalable but opaque RFG pipelines, manual CFG yields granular, interpretable features but requires substantial clinician time per-patient, creating a practical bottleneck for clinical research.

Recent advances in Large Language Models (LLMs) have introduced a promising middle ground: \emph{Clinician-Guided LLM Feature Generation (CLFG)}. Studies demonstrate that general-purpose models like ChatGPT and DeepSeek R1, as well as domain-specific models such as ClinicalMamba and GatorTron, can accurately extract structured information when guided by expert prompts or fine-tuning\citep{huang2024chatgptstructured, ntinopoulos2025large, yang-etal-2024-clinicalmamba, yang2022large,nori_capabilities_2023}. While these semi-automated features enhance prediction accuracy compared to raw embeddings\citep{anderson_paging_2025, mcinerney-etal-2023-chill,goel_llms_2023}, they do not fully solve the scalability bottleneck. CLFG still relies heavily on domain experts to \emph{a priori} define target features, construct task-specific prompts, and manually validate the logic for each new predictive task. Furthermore, whereas emerging agentic systems can iteratively refine intermediate outputs through self-correction and re-planning\citep{shinn_reflexion_2023,madaan_selfrefine_2023,goodell_clinical_agents_2025}, static CLFG workflows typically lack such mechanisms, limiting their robustness to the heterogeneity and temporal structure of longitudinal clinical notes.

In many EHR-based studies, the limiting factor is not the model architecture but rather the notes-to-features latency: producing an auditable, clinically faithful set of structured features from narrative notes often requires a significant number of hours of manual chart abstraction. This leads to constrained cohort sizes, slower iteration across endpoints, and limited reproducibility across institutions. To overcome many of these limitations, our purpose was to build a workflow that compresses expert feature generation into a reusable and transparent system, preserving clinical reasoning while making note-derived feature generation fast and rigorous to support iterative clinical research. To resolve the tension between scalability and clinical fidelity, we introduce the \emph{Scalable Note-to-Outcome Workflow (SNOW)}. As illustrated in Figure~\ref{fig:main}(b), SNOW is a transparent, multi-agent LLM system that mirrors the steps of manual feature generation: it proposes candidate variables, grounds them in cohort-wide evidence, and iteratively refines extraction rules through self-correction to produce auditable patient-level feature tables at scale.

%% file: agentic_method.tex
Having systematically defined the CFG process, we then designed SNOW to transparently assist in scaling this methodology. SNOW consists of a sequence of specialized LLM agents, each corresponding to a core subtask in our systematic CFG pipeline. This transparent architecture allows for optional human oversight at each stage while enabling scalable deployment.

\begin{figure}[htbp]
\centering
\includegraphics[width=0.9\linewidth]{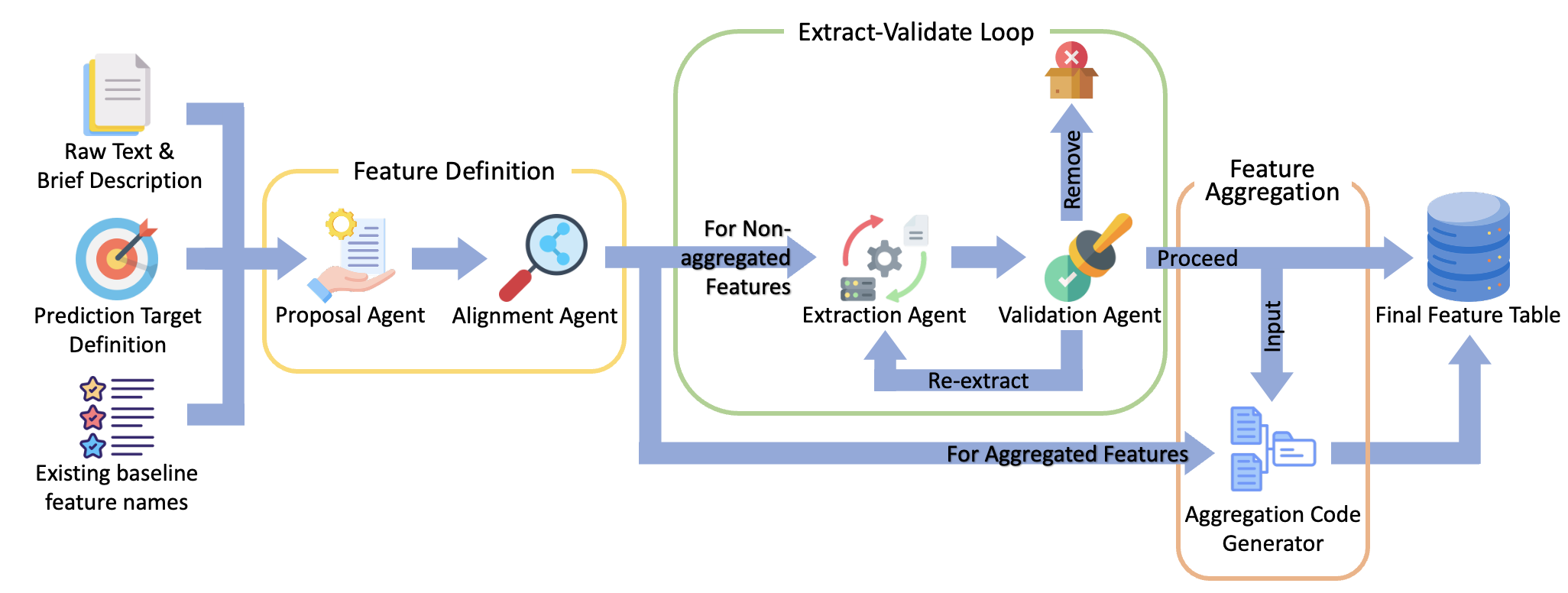}
\caption{\textbf{Architecture of SNOW, a modular multi-agent Scalable Note-to-Outcome Workflow.} 
SNOW decomposes feature generation into specialized LLM agents: the Proposal and Alignment Agents for feature definition, and the Extraction and Validation Agents operating in a loop to refine extraction logic. The Aggregation Code Generator compiles aggregated features from extracted values.
Arrows indicate information flow between agents. The system allows for human oversight at each stage, enabling experts to review, refine, or override intermediate outputs.}
\label{fig:SNOW}
\end{figure}

To transparently and scalably implement systematic CFG process, we designed a modular agentic system, SNOW (Scalable Note-to-Outcome Workflow), where each agent corresponds to a specific CFG subtask (Figure~\ref{fig:SNOW}). In this context, an ``agentic system'' refers to a collection of distinct LLMs, each governed by a specialized system prompt and set of tools tasked with a specific function within the pipeline (full system prompts and LLM specifications are described in Appendix~\ref{appendix:llm-specs}). Unlike end-to-end black-box models, SNOW agents receive only a natural-language description of the prediction target and clinical context, operating through three distinct stages: Feature Definition, the Extract-Validate Loop, and Feature Aggregation.

While SNOW can operate fully autonomously, its design explicitly exposes intermediate outputs to allow clinicians to review and refine the process at key decision points: (1) reviewing proposed features after Feature Definition, (2) approving extraction logic after the Extract-Validate Loop, and (3) validating aggregation code. This flexibility enables institutions to calibrate the balance between automation efficiency and clinical oversight.

\subsubsection{Feature Definition}

The Feature Definition stage involves two sequential agents that mimic the clinical reasoning process of hypothesis formation followed by data-driven adaptation (Figure~\ref{fig:SNOW}, yellow panel).

Given only the prediction-task description, the \emph{Feature Proposal Agent} reads a random sample of (we used $6$) notes and proposes an initial set of candidate variables that are clinically interpretable and suitable for modeling. It avoids duplicating variables available from structured baseline features, annotates feature type (numeric, categorical, boolean), and flags subgroup or repeated-context structures (e.g., anatomical sites or timepoints) that may require later aggregation.

The \emph{Feature Alignment Agent} subsequently scans clinical notes to refine and expand the feature set based on actual documentation patterns. Designed for adaptability, this agent functions as an optional domain-calibration step: it can be configured to scan the full corpus, restricted to a representative sample to reduce latency, or omitted entirely to maximize speed. This architecture ensures that SNOW remains broadly generalizable via the initial Proposal Agent, while offering the flexibility to capture institution-specific reporting nuances when enabled.

\subsubsection{Extract-Validate Loop}

The core of the system is the iterative Extract-Validate Loop (Figure~\ref{fig:SNOW}, green panel), designed to handle the noise and variability of clinical text.

The \emph{Feature Extraction Agent} processes individual clinical notes to extract values for each proposed feature. For initial extraction attempts, it applies instructions generated by the Feature Proposal Agent. For subsequent re-extractions, it incorporates refined guidance provided by the Validation Agent.

Following extraction, the \emph{Feature Validation Agent} performs quality control by reviewing a sample of clinical notes alongside the extracted values to assess accuracy, completeness, and consistency. Based on this assessment, it makes one of three decisions: proceed with the feature; remove it due to persistent poor quality or infeasibility; or re-extract with revised instructions. This initiates a feedback loop in which features flagged for re-extraction return to the Extraction Agent with updated instructions. 

This iterative refinement allows the system to autonomously learn nuances of feature extraction that typically require expert knowledge, such as handling inconsistent terminology, resolving negation, or normalizing units (see Appendix \ref{appendix:SNOW-examples} for examples of iterative refinement). The loop continues until the Validation Agent either approves the feature or determines that it cannot be consistently extracted.

\subsubsection{Feature Aggregation}

For features defined as aggregates of previously extracted base features, such as the maximum Gleason score across anatomical regions or the percentage of positive biopsy cores, the \emph{Aggregation Code Generator} (Figure~\ref{fig:SNOW}, orange panel) creates executable Python code to compute these values. The generated code handles missing values appropriate to the data type and adheres to the intended computation logic, ensuring clinical validity without manual coding.

\subsubsection{Clinical Oversight}

SNOW is designed to support clinician involvement at every major stage of the workflow, enabling expert review without requiring continuous supervision. This oversight capability ensures that feature generation remains transparent, auditable, and clinically grounded. The system exposes all intermediate artifacts, including proposed feature lists, extraction instructions, extracted values, and aggregation code, so that clinicians can verify correctness and modify system behavior when necessary.

Clinicians may intervene through three primary mechanisms:
\begin{enumerate}
    \item \emph{Feature Review and Refinement:} After the Feature Proposal and Alignment stages, clinicians can review candidate feature lists to assess clinical relevance, remove implausible or redundant features, and manually add variables based on domain knowledge.
    \item \emph{Instruction and Extraction Oversight:} During the Extract-Validate loop, clinicians may audit extraction behavior by examining representative note-value pairs and suggesting corrections to the extraction instructions. These corrections are incorporated directly into subsequent extraction cycles.
    \item \emph{Aggregation Logic Verification:} For derived features, clinicians can review the auto-generated aggregation code to ensure that clinical intent is preserved and aligns with practice guidelines.
\end{enumerate}

In this study, clinicians exercised oversight at key points to confirm the plausibility and clinical validity of the system’s outputs, but the system itself performed all feature definition, extraction, and aggregation. This hybrid design allows SNOW to scale the manual clinician feature generation process while preserving the transparency and control required for rigorous clinical research.

%% file: new_rfg.tex
To retrieve features from progress notes using RFG, we apply different natural language processing (NLP) methods. For each patient, we create a concatenated progress note text, by combining the pre-treatment clinical note that has clinical T stage information present closest to the start date of treatment, identified using regular expressions, with the pre-treatment biopsy report from which we retrieve the subset of CFG features per systematic core (i.e., grade group, percent involved, and percent Gleason pattern 4/5 per systematic core). We then apply three different NLP methods to this concatenated text to generate features: Bag-of-Words (BoW Classic), Bag-of-Words Term Frequency-Inverse Document Frequency (BoW TF-IDF), and the Gemini Embedding model \citep{lee2025geminiembeddinggeneralizableembeddings}. 

BoW provides a transparent and interpretable representation of clinical text. To capture varying levels of local context, we constructed BoW Classic and BoW TF-IDF embeddings with n-gram features ranging from unigrams to 5-grams, with a maximum vocabulary size of 10,000 features. BoW Classic refers to the version in which the feature value for each n-gram is the raw frequency count of its appearance in a clinical note. BoW TF-IDF is a variant that uses the same n-gram extraction and preprocessing pipeline but reweights n-gram frequencies using term frequency–inverse document frequency to downweight commonly occurring terms that are less informative across the corpus. We constructed the corpus vocabulary from all 147 patient notes.

To assess whether embedding models could extract clinically meaningful signal beyond sparse lexical representations, we evaluated embeddings generated using the latest Gemini Embedding model, which achieves leading performance on the HuggingFace MTEB benchmark.\citep{muennighoff2023mteb} This allowed us to assess whether general-purpose, proprietary embedding models could extract clinically useful signal in comparison to BoW methods.  We obtained document-level vectors from the Gemini embedding model. For notes exceeding model token limits, we embedded overlapping chunks and mean-pool the resulting vectors to form a single patient-level representation.

To further explore domain-adapted representation learning, we developed a fine-tuned embedding model based on the open-source Mistral-7B-v0.3 language model \citep{jiang2023mistral7b}. The fine-tuning corpus consisted of 16,872 pathology reports, with all notes corresponding to patients in the study cohort explicitly excluded to prevent information leakage. Fine-tuning was performed using a causal language modeling objective, training the model autoregressively to predict the next token given preceding context. This objective encourages the model to learn domain-specific semantic structure that can be leveraged for downstream embedding extraction.

We fine-tuned the model using Low-Rank Adaptation (LoRA) \citep{hu2021loralowrankadaptationlarge} with 8-bit quantization to improve computational efficiency. LoRA was configured with rank 64 and scaling factor 128, targeting the query, key, value, and embedding layers, with a dropout rate of 0.1 and no bias training. Training was conducted for five epochs with a per-device batch size of 2 and gradient accumulation over 8 steps, yielding an effective batch size of 16, and was performed on a single NVIDIA A100 GPU. After fine-tuning, we generated embeddings using both mean pooling over all token-level hidden states and by extracting the hidden state corresponding to the end-of-sequence (EOS) token. Despite this domain adaptation, the fine-tuned Mistral-based embeddings did not outperform the off-the-shelf Gemini Embedding model on the prostate cancer recurrence prediction task. Consequently, results from the fine-tuned Mistral model are not included in the final analyses.

For all RFG methods, we implemented two preprocessing approaches: standard preprocessing and domain-informed preprocessing optimized for clinical text recommended by GPT-5,\citep{openai_gpt5_2025} which we refer to as LLM-informed RFG. For BoW Classic and BoW TF-IDF, standard preprocessing included (1) text normalization (lowercasing, whitespace collapsing); (2) lemmatization; (3) punctuation removal; (4) removal of common English stopwords. Domain-informed proccessing enhanced this by: (1) removing non-medical content (URLs, email addresses); (2) standardizing measurements (inserting spaces between numbers and units, e.g., ``3.5cm'' $\to$ ``3.5 cm''); (3) removing punctuation while preserving clinical codes (e.g., ``T2'', ``N0'', ``ER+''); and (4) crucially, employing a clinical stopword list that retained semantically important tokens typically discarded in generic NLP pipelines, including negations, polarity terms, measurement comparators, and quantifiers.

For the Gemini Embedding model, the standard approach was to use raw text without preprocessing because it is a LLM–based embedding trained on raw text, and is designed to operate directly on natural language without additional normalization. For domain-informed preprocessing, recommended by GPT-5, we applied lighter normalization to preserve sentence structure and clinical context: we removed URLs and emails, collapsed whitespace, and de-duplicated excessive punctuation while keeping medical terminology and numbers intact.

To ensure consistency across RFG methods and mitigate potential overfitting in our relatively small dataset, we applied singular value decomposition (SVD) to reduce the dimensionality of the feature matrix for each method. The number of retained components was treated as a tunable hyperparameter, optimized in the model selection stage (Section~\ref{sec:patient-level-methods-ml-models}). The candidate dimensions evaluated were [30, 60, 90, 120].


%% file: llm_method.tex
In settings where large, manually generated features exist, feature-specific NLP extraction systems (rule-based or supervised) can be trained to recover structured cancer variables such as stage, grade, and therapies at high accuracy.~\cite{wang2019natural}. In our prostate cohort ($n=147$) and across a broader clinician-defined feature set, training and reliably validating multiple supervised extractors would be underpowered. We therefore explored alternatives using LLM prompting, that we call optimized clinician-guided feature generation (CLFG).

Specifically, we explored using LLMs to generate patient-level features through expert-guided prompts. In each prompt, we ask the LLM to identify and extract the same set of clinician-defined features used in CFG. To support accurate extraction, we include detailed instructions based on our experience manually processing these features. For instance, when core-length involvement is reported as a raw measurement (e.g., in millimeters) rather than a percentage, we instruct the model to perform the necessary division. The LLM processes each note individually, extracting all relevant features one note at a time. For features that depend on other extracted values, such as percent Gleason pattern 4/5 or maximum Gleason score which are derived from individual Gleason scores, we apply post-processing code that follows clinical rules based on expert knowledge and standard medical guidelines.

%% file: new_outcome_model.tex
To obtain an unbiased estimate of model performance with sample size of $n=147$, we use nested cross-validation. Specifically, we implement a 3-fold outer cross-validation to evaluate generalization performance. For each outer fold, the remaining two-thirds of the data, i.e. the outer training set, are used to perform a 3-fold inner cross-validation to select the optimal hyperparameters of various machine learning models based on area under the receiver operating characteristic curve (AUC-ROC) as the performance metric (see Appendix~\ref{appendix:aupr-f1} for sensitivity analysis with area under the precision-recall curve, as well as F1 measure). After hyperparameter tuning within the inner folds, we retrain the model on the full outer training set using the best hyper parameters, and evaluate its performance on the held-out outer fold. This procedure ensures that the outer test data remain entirely separate from both model fitting and hyperparameter selection.

Since it is known that formal statistical tests comparing AUC-ROC between different models are unreliable with small samples~\citep{newcombe_confidence_2006, feng_comparison_2017}, we repeat the entire nested cross-validation procedure 50 times, each time using a different random seed to generate the outer and inner cross-validation splits. We then compare the distribution of the AUC-ROC of the models across the 50 seeds.

We evaluated three machine learning models with hyperparameter tuning via nested cross-validation: (1) \emph{Logistic Regression} with L1 and L2 penalties; (2) \emph{Random Feature Model}, a custom kernel approximation classifier; and (3) \emph{K-Nearest Neighbors} (KNN). 

The Random Feature Model was implemented as follows: We first generated $d$ random Gaussian vectors $\mathbf{w}_1, \ldots, \mathbf{w}_d \sim \mathcal{N}(0, \mathbf{I})$ and normalized each to unit length. For an input feature vector $\mathbf{x} \in \mathbb{R}^p$, the random feature transformation was defined as $\phi(\mathbf{x}) = [\text{ReLU}(\mathbf{x}^T\mathbf{w}_1), \ldots, \text{ReLU}(\mathbf{x}^T\mathbf{w}_d)]^T$, where $\text{ReLU}(z) = \max(0, z)$. This transformation approximates a kernel function through random projection. A ridge classifier was then trained on the transformed features $\phi(\mathbf{x})$, providing a computationally efficient approximation to kernel methods while maintaining linear scalability with sample size.

We used the following parameter grids: Logistic Regression with inverse regularization parameter $C \in [10^{-4}, 1]$ (9 logarithmically-spaced values) and both liblinear and saga solvers; Random Feature Model with $d \in \{64, 128, 256, 512\}$ random projections and regularization $\alpha \in [10^{-4}, 10^{4}]$ (9 logarithmically-spaced values); KNN with $k \in \{5, 7, 9, 11, 13\}$ neighbors. All models considered both uniform and distance-based weighting schemes and both Euclidean and Manhattan distance metrics where applicable.

Missing values were handled using singular value decomposition (SVD) imputation. A sensitivity analysis utilizing other imputation methods, e.g. MICE, is included in Appendix~\ref{appendix:imputation-analysis}.

%% file: mimic.tex
To evaluate the generalizability of SNOW beyond prostate cancer, we applied it to a distinct clinical prediction task: mortality prediction in patients hospitalized with heart failure with preserved ejection fraction (HFpEF). Unlike the prostate cancer cohort where we conducted extensive manual CFG over a year-long period, we did not perform CFG or CLFG for the HFpEF cohort due to the manual and resource-intensive nature of these approaches. Instead, we focus on comparing SNOW's performance against baseline features and RFG methods. This comparison tests SNOW's ability against existing automated methods without requiring expert involvement, which is critical for assessing its practical utility in real-world settings where manual curation is often expensive.

We constructed a validation cohort using the MIMIC-IV (version 3.1) and MIMIC-IV-Note (version 2.2) databases, publicly available deidentified electronic health record datasets from Beth Israel Deaconess Medical Center via PhysioNet \citep{mimiciv-v3-1-physionet-2024,johnson2023mimiciv,mimicivnote-v2-2-physionet-2023,goldberger2000physionet}. Our cohort definition, baseline feature set, and prediction targets were informed by prior work on HFpEF mortality prediction \citep{shin2024predicting}, adapted to match data availability and modeling requirements in our setting.

We identified 2,084 adult patients (age $\geq$ 18 years) in MIMIC-IV for whom HFpEF was recorded as the primary diagnosis, based on ICD-9 and ICD-10 codes. For patients with multiple qualifying hospitalizations, we selected the earliest as the index admission. Patients who died during the index admission were excluded to enable measurement of post-discharge outcomes.

To ensure availability of unstructured clinical text, we required that the index admission had an associated discharge summary in MIMIC-IV-Note, prior to patient's discharge.
This consolidated discharge summary served as the unstructured input to SNOW. We gathered structured baseline features from MIMIC-IV, including demographics, vital signs, laboratory measurements, and comorbidity indicators (complete list provided in Appendix \ref{appendix:hfpef-features}). The prediction targets were 30-day and 1-year all cause mortality following hospital discharge.

We applied the same feature generation and modeling approach used for the prostate cancer 
cohort. Specifically, we used SNOW to extract features from discharge summaries, applied 
the RFG methods described in Section~\ref{sec:rfg-method} (BoW Classic, 
BoW TF-IDF, and Gemini embeddings), and evaluated performance using the machine learning 
models described in Section~\ref{sec:patient-level-methods-ml-models}, via nested cross-validation. The only modifications were to the hyperparameter 
grids, which we expanded to accommodate the larger sample size: Logistic Regression with $C \in [10^{-5}, 10]$ (7 logarithmically-spaced values); Random Feature Model with $d \in \{256, 512, 1024, 2048, 4096\}$ random projections and the same regularization range; KNN with $k \in \{5, 10, 20, 30, 40, 50\}$ neighbors. The increased dimensionality of random features and larger neighborhood sizes for the HFpEF cohort reflect the greater statistical power available with more training samples, allowing for more complex model architectures without overfitting risk. Additionally, for RFG methods requiring tunable dimensionality reduction, the candidate dimensions were [50, 100, 200, 300]. For simplicity, median imputation was applied to missing values in this larger cohort.

%% file: new_results.tex
\begin{figure}[H]
\centering
\includegraphics[width=0.9\linewidth]{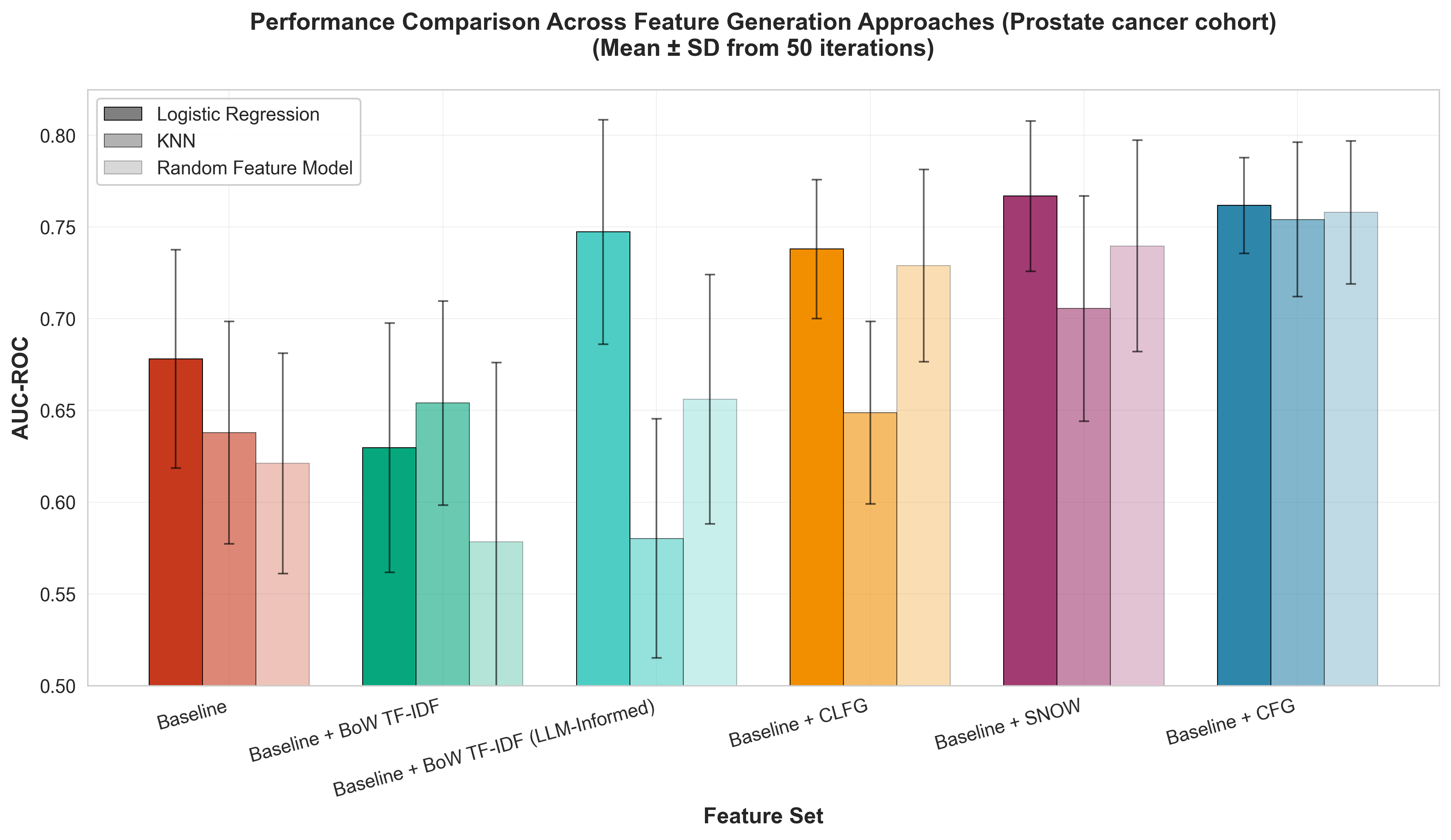}
\caption{\textbf{Performance of feature generation methods for 5-year prostate cancer recurrence prediction.} 
Distributions of area under the receiver operating characteristic curve (AUC-ROC) across 50 repetitions of nested cross-validation for regularized logistic regression, $k$-nearest neighbors, and random feature models trained with different feature sets: Baseline only, Baseline + Bag-of-Words (BoW) TF–IDF, Baseline + BoW TF–IDF, Baseline + CLFG, Baseline + SNOW, and Baseline + CFG. 
CFG substantially improves performance relative to Baseline, and SNOW achieves AUC-ROC comparable to CFG while outperforming all RFG approaches, indicating that task-adapted agentic feature generation better harvests prognostic signal from notes than off-the-shelf embeddings. Among RFG methods, BoW TF-IDF is the best-performing non–LLM informed variant and BoW TF-IDF (LLM-informed) is the best-performing LLM-informed RFG variant; a full comparison of all RFG methods is provided in Appendix \ref{appendix:rfg_comparison}.}
\label{fig:prostate_main}
\end{figure}

Figure~\ref{fig:prostate_main} presents the distribution of AUC-ROC scores across 50 iterations of nested cross-validations on Logistic Regression, KNN, and Random Feature Model, comparing six feature sets: Baseline, Baseline + BoW TF-IDF (best-performing non-LLM-informed RFG), Baseline + BoW TF-IDF (LLM-informed), best-performing RFG, Baseline + CLFG, Baseline + \AFG, and Baseline + CFG. In the sections below, we report performance metrics (mean $\pm$ standard deviation) derived from the regularized Logistic Regression classifier, as it consistently demonstrated the highest discrimination among the evaluated machine learning models.

The ‘Baseline + CFG’ model (AUC-ROC: $0.762 \pm 0.026$) achieves a substantially higher mean AUC-ROC compared to the ‘Baseline’ model (AUC-ROC: $0.678 \pm 0.060$) across all three machine learning models, demonstrating that clinician-guided patient-level feature generation (CFG) from progress notes significantly enhances predictive performance. This finding aligns with prior work on leveraging unstructured clinical notes for predictive modeling~\citep{hsu_characterizing_2020, liu_deep_2018, ford_extracting_2016}.

The ‘Baseline + CLFG’ model (AUC-ROC: $0.738 \pm 0.038$), which leverages LLMs guided by clinician-written instructions, performs slightly worse than the `Baseline + CFG' model but still substantially better than the `Baseline' model alone, suggesting that LLMs can effectively extract information from unstructured notes following expert guidance. However, this approach requires substantial expert prompt optimization.

Notably, the `Baseline + \AFG' model (AUC-ROC: $0.767 \pm 0.041$), which efficiently scales the feature generation process using a multi-agent LLM system, achieves performance on par with the manual CFG process. This result demonstrates that our agentic LLM system can effectively scale labor-intensive expert-driven CFG. The feature specifications generated by SNOW are provided in Appendix~\ref{appendix:afg-features}. Furthermore, feature importance analysis confirms that SNOW, similar to manual CFG, shifts the model's reliance away from sociodemographic proxies (e.g., marital status) toward granular clinical metrics, thereby capturing the intended prognostic signal (see Appendix \ref{appendix:feature_analysis}).

RFG methods performed notably worse. The strongest variants, Baseline + BoW TF-IDF and Baseline + BoW TF-IDF (LLM-informed), achieved AUC-ROC values of $0.630 \pm 0.068$ and $0.747 \pm 0.061$ with regularized Logistic Regression, respectively. However, the apparent improvement from LLM-informed preprocessing was not robust: performance dropped sharply to $0.580 \pm 0.065$ with KNN, indicating substantial machine learning model sensitivity. Appendix~\ref{appendix:rfg_comparison} provides a full comparison of all RFG variants and discusses why sparse lexical representations outperform dense embeddings in this setting.

The sensitivity analyses show that the conclusions are
consistent with varying imputation methods (Appendix~\ref{appendix:imputation-analysis}), removal of features race, ethnicity, and language (Appendix~\ref{appendix:sociodemographic_analysis}), alternative definition of the outcome (Appendix~\ref{appendix/alternative-BF}), cohort selection (Appendix~\ref{appendix:168_analysis}), and performance metric (Appendix~\ref{appendix:aupr-f1}).

\subsection{Efficiency and Research-Cycle Acceleration} 

Beyond predictive performance, we evaluated the operational efficiency of feature generation, contrasting the resource requirements of the manual CFG process against the SNOW pipeline. The CFG protocol was resource-intensive, necessitating a year-long collaborative effort. This process involved bi-weekly multidisciplinary meetings attended by clinician and data science investigators, residents, and researchers to define concepts, followed by a rigorous manual abstraction phase performed by clinically trained experts. In total, generating the gold-standard feature set for 147 patients required an estimated 240 hours of human expert time (Table~\ref{tab:efficiency}).

In contrast, SNOW decoupled the scale of extraction from human effort. We compared the deployment time of the pre-built SNOW system against the execution time of the manual protocol. Once the prediction task was defined, the system generated the full patient-level feature table in approximately 12 hours of wall-clock time. Importantly, the burden on domain experts shifted from manual abstraction to high-level auditing: clinicians spent only 5 hours reviewing feature proposals and validating extraction logic during the Extract-Validate loop. This represents nearly a 48-fold reduction in human effort for the specific task. Furthermore, while the manual effort scales linearly with cohort size, rendering granular extraction for large populations like the HFpEF cohort (N=2,084) impractical, SNOW's marginal cost per additional patient is negligible.

\begin{table}[htbp] 
\centering 
\caption{
\textbf{Deployment efficiency comparison: manual vs. agentic feature generation.} 
Time estimates reflect effort required for the prostate cancer cohort (N=147). 
The agentic workflow (SNOW) achieved a 48-fold reduction in human expert time while reducing deployment latency from over 12 months to under 1 week.
} 
\label{tab:efficiency} 
\begin{tabularx}{\textwidth}{@{}lXX@{}} 
\toprule 
& \textbf{Manual Effort for CFG} & \textbf{SNOW} \\
\midrule 
\textbf{Deployment Time} & & \\
\quad Calendar time & $>$12 months & $<$1 week \\[2pt]
\quad Human expert effort & $\sim$240 person-hours & $\sim$5 person-hours \\[6pt]

\textbf{Effort Breakdown} & & \\
\quad Task definition & $\sim$190 hrs & $<$1 hr \\
& \small (Year-long team meetings) & \small (Initial configuration) \\[2pt]
\quad Feature extraction & $\sim$50 hrs & 0 hrs \\
& \small (Manual chart review) & \small (Automated) \\[2pt]
\quad Validation & $\sim$5 hrs & $\sim$4 hrs \\
& \small (Spot checking) & \small (Log review) \\[6pt]

\textbf{Scalability} & Linear: $O(N)$ & Constant: $O(1)$ \\
& \small Impractical for large cohorts & \small Human effort fixed \\
\bottomrule 
\end{tabularx} 
\end{table}

\subsection{Clinical Oversight and Qualitative Analysis}
\input{clinical_insights}

\subsection{External Validation Result}

\begin{figure}[H]
\centering
\includegraphics[width=0.9\linewidth]{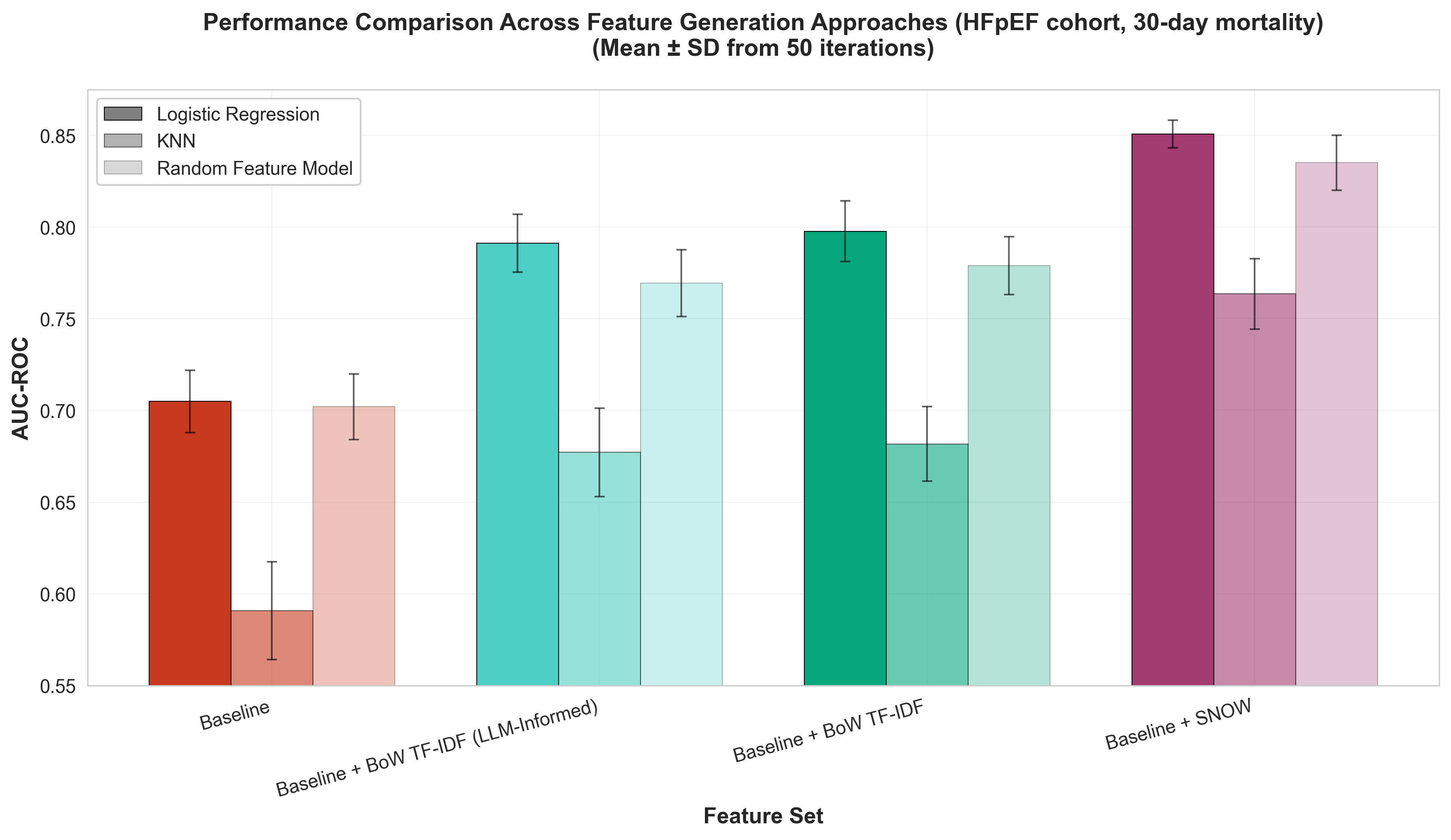}
\caption{\textbf{External validation of SNOW for 30-day mortality prediction in HFpEF.} 
Distributions of AUC-ROC across 50 repetitions of nested cross-validation for models predicting 30-day all-cause mortality after discharge in patients with HFpEF in MIMIC-IV dataset. 
Feature sets include Baseline only, Baseline + BoW TF–IDF, Baseline + BoW TF–IDF (LLM-informed), and Baseline + SNOW. 
Augmenting baseline features with SNOW consistently yields the highest mean AUC-ROC and reduced variability, demonstrating that the agent-based feature generation approach generalizes to a new disease area and dataset without task-specific manual curation. Among RFG methods, BoW TF-IDF is the best-performing non–LLM-informed variant and BoW TF-IDF (LLM-informed) is the best-performing LLM-informed RFG variant; a full comparison of all RFG methods is provided in Appendix \ref{appendix:rfg_comparison}.}
\label{fig:hfpef_30days_main}
\end{figure}

\begin{figure}[H]
\centering
\includegraphics[width=0.9\linewidth]{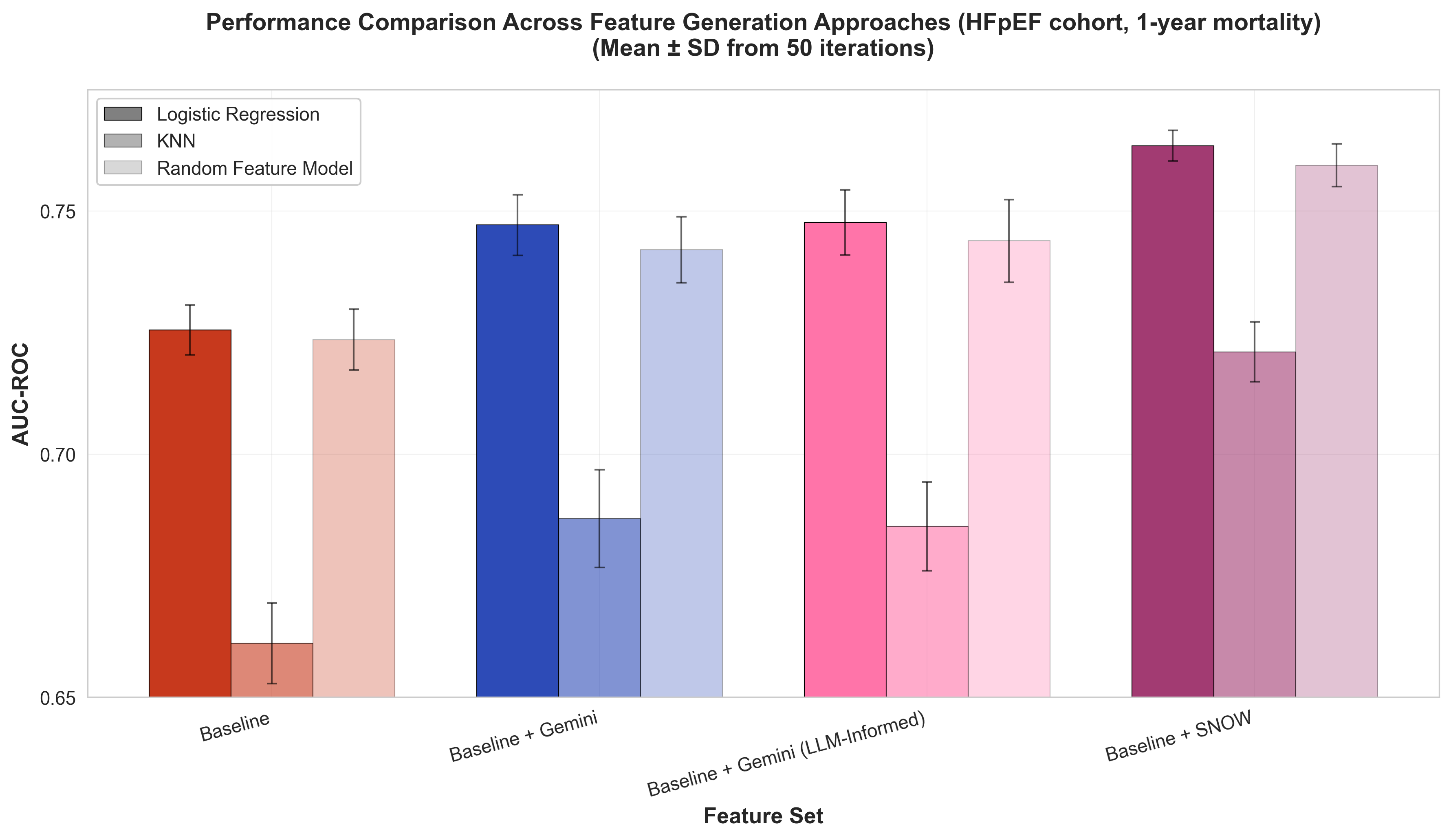}
\caption{\textbf{External validation of SNOW for 1-year mortality prediction in HFpEF.} 
Distributions of AUC-ROC across 50 repetitions of nested cross-validation for models predicting 1-year all-cause mortality after discharge in the HFpEF cohort. 
As for 30-day mortality, models using Baseline + SNOW features outperform those using Baseline alone or Baseline combined with RFG features. 
These results indicate that SNOW-derived features provide stable, incremental prognostic signal from discharge summaries over longer time horizons in an independent health system. Among RFG methods, Gemini embeddings is the best-performing non–LLM-informed variant and Gemini embeddings (LLM-informed) is the best-performing LLM-informed RFG variant; a full comparison of all RFG methods is provided in Appendix \ref{appendix:rfg_comparison}.}
\label{fig:hfpef_1year_main}
\end{figure}

Figure~\ref{fig:hfpef_30days_main} and Figure~\ref{fig:hfpef_1year_main} show the distribution of AUC-ROC scores across 50 iterations of nested cross-validation for 30-day and 1-year mortality prediction in the HFpEF cohort, respectively, comparing all feature generation approaches.

Across both prediction tasks, adding SNOW features to the structured baseline features yields the strongest performance. The ‘Baseline + \AFG’ models (30-day AUC-ROC: $0.851 \pm 0.008$, 1-year AUC-ROC: $0.763 \pm 0.003$) achieve the highest mean AUC-ROC and exhibit lower standard deviations than the other feature sets, indicating that SNOW features provide robust incremental signal beyond baseline features alone. This effect is observed for both 30-day and 1-year mortality, suggesting that the agentic feature generation process can extract clinically meaningful information from discharge summaries in a disease area for which no manual CFG or CLFG was performed.

In contrast, augmenting the baseline features with BoW TF-IDF representations yields only modest performance gains over the Baseline model. These RFG approaches underperform the ‘Baseline + \AFG’ models for both time horizons. Together, these results indicate that while automated representational features can capture some additional risk information from clinical text, they do not fully exploit the prognostic content of discharge summaries in this setting.

Taken together with the prostate cancer results, the HFpEF external validation demonstrates that SNOW generalizes beyond its development cohort: without any task-specific manual feature curation or clinician-written extraction instructions, the \AFG{} features consistently enhance mortality prediction over both structured baselines and RFG methods in a large, independent public dataset.

%% file: clinical_insights.tex
Although SNOW operates autonomously, we subjected its decision-making process to rigorous clinical auditing to ensure medical validity. Two radiation oncology physicians specializing in prostate cancer treatment evaluated the intermediate and final outputs of the agentic workflow.

\subsubsection{Feature Proposal and Alignment}
The \emph{Feature Proposal Agent} initially generated a candidate list of 24 features (Table~\ref{tab:initial_features}). The clinicians deemed 100\% of these candidate features (24/24) appropriate based on clinical utility and plausibility for outcome modeling. Following the corpus-wide scan by the \emph{Feature Alignment Agent}, this list was expanded to 58 features (Table~\ref{tab:updated_features}). All initial features were retained, and the clinicians confirmed that the expanded set captured a comprehensive range of biologically plausible predictors for prostate cancer recurrence.

One variable in this updated feature set, `immunocompromised status', warranted specific clinical scrutiny. While immunocompromised patients have higher overall mortality, the literature regarding the association between immunosuppression (e.g., transplant, HIV/AIDS) and prostate cancer \emph{recurrence} remains conflicting. Small retrospective studies have suggested higher cancer-specific mortality in specific subgroups,\citep{sanguedolce2025immunocompromised, ramey2015immunocompromised-ebrt, kahn2012hiv-ebrt} yet the biological link to recurrence is not definitively established. The clinicians concluded that while the likelihood of a causal relationship is low, the agent's decision to include it as an exploratory feature at the alignment stage was appropriate given the mixed evidence.

\subsubsection{Agentic Self-Correction and Extraction Validity}

The \emph{Extract-Validate Loop} demonstrated emergent error-correction capabilities, acting as an autonomous data curator before human review. 
As detailed in the logs (Appendix~\ref{appendix:SNOW-examples}), the Validation Agent iteratively identified and resolved logic errors that standard extraction would have missed. %
For example, in extracting ``PSA velocity'', the initial pass yielded mathematically correct but clinically implausible outliers (e.g., $>$600 ng/mL/year) driven by short follow-up intervals. %
Rather than halluncinating values or failing silently, the agent autonomously updated the extraction instructions to strictly enforce clinical capping rules (-10 to +40 ng/mL/year) and exclude treatment-confounded measurements, stabilizing the feature without human intervention.

Following this autonomous refinement, the system produced a granular feature set (see Appendix~\ref{appendix:final-features}). 
The clinicians identified no missing signal; the final list captured key tumor metrics, biopsy details, and relevant clinical history. 
The human audit revealed a notable nuance in how the agents defined ``PSA velocity'' compared to clinical convention. The agent defined the feature as calculable with a minimum of two PSA values. While clinically valid, physicians typically recommend at least three values to minimize noise and provide reliable calculations.\citep{vickers2012psa-kinetics} However, the physicians noted that the agent simultaneously extracted ``PSA doubling time'', which models the exponential growth typical of tumors, which ensures the predictive signal was captured, despite the definition discrepancy in velocity.\citep{freedland2001psa-velocity, loeb2008psadt-vs-psav}



\subsubsection{Autonomous Feature Pruning}

A key finding of the oversight process was the system's ability to autonomously identify and discard features that were clinically relevant but practically unusable within this specific dataset. Between the updated feature list and the final model inputs, SNOW dropped nine variables (Appendix~\ref{appendix:final-features}), including `immunocompromised status', `surgical margin status', and `capsular invasion (biopsy)'.

Inspection of the agent logs revealed that these removals were driven by correct data-centric reasoning rather than extraction errors, examples include:
\begin{itemize}
    \item \emph{Immunocompromised status:} The agent removed this due to a ``high missing rate (97.96\%)'' that correctly reflected the true absence (sparsity) of these conditions in the general population, noting in its reasoning ``the rarity... makes the feature unsuitable for meaningful regression modeling.''
    \item \emph{Surgical margin status:} Despite acknowledging its high prognostic value, the agent dropped this feature, reasoning: ``The dataset appears to primarily consist of biopsy reports and pre-surgical evaluations... [this feature] requires the presence of post-surgical pathology data which appears to be absent.''
\end{itemize}
This demonstrates SNOW's capacity to discern between \emph{clinical importance} (what matters in theory) and \emph{data availability} (what matters for the specific model), a distinction usually requiring human judgment. Conversely, the system correctly retained `nerve-sparing surgery', a surgery-specific feature, after identifying sufficient mentions in the non-surgical pathology notes to warrant inclusion.

\subsubsection{Granularity and External Validity}

The final feature set was characterized by high granularity. For instance, SNOW extracted features specifically regarding the ``transition zone'' (Table~\ref{tab:updated_features}). While appropriate for this cohort and clinically plausible, clinicians noted that not all institutions regularly sample the transition zone during standard biopsy protocols.\citep{richard2012transition-zone} By parsing data at this level of granularity, SNOW maximizes predictive power for the training distribution but highlights a potential trade-off: highly specific feature definitions may require domain adaptation when generalizing to datasets with different documentation or sampling protocols.

%% file: new_discussion.tex
Unstructured clinical notes encode clinically nuanced information that are prognostic for important outcomes, yet these signals often remain difficult to use for cohort-scale prediction because they are hard to translate into structured, patient-level variables. In our experiments, scalable representational feature generation (RFG) approaches provided limited incremental benefit over structured EHR variables, suggesting that simply embedding notes does not reliably surface the relevant clinical abstractions. We demonstrated that a systematic clinician feature generation (CFG) protocol yielded improved prediction compared to both structured baselines and RFG, reinforcing the notion that there is valuable signal in narrative text which requires clinically grounded extraction to unlock. Building on this foundation, we introduce SNOW, a transparent multi-agent workflow that executes the CFG process with targeted clinician oversight through auditable intermediate outputs, including feature specification, extraction, validation, and aggregation, without requiring manual per-patient chart abstraction. Across prostate cancer recurrence prediction and an external HFpEF mortality cohort, SNOW achieved clinician-grade performance while demonstrating scalability and preserving interpretability and generalizability. To facilitate reuse and independent evaluation, we have open-sourced the SNOW code, prompts, and documentation for application in new cohorts and tasks.

A key takeaway from this study is an explanation as to why scalable representational feature generation (RFG) approaches can underperform when the goal is to recover clinically grounded variables from narrative notes. Off-the-shelf embeddings and related pretrained text representations can summarize large volumes of text, but they do not, by default, enforce the explicit definitions, consistency constraints, and error checking that clinicians apply during chart abstraction. In our experiments, task-adapted variants of RFG and prompt-based extraction (CLFG) narrowed the gap but remained limited by opacity, incomplete capture of clinically specific constructs, and brittle alignment between narrative phrasing and variable definitions. SNOW addresses these failure modes by treating clinician feature generation as a structured workflow rather than a single modeling step: it proposes candidate features, aligns them to cohort context, produces explicit extraction instructions, validates and repairs outputs, and aggregates raw extractions into modeling-ready variables. Critically, the system surfaces intermediate artifacts for targeted clinician review and refinement, so the clinician’s role shifts from manual per-patient extraction to oversight of definitions, audits, and interpretation.

The principal impact of SNOW is therefore methodological: it lowers the marginal cost of generating clinically nuanced, interpretable features from unstructured notes and shortens the time from clinical question to an analysis-ready feature table.
This distinction is important for interpreting the paper’s contribution. SNOW is not intended to claim discovery of new biological mechanisms; rather, it makes rigorous measurement of clinically meaningful factors feasible at scale, which is often the limiting step in retrospective studies even when the underlying concepts are well understood. 
For example, the percentage of Gleason pattern 4/5 in biopsy tissue has been shown to predict biochemical and clinical outcomes and is used in prognostic modeling, yet it is often embedded in narrative pathology reporting rather than available as a structured variable.~\cite{dambrosio2007gleason45proportion,stoyanova2013doseescalatedrtadt} When such established predictors are expensive to extract manually, their use in research and downstream risk stratification can vary across sites and teams, and may depend on subspecialty experience; scalable feature generation can help standardize access to these clinically grounded variables while keeping clinician oversight in the loop. By automating repetitive chart abstraction while producing outputs that can be audited and corrected, SNOW lowers the cost of adding clinically specific variables and accelerates iteration over feature sets. This makes it feasible to test what information in notes is truly additive beyond structured EHR data, to identify gaps in structured capture, and to shift clinician-researcher effort toward higher-value work such as refining definitions and interpreting model behavior.

The prostate cancer recurrence task studied here is a small-sample, low-event-rate setting (11/147 events; 7.5\%), which places a constraint on performance and yields high-variance estimates of discrimination. Accordingly, absolute AUC-ROC/AUPRC values should not be interpreted as evidence of an immediately deployable clinical prognostic tool. 
Instead, the prostate cohort serves as a high-resolution methodological testbed, where the high cost of manual curation was invested to create a dense `gold standard' for validating agentic reasoning. This allowed us to confirm that SNOW truly replicates expert logic before deploying it to the HFpEF cohort, where the system operated on a scale that precluded human verification. Under this controlled setting, clinician-grounded features (manual CFG and SNOW) consistently outperform scalable representational baselines, supporting the conclusion that the principal bottleneck is not model class but the ability to translate narrative documentation into clinically coherent, auditable variables.

Our prostate cancer cohort reflects treatment practices from 2005–2015; performance and feature distributions may differ under contemporary imaging, grading, and treatment protocols. While this limits claims about the transportability of the specific recurrence predictor, the external HFpEF evaluation reinforces that SNOW generalizes as a note-to-variable workflow to a distinct disease, institution, and dataset without task-specific manual curation. Further studies are warranted to evaluate SNOW prospectively and across multiple institutions for any of the indications in or study or across other diseases, ideally with standardized feature definitions and audit protocols to quantify both extraction fidelity and downstream decision-analytic utility.

Several limitations warrant emphasis. First, our prostate cancer cohort definition required a PSA value available beyond five years after treatment to support consistent outcome determination and reduce misclassification from insufficient follow-up. This design choice may introduce survivorship bias by excluding patients with shorter follow-up due to loss to follow-up or death, and may limit transportability of absolute performance estimates to settings with different follow-up completeness. Second, SNOW relies on modular agents connected by hand-designed prompts and rules, and these components are tuned independently rather than optimized end-to-end. Future work could evaluate systematic prompt optimization and joint training objectives that better align extraction fidelity with downstream research goals. This process could elucidate when the divide-and-conquer design improves reliability relative to simpler pipelines.
Third, SNOW assumes access to secure, compliant LLM infrastructure and reliable integration with EHR data systems, which may constrain adoption in some settings.
Fourth, our experiments focus on text notes and structured variables, leaving unmodeled prognostic information in modalities such as imaging, waveforms, and audio.

Fifth, our prostate recurrence analysis is intended as a methodological testbed rather than a clinically deployable risk model; accordingly, we did not benchmark predictions against established biochemical recurrence nomograms. Such comparisons would require careful harmonization of endpoints and treatment-era assumptions and complete availability of the nomograms’ input variables (including detailed treatment and pathologic factors), which were not uniformly captured in structured form in this retrospective cohort.
Future studies should extend the framework to multimodal sources and evaluate robustness and reproducibility across institutions, including explicit measurement of human oversight requirements and end-to-end time savings in real-world research pipelines, as well as benchmarking against established clinical risk tools when variables and endpoints can be standardized.

In summary, SNOW reframes the use of clinical notes in prediction as a problem of scalable, auditable measurement rather than purely representational learning. By operationalizing clinician feature generation into explicit intermediate artifacts, feature definitions, extraction rules, validation logs, and aggregation code, it enables clinically grounded variables to be recovered from narrative documentation with limited, targeted expert oversight, reducing notes-to-analysis latency and the marginal cost of adding clinically specific features. This measurement-first approach is complementary to end-to-end modeling: it supports hypothesis-driven studies, improves interpretability, and facilitates reproducible cross-team evaluation of what information in notes is truly additive beyond structured EHR data.  Looking ahead, establishing multi-institution benchmarks with standardized audit protocols, quantifying feature-level extraction uncertainty and oversight requirements, and extending the workflow to multimodal sources will be essential for translating agentic note-to-variable pipelines into robust infrastructure for clinical research.

%% file: code_availability.tex
The full implementation of the SNOW (Scalable Note-to-Outcome Workflow) framework is publicly available at
\url{https://github.com/JoyeeWang01/SNOW-Scalable-Note-to-Outcome-Workflow}.
The repository includes SQL queries for constructing the HFpEF cohort from the MIMIC-IV dataset, the complete SNOW multi-agent pipeline and prompts, a clinician-guided LLM feature generation (CLFG) implementation tailored to the prostate cancer cohort, evaluation and modeling code, and configuration files required to reproduce the experiments reported in this study.

Due to patient privacy protections and data use agreements, the underlying clinical notes and patient-level data are not shared. However, the repository provides detailed documentation and templates that enable application of SNOW to new cohorts and prediction tasks using locally available data. The code is released under the Apache License, Version 2.0.

%% file: appendix/cfg_table.tex
\begin{table}[H]
\caption{Clinical concepts and representative features.}
\centering
\renewcommand{\arraystretch}{1.2}
\begin{tabular}{p{0.28\linewidth} p{0.65\linewidth}}
\toprule
\textbf{Clinical Concept} & \textbf{Features Representative of Concept} \\
\midrule
Tumor Volume &
Percent involved and percent Gleason pattern 4/5 per systematic core; maximum
percent Gleason pattern 4/5, mean percent Gleason pattern 4/5, mean percent
involved, and percent of positive regions; clinical T stage \\
\midrule
Tumor Grade &
Percent Gleason pattern 4/5 and grade group per systematic core; maximum
percent Gleason pattern 4/5 and mean percent Gleason pattern 4/5; maximum
pre-treatment Gleason score and corresponding grade group \\
\midrule
Extraprostatic Extension & Clinical T stage \\
\midrule
Regional Metastasis & Clinical N stage \\
\bottomrule
\end{tabular}
\label{tab:clinical_concepts}
\end{table}

%% file: appendix/feature_table.tex
\subsection{Initial feature list}

\small
\begin{longtable}{@{}L{1.8cm}L{2.2cm}L{3cm}L{3.5cm}C{0.8cm}L{2cm}@{}}
\caption{Feature specifications for prostate cancer prediction model generated by \AFG.} \\
\toprule
\textbf{Feature Name} & \textbf{Specific Subgroups} & \textbf{Description} & \textbf{Instructions} & \textbf{Agg.} & \textbf{Agg. Source} \\
\midrule
\endfirsthead

\multicolumn{6}{c}{\tablename\ \thetable\ -- \textit{Continued from previous page}} \\
\toprule
\textbf{Feature Name} & \textbf{Specific Subgroups} & \textbf{Description} & \textbf{Instructions} & \textbf{Agg.} & \textbf{Agg. Source} \\
\midrule
\endhead

\midrule
\multicolumn{6}{r}{\textit{Continued on next page}} \\
\endfoot

\bottomrule
\endlastfoot

prostate volume & -- &
Volume of the prostate in cubic centimeters. Larger prostate volumes may correlate with more advanced disease and higher risk of biological failure. &
Extract the numeric value following phrases such as \texttt{Volume =} or \texttt{Vol:} in cubic cm/cc. If dimensions are given, calculate volume using $0.52 \times \text{length} \times \text{width} \times \text{height}$. &
FALSE & \makecell[tl]{--} \\
\midrule

PSA at diagnosis & -- &
Initial PSA level at diagnosis. Higher PSA at diagnosis is associated with greater risk of biological failure after treatment. &
Extract the numeric PSA value from the clinical history section, typically the most recent value before biopsy, measured in ng/mL. &
FALSE & \makecell[tl]{--} \\
\midrule

PSA velocity & -- &
Rate of PSA change over time prior to treatment. Faster PSA velocity indicates more aggressive disease and higher risk of biological failure. &
Compute change in PSA per year from sequential PSA values when available. If multiple values are present, use the linear regression slope of PSA values over time in ng/mL per year. If fewer than two PSA values are available, encode as missing (null). &
FALSE & \makecell[tl]{--} \\
\midrule

Gleason score & All 14 prostate
regions* &
Gleason score from biopsy cores by anatomical location. Higher scores, particularly patterns 4 or 5, strongly predict biological failure after treatment. &
For each location, extract the Gleason score as a numeric sum from the diagnosis section (e.g., \texttt{3+4=7} encoded as 7). For benign cores, encode as 0. &
FALSE & \makecell[tl]{--} \\
\midrule

percent core involvement & All 14 prostate
regions* &
Percentage of each biopsy core involved with cancer by anatomical location. Higher tumor volume in cores predicts higher risk of biological failure. &
For each core, extract the numeric percentage from the diagnosis section. If only a length is given (e.g., ``4 MM FOCUS IN A 13 MM CORE''), calculate percentage as $(\text{tumor length}/\text{core length}) \times 100$. For cores reported as ``$<X\%$'', encode as $X/2$. For benign cores, encode as 0\%. &
FALSE & \makecell[tl]{--} \\
\midrule

maximum Gleason primary & -- &
Predominant Gleason pattern in the highest Gleason score. Primary pattern 4 or 5 indicates more aggressive disease with higher risk of biological failure. &
Extract the first number from the highest Gleason score (e.g., from \texttt{4+3=7}, extract 4). &
FALSE & \makecell[tl]{--} \\
\midrule

maximum Gleason secondary & -- &
Secondary Gleason pattern in the highest Gleason score. Secondary pattern 4 or 5 indicates more aggressive disease and worse prognosis. &
Extract the second number from the highest Gleason score (e.g., from \texttt{4+3=7}, extract 3). &
FALSE & \makecell[tl]{--} \\
\midrule

total number of cores & -- &
Total number of biopsy cores taken. Used to quantify sampling extent and to derive percentage of positive cores. &
Count the total number of biopsy cores mentioned in the pathology report. &
FALSE & \makecell[tl]{--} \\
\midrule

number of positive cores & -- &
Number of biopsy cores positive for cancer. More positive cores correlate with more extensive disease and higher risk of biological failure. &
Count the number of cores reported as containing adenocarcinoma. &
FALSE & \makecell[tl]{--} \\
\midrule

clinical T stage & -- &
Clinical T stage of prostate cancer. Higher T stage indicates more locally advanced disease and higher risk of biological failure after treatment. &
Extract T stage from clinical notes (T1a, T1b, T1c, T2a, T2b, T2c, T3a, T3b, T4) and encode as:
T1a=1, T1b=2, T1c=3, T2a=4, T2b=5, T2c=6, T3a=7, T3b=8, T4=9. If not explicitly mentioned, derive from TNM notation (e.g., \texttt{T2bN0M0} encoded as 5). &
FALSE & \makecell[tl]{--} \\
\midrule

nodal status & -- &
Clinical N stage indicating lymph node involvement. Positive nodes (N1) greatly increase risk of biological failure. &
Extract from clinical notes as N0 or N1 and encode as: N0 = 0 (no regional lymph node metastasis); N1 = 1 (metastasis in regional lymph node(s)). &
FALSE & \makecell[tl]{--} \\
\midrule

DRE findings & -- &
Findings from digital rectal examination. Abnormal DRE findings correlate with higher risk of adverse outcomes. &
Encode as: 0 = normal/negative DRE; 1 = abnormal DRE (described as nodular, indurated, firm, asymmetric, ``positive DRE'', etc.). &
FALSE & \makecell[tl]{--} \\
\midrule

family history of prostate cancer & -- &
Presence of family history of prostate cancer, which may indicate a genetic predisposition to more aggressive disease. &
Encode as: 0 = negative family history; 1 = positive family history of prostate cancer in any relative (not limited to first-degree relatives). &
FALSE & \makecell[tl]{--} \\
\midrule

on 5$\alpha$-reductase inhibitor & -- &
Use of 5$\alpha$-reductase inhibitors (e.g., Proscar/finasteride, Avodart/dutasteride) before diagnosis. May affect PSA interpretation and indicate prior BPH treatment. &
Encode as: 0 = not taking a 5$\alpha$-reductase inhibitor; 1 = taking a 5$\alpha$-reductase inhibitor. &
FALSE & \makecell[tl]{--} \\
\midrule

treatment type & -- &
Primary treatment modality. Type of initial treatment affects risk of biological failure. &
Encode as: 1 = radical prostatectomy; 2 = radiation therapy; 3 = combined modality; 4 = other. &
FALSE & \makecell[tl]{--} \\
\midrule

androgen deprivation therapy & -- &
Use of androgen deprivation therapy (ADT). ADT use affects PSA trajectory and may delay biological failure. &
Encode as: 0 = no ADT; 1 = antiandrogen only (e.g., Casodex/bicalutamide); 2 = LHRH agonist/antagonist only (e.g., Lupron); 3 = combined androgen blockade (both antiandrogen and LHRH agonist/antagonist). &
FALSE & \makecell[tl]{--} \\
\midrule

radiation dose & -- &
Total planned radiation dose in Gray (Gy) for patients receiving radiotherapy. Higher doses may reduce risk of biological failure in higher-risk patients. &
Extract the numeric value in Gy from treatment notes for radiation patients. &
FALSE & \makecell[tl]{--} \\
\midrule

perineural invasion & -- &
Presence of perineural invasion in any biopsy core. Perineural invasion is associated with more aggressive disease and higher risk of biological failure. &
Encode as: 0 = not present/not mentioned; 1 = present in at least one core as described in the pathology report. &
FALSE & \makecell[tl]{--} \\
\midrule

PSA density & -- &
PSA level relative to prostate volume. Higher PSA density correlates with more aggressive disease and increased risk of biological failure. &
Calculate by dividing PSA (ng/mL) by prostate volume (cc). &
TRUE & PSA at diagnosis, prostate volume \\
\midrule

maximum Gleason score & -- &
Maximum Gleason score found in any core. A powerful predictor of biological failure risk. &
Determine the maximum Gleason score across all biopsy cores. &
TRUE & Gleason score (all 14 regions*) \\
\midrule

percent positive regions & -- &
Percentage of biopsy cores positive for cancer. Higher values indicate more extensive disease and increased risk of biological failure. &
Calculate as (number of cores with cancer / total cores sampled) $\times 100$. &
TRUE & total number of cores, number of positive cores\\
\midrule

maximum percent core involvement & -- &
Maximum percentage of cancer involvement in any single biopsy core. Higher maximum involvement suggests larger tumor volume and higher risk of biological failure. &
Extract the highest percentage value of cancer involvement from any core. &
TRUE & percent core involvement (all 14 regions*) \\
\midrule

bilateral disease & -- &
Presence of cancer in both lobes of the prostate. Bilateral disease indicates more extensive cancer and higher risk of biological failure. &
Encode as: 0 = unilateral (cancer confined to left \emph{or} right lobe only); 1 = bilateral (cancer present in both left \emph{and} right lobes). &
TRUE & Gleason score (all 14 regions*) \\
\midrule

surgical margin status & -- &
Status of surgical margins for patients who underwent radical prostatectomy. Positive margins increase risk of biological failure after surgery. &
Encode as: 0 = negative margins, 1 = positive margins. &
FALSE & \makecell[tl]{--} \\
\midrule
\label{tab:initial_features}
\end{longtable}

\footnotesize *14 prostate regions: left/right × (apex medial, apex lateral, mid medial, mid lateral, base medial, base lateral, anterior apex)

\subsection{Updated feature list}

\small
\begin{longtable}{@{}L{1.8cm}L{2.2cm}L{3cm}L{3.5cm}C{0.8cm}L{2cm}@{}}
\caption{Feature specifications for prostate cancer prediction model generated by \AFG.} \\
\toprule
\textbf{Feature Name} & \textbf{Specific Subgroups} & \textbf{Description} & \textbf{Instructions} & \textbf{Agg.} & \textbf{Agg. Source} \\
\midrule
\endfirsthead

\multicolumn{6}{c}{\tablename\ \thetable\ -- \textit{Continued from previous page}} \\
\toprule
\textbf{Feature Name} & \textbf{Specific Subgroups} & \textbf{Description} & \textbf{Instructions} & \textbf{Agg.} & \textbf{Agg. Source} \\
\midrule
\endhead

\midrule
\multicolumn{6}{r}{\textit{Continued on next page}} \\
\endfoot

\bottomrule
\endlastfoot

capsular invasion & -- &
Evidence of tumor invading into but not through the prostatic capsule in biopsy specimens. Can indicate more aggressive disease and higher risk of extraprostatic extension. &
Encode as: 0 = not present/not mentioned, 1 = present (mentioned in pathology report). &
FALSE & \makecell[tl]{--} \\
\midrule

immunocompromised status & -- &
Presence of immunocompromised condition (e.g., HIV, organ transplant) that could affect treatment choices and outcomes. &
Encode as: 0 = not immunocompromised/not mentioned, 1 = HIV positive, 2 = organ transplant recipient, 3 = other immunocompromised condition.&
FALSE & \makecell[tl]{--} \\
\midrule

free PSA percent & -- &
Percentage of free (unbound) PSA relative to total PSA. Lower free PSA percentage (\textless 25\%) indicates higher risk of prostate cancer and potentially more aggressive disease. &
Extract numeric percentage value when reported (typically in clinical history section).&
FALSE & \makecell[tl]{--} \\
\midrule

radiation dose & -- &
Total planned radiation dose in Gray (Gy) for patients receiving radiotherapy. Higher doses may reduce risk of biological failure in higher-risk patients. &
Extract the numeric value in Gy from treatment notes for radiation patients. &
FALSE & \makecell[tl]{--} \\
\midrule

perineural invasion & -- &
Presence of perineural invasion in any biopsy core. Perineural invasion is associated with more aggressive disease and higher risk of biological failure. &
Encode as: 0 = not present/not mentioned; 1 = present in at least one core as described in the pathology report. &
FALSE & \makecell[tl]{--} \\
\midrule

extracapsular extension & -- &
Presence of cancer extending beyond the prostate capsule based on imaging or pathology. Indicates T3a disease and significantly increases risk of biological failure. &
Encode as: 0 = not present/not mentioned; 1 = suspicious for extracapsular extension; 2 = definitive extracapsular extension noted. &
FALSE & \makecell[tl]{--} \\
\midrule

seminal vesicle invasion & -- &
Presence of cancer invading the seminal vesicles based on imaging or pathology. Indicates T3b disease and substantially increases risk of biological failure. &
Encode as: 0 = not present/not mentioned; 1 = suspicious for seminal vesicle invasion; 2 = definitive seminal vesicle invasion noted. &
FALSE & \makecell[tl]{--} \\
\midrule

hypoechoic lesion present & -- &
Presence of hypoechoic lesions on ultrasound during biopsy. These often represent areas suspicious for cancer and may indicate greater tumor burden. &
Encode as: 0 = no hypoechoic lesions mentioned; 1 = hypoechoic lesions present. &
FALSE & \makecell[tl]{--} \\
\midrule

high-grade PIN & -- &
Presence of high-grade prostatic intraepithelial neoplasia (PIN) in any biopsy core. High-grade PIN may be associated with concurrent cancer or increased risk of subsequent cancer. &
Encode as: 0 = not present; 1 = present in at least one core. &
FALSE & \makecell[tl]{--} \\
\midrule

ASAP (atypical small acinar proliferation) & -- &
Presence of atypical small acinar proliferation (ASAP) in any biopsy core. ASAP represents suspicious glands that fall short of diagnostic criteria for cancer but increase risk of subsequent cancer diagnosis. &
Encode as: 0 = not present; 1 = present in at least one core. &
FALSE & \makecell[tl]{--} \\
\midrule

age at diagnosis & -- &
Age of the patient at time of diagnosis. Age can impact treatment decisions and may correlate with outcomes. &
Extract numeric age in years from the clinical history section. &
FALSE & \makecell[tl]{--} \\
\midrule

cribriform pattern present & -- &
Presence of cribriform architecture in prostate adenocarcinoma. Cribriform pattern is associated with more aggressive disease and higher risk of biological failure, even within the same Gleason grade. &
Encode as: 0 = not present/not mentioned; 1 = present in the pathology report. &
FALSE & \makecell[tl]{--} \\
\midrule

lymphovascular invasion & -- &
Presence of cancer cells within lymphatic or blood vessels. Indicates aggressive disease with higher risk of metastasis and biological failure. &
Encode as: 0 = not present/not mentioned; 1 = present (mentioned in the pathology report). &
FALSE & \makecell[tl]{--} \\
\midrule

intraductal carcinoma & -- &
Presence of intraductal carcinoma component. Associated with more aggressive disease and higher risk of biological failure even when controlling for Gleason score. &
Encode as: 0 = not present/not mentioned; 1 = present (mentioned in the pathology report). &
FALSE & \makecell[tl]{--} \\
\midrule

history of active surveillance & -- &
History of being on an active surveillance protocol before definitive treatment. May indicate initially favorable disease or subsequent progression requiring intervention. &
Encode as: 0 = no history of active surveillance; 1 = history of active surveillance prior to treatment. &
FALSE & \makecell[tl]{--} \\
\midrule

time on active surveillance & -- &
Duration of active surveillance prior to definitive treatment. Longer duration may indicate slower disease progression. &
Extract time in months from initial diagnosis to treatment decision. For patients without active surveillance history, encode as 0. &
FALSE & \makecell[tl]{--} \\
\midrule

inflammation & -- &
Presence of inflammation in the prostate (prostatitis) in any biopsy core. Inflammation may affect PSA levels and influence management. &
Encode as: 0 = not present/not mentioned; 1 = present in at least one core (acute or chronic inflammation, prostatitis). &
FALSE & \makecell[tl]{--} \\
\midrule

Karnofsky performance status & -- &
Patient functional status on the Karnofsky Performance Scale. Lower scores correlate with poorer performance status and higher risk of complications. &
Extract numeric value (0--100) from clinical notes when available. If not explicitly mentioned, do not impute from context. &
FALSE & \makecell[tl]{--} \\
\midrule

nerve-sparing procedure & -- &
Whether nerve-sparing technique was used during radical prostatectomy. Can affect functional outcomes and may be related to margin status. &
Encode as: 0 = non–nerve sparing; 1 = unilateral nerve sparing; 2 = bilateral nerve sparing. &
FALSE & \makecell[tl]{--} \\
\midrule

prior prostate treatment & -- &
Previous treatment for prostate conditions prior to the current diagnosis/treatment. Prior interventions may affect disease progression and risk of biological failure. &
Encode as: 0 = no prior treatment; 1 = prior TURP; 2 = prior radiation; 3 = prior hormonal therapy; 4 = prior cryotherapy; 5 = multiple prior treatments. &
FALSE & \makecell[tl]{--} \\
\midrule

smoking status & -- &
Patient's smoking status at diagnosis. Smoking may affect overall health and treatment outcomes. &
Encode as: 0 = never smoker; 1 = former smoker; 2 = current smoker; $-1$ = not mentioned. &
FALSE & \makecell[tl]{--} \\
\midrule

urinary symptoms & -- &
Presence and severity of lower urinary tract symptoms at diagnosis, which may correlate with disease burden and impact quality of life. &
Encode as: 0 = no urinary symptoms reported; 1 = mild symptoms (e.g., occasional nocturia $\leq 2$ times, mild frequency); 2 = moderate–severe symptoms (nocturia $>2$ times, frequency, urgency, poor flow, hesitancy). &
FALSE & \makecell[tl]{--} \\
\midrule

PSA doubling time & -- &
Time required for PSA level to double based on sequential PSA measurements. Rapid doubling time (e.g., $< 12$ months) is strongly associated with higher risk of biological failure. &
When at least three PSA values are available, calculate PSADT using:
$\text{PSADT} = \ln(2) \times (\text{time between first and last PSA}) / [\ln(\text{last PSA}) - \ln(\text{first PSA})]$.
Report in months. If fewer than three PSA values are available, encode as missing (null). &
FALSE & \makecell[tl]{--} \\
\midrule

core length (mm) & All 14 prostate
regions* &
Total length of each biopsy core in millimeters by anatomical location. Core length provides context for interpreting cancer extent and sampling adequacy. &
Extract numeric length of the entire core in mm from the gross description section for each core location. If not explicitly provided, encode as missing (null). &
FALSE & \makecell[tl]{--} \\
\midrule

cancer length (mm) & All 14 prostate
regions* &
Length of cancer involvement in each core in millimeters by anatomical location. Longer cancer lengths indicate higher tumor volume and correlate with increased risk of biological failure. &
Extract the numeric length of cancer in mm from the pathology report for each core. If only a percentage is reported, estimate length by multiplying the percentage by the total core length when available. For benign cores, encode as 0 mm. &
FALSE & \makecell[tl]{--} \\
\midrule

tertiary Gleason pattern & -- &
Presence of a tertiary (third) Gleason pattern, particularly pattern 5, which indicates more aggressive disease beyond what the standard Gleason score captures. &
Encode as: 0 = not mentioned; 3 = tertiary pattern 3; 4 = tertiary pattern 4; 5 = tertiary pattern 5. Only code as present if explicitly mentioned in the pathology report. &
FALSE & \makecell[tl]{--} \\
\midrule

transition zone involvement & -- &
Presence of cancer in the transition zone or anterior region of the prostate. Tumors in these regions may have different biological behavior than peripheral zone tumors. &
Encode as: 0 = no transition zone involvement reported; 1 = cancer reported in transition zone, anterior region, anterior apex, or explicitly described as ``central gland''. &
FALSE & \makecell[tl]{--} \\
\midrule

prior negative biopsy & -- &
History of previous negative prostate biopsies before cancer diagnosis. May indicate slower-growing disease or prior sampling error. &
Encode as: 0 = no prior biopsies; 1 = prior negative biopsy/biopsies. &
FALSE & \makecell[tl]{--} \\
\midrule

number of previous negative biopsies & -- &
Count of previous negative prostate biopsies before cancer diagnosis. Multiple prior negatives may suggest distinct tumor biology, missed sampling, or evolving disease. &
Extract numeric count from clinical history. Code 0 if no previous biopsies mentioned, 1 for one previous negative biopsy, 2 for two previous negative biopsies, etc. &
FALSE & \makecell[tl]{--} \\
\midrule

any pattern 5 present & -- &
Presence of any Gleason pattern 5 component, indicating very aggressive disease and substantially higher risk of biological failure. &
Encode as: 0 = no pattern 5 component in any core; 1 = pattern 5 present in at least one core (as primary, secondary, or tertiary pattern). &
TRUE & maximum Gleason primary, maximum Gleason secondary, tertiary Gleason pattern   \\
\midrule

maximum cancer length (mm) & -- &
Maximum length of cancer in any single core in millimeters. Longer cancer involvement indicates higher tumor volume and correlates with increased risk of biological failure. &
Extract the maximum length of cancer reported in any core (in mm). If only percentage involvement is reported, estimate mm using core length when available. &
TRUE & cancer length (all 14 regions*) \\
\midrule

total tumor length (mm) & -- &
Total length of cancer across all positive cores in millimeters. Greater total cancer length indicates higher tumor burden and increased risk of biological failure. &
Sum the lengths in mm of cancer involvement across all positive cores. For cores where only percentage is given, estimate length by multiplying the percentage by total core length. &
TRUE & cancer length (all 14 regions*) \\
\midrule

multifocal disease & -- &
Cancer present in multiple non-adjacent areas of the prostate rather than a single focus. Multifocal disease may reflect distinct biology and impact treatment response. &
Encode as: 0 = unifocal (only one region involved); 1 = multifocal (multiple non-adjacent regions involved), based on the pattern of positive cores. &
TRUE & Gleason score (all 14 regions*) \\
\midrule

percent pattern 4 5 & -- &
Estimated percentage of tumor that consists of high-grade (pattern 4 or 5) disease. Higher percentage of high-grade disease is associated with increased risk of biological failure. &
Calculate by summing the product of each core's percent involvement and proportion of pattern 4-5 in that core, divided by the sum of all percent involvements. Encode as a percentage from 0-100. &
TRUE & Gleason score (all 14 regions*), percent core involvement (all 14 regions*) \\
\midrule

number of high risk features & -- &
Count of high-risk disease features present. Multiple high-risk features compound the risk of biological failure. &
Count presence of: 1) PSA \textgreater 20, 2) Gleason score \textgreater 8, 3) Clinical stage \textgreater T3a, 4) Positive nodes, 5) \textgreater 0\% positive cores. &
TRUE & PSA at diagnosis, clinical T stage, nodal status, percent positive regions, maximum Gleason score \\
\midrule
\label{tab:updated_features}
\end{longtable}
\footnotesize *14 prostate regions: left/right × (apex medial, apex lateral, mid medial, mid lateral, base medial, base lateral, anterior apex)

\subsection{Final feature list}\label{appendix:final-features}

The final feature set used for modeling comprises the variables presented in the \textit{Updated feature list} (Table~\ref{tab:updated_features}), with the exception of five features that were autonomously removed by the system prior to final model training: \textit{Free PSA percent}, \textit{Capsular invasion}, \textit{Immunocompromised status}, \textit{Surgical margin status}, and \textit{Number of high risk features}.

For the retained features, while the variable definitions remained consistent with the updated list, the specific extraction instructions were iteratively refined during the Extract-Validate loop. This refinement process was particularly important for complex, multi-site features, such as \textit{Core Length (mm)}, where instructions were optimized to ensure consistent extraction logic was applied across all 14 anatomical regions of the prostate.

%% file: appendix/hfpef_summary_stats.tex
\begin{table}[H]
\centering
\caption{Baseline Characteristics of Patients by Survival Status (N=2084)}
\label{tab:baseline}
\scriptsize
\setlength{\tabcolsep}{3pt} 
\resizebox{\textwidth}{!}{%
\begin{tabular}{lcccc}
\hline
 & \multicolumn{2}{c}{30-Day} & \multicolumn{2}{c}{1-Year} \\
\cline{2-3} \cline{4-5}
 & Survived & Death & Survived & Death \\
 & ($n=1993$) & ($n=91$) & ($n=1586$) & ($n=498$) \\
\hline
\textbf{Demographics} & & & & \\
Age, years (mean $\pm$ std) & 76.03 $\pm$ 13.31 & 81.82 $\pm$ 10.55 & 74.73 $\pm$ 13.46 & 81.23 $\pm$ 11.25 \\
Gender, $n$ (\%) & & & & \\
\quad Male & 1193 (59.86) & 49 (53.85) & 950 (59.90) & 292 (58.63) \\
\quad Female & 800 (40.14) & 42 (46.15) & 636 (40.10) & 206 (41.37) \\
\textbf{Vital signs (mean $\pm$ std)} & & & & \\
\quad Temperature, $^\circ$C & 36.70 $\pm$ 0.38 & 36.71 $\pm$ 0.56 & 36.72 $\pm$ 0.38 & 36.67 $\pm$ 0.44 \\
\quad Heart rate, bpm & 79.54 $\pm$ 15.62 & 83.30 $\pm$ 18.37 & 79.26 $\pm$ 15.76 & 81.21 $\pm$ 15.78 \\
\quad Oxygen saturation, \% & 96.68 $\pm$ 2.17 & 96.59 $\pm$ 1.90 & 96.70 $\pm$ 2.06 & 96.60 $\pm$ 2.42 \\
\quad Systolic BP, mmHg & 136.06 $\pm$ 21.60 & 124.59 $\pm$ 18.98 & 137.24 $\pm$ 21.93 & 129.94 $\pm$ 19.64 \\
\quad BMI, $kg/m^2$ & 32.50 $\pm$ 9.82 & 28.43 $\pm$ 6.17 & 33.19 $\pm$ 10.00 & 29.50 $\pm$ 8.16 \\
\textbf{Lab values (mean $\pm$ std)} & & & & \\
\quad Bicarbonate, mmol/L & 28.91 $\pm$ 4.37 & 29.38 $\pm$ 5.21 & 28.78 $\pm$ 4.30 & 29.43 $\pm$ 4.73 \\
\quad Creatinine, mg/dL & 1.71 $\pm$ 1.46 & 1.61 $\pm$ 0.96 & 1.69 $\pm$ 1.50 & 1.78 $\pm$ 1.25 \\
\quad Hemoglobin, g/dL & 10.74 $\pm$ 1.87 & 9.96 $\pm$ 1.56 & 10.84 $\pm$ 1.88 & 10.29 $\pm$ 1.73 \\
\quad INR & 1.73 $\pm$ 0.79 & 1.69 $\pm$ 0.74 & 1.72 $\pm$ 0.79 & 1.74 $\pm$ 0.81 \\
\quad Platelet count, $10^3/\mu$L & 237.83 $\pm$ 95.07 & 242.03 $\pm$ 101.95 & 239.31 $\pm$ 91.69 & 233.88 $\pm$ 106.15 \\
\quad Potassium, mmol/L & 4.13 $\pm$ 0.42 & 4.16 $\pm$ 0.42 & 4.13 $\pm$ 0.42 & 4.13 $\pm$ 0.42 \\
\quad WBC count, $10^3/\mu$L & 7.96 $\pm$ 5.66 & 9.42 $\pm$ 5.01 & 7.80 $\pm$ 5.12 & 8.71 $\pm$ 7.00 \\
\quad Sodium, mmol/L & 138.96 $\pm$ 3.52 & 137.98 $\pm$ 4.44 & 138.98 $\pm$ 3.51 & 138.71 $\pm$ 3.76 \\
\quad NT-proBNP, pg/mL & 5447.18 $\pm$ 8193.46 & 11036.54 $\pm$ 11462.30 & 4525.04 $\pm$ 6990.50 & 10107.48 $\pm$ 11433.67 \\
\quad Troponin, ng/mL & 0.19 $\pm$ 1.17 & 0.15 $\pm$ 0.21 & 0.17 $\pm$ 1.00 & 0.22 $\pm$ 1.39 \\
\textbf{Comorbidities, $n$ (\%)} & & & & \\
\quad Acute myocardial infarction & 258 (12.95) & 13 (14.29) & 198 (12.48) & 73 (14.66) \\
\quad Peripheral vascular disease & 215 (10.79) & 16 (17.58) & 164 (10.34) & 67 (13.45) \\
\quad Cerebrovascular disease & 122 (6.12) & 5 (5.49) & 92 (5.80) & 35 (7.03) \\
\quad Dementia & 70 (3.51) & 6 (6.59) & 51 (3.22) & 25 (5.02) \\
\quad Chronic obstructive pulmonary disease & 893 (44.81) & 41 (45.05) & 681 (42.94) & 253 (50.80) \\
\quad Rheumatoid disease & 93 (4.67) & 3 (3.30) & 71 (4.48) & 25 (5.02) \\
\quad Peptic ulcer disease & 27 (1.35) & 0 (0.00) & 22 (1.39) & 5 (1.00) \\
\quad Mild liver disease & 105 (5.27) & 6 (6.59) & 81 (5.11) & 30 (6.02) \\
\quad Diabetes & 623 (31.26) & 22 (24.18) & 506 (31.90) & 139 (27.91) \\
\quad Diabetes complications & 280 (14.05) & 16 (17.58) & 228 (14.38) & 68 (13.65) \\
\quad Hemiplegia paraplegia & 7 (0.35) & 0 (0.00) & 5 (0.32) & 2 (0.40) \\
\quad Renal disease & 881 (44.20) & 42 (46.15) & 666 (41.99) & 257 (51.61) \\
\quad Cancer & 141 (7.07) & 10 (10.99) & 91 (5.74) & 60 (12.05) \\
\quad Moderate severe liver disease & 22 (1.10) & 4 (4.40) & 15 (0.95) & 11 (2.21) \\
\quad Metastatic cancer & 41 (2.06) & 8 (8.79) & 18 (1.13) & 31 (6.22) \\
\quad Hypertension & 1627 (81.64) & 66 (72.53) & 1316 (82.98) & 377 (75.70) \\
\quad Coronary artery disease & 788 (39.54) & 34 (37.36) & 603 (38.02) & 219 (43.98) \\
\quad Pulmonary hypertension & 510 (25.59) & 17 (18.68) & 389 (24.53) & 138 (27.71) \\
\quad Atrial fibrillation & 974 (48.87) & 59 (64.84) & 731 (46.09) & 302 (60.64) \\
\hline
\end{tabular}%
}
\end{table}

%% file: appendix/rfg_comparison.tex
Across both the prostate cancer cohort and the HFpEF external validation cohort, we systematically evaluated six RFG methods (BoW Classic, BoW TF-IDF, Gemini Embeddings, and their LLM-informed variants) to understand the extent to which automated RFG approaches can extract clinically meaningful predictive signal from unstructured text.

\begin{figure}[H]
\centering
\includegraphics[width=0.9\linewidth]{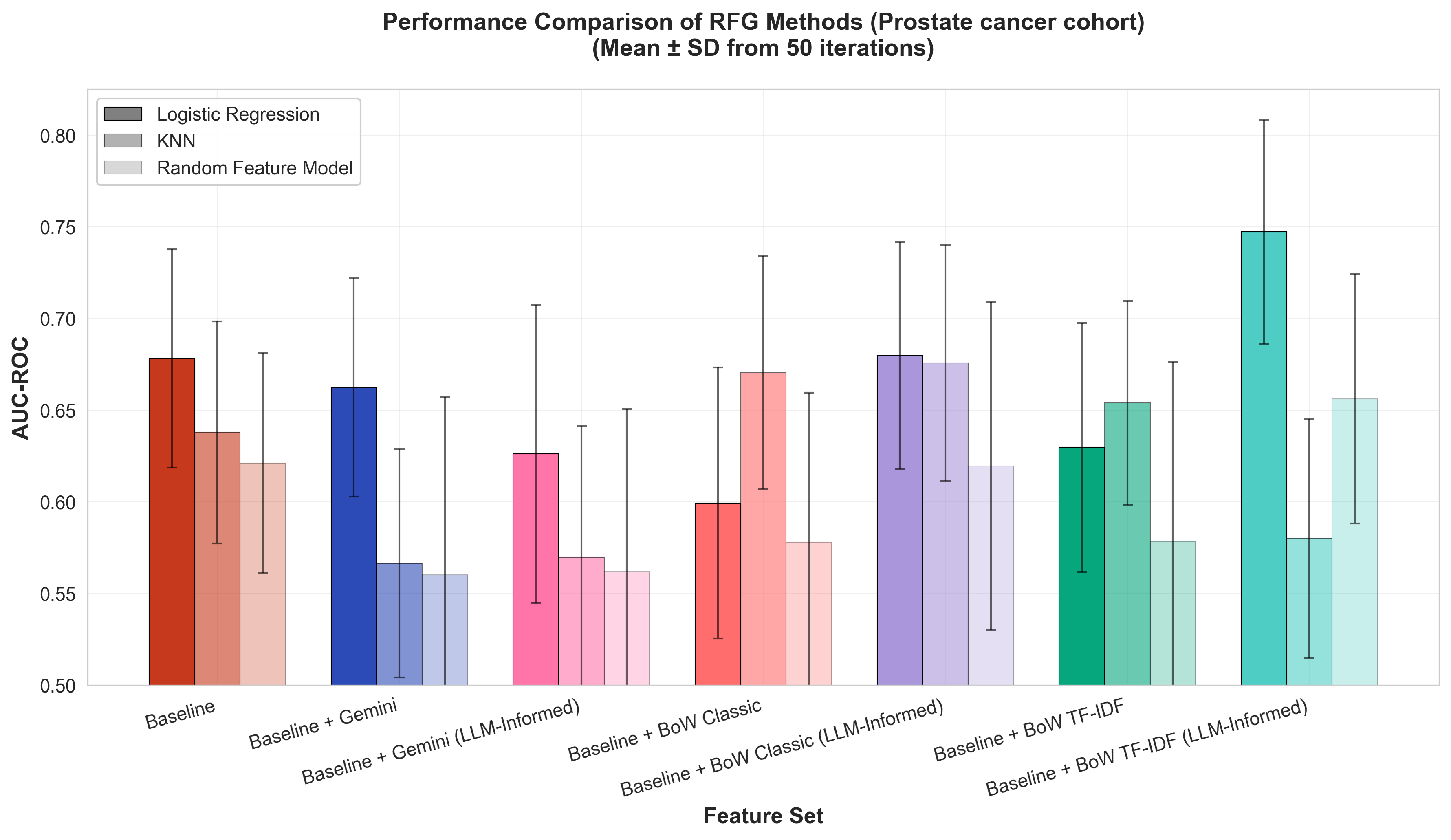}
\caption{Comparison of AUC-ROC distributions across RFG methods on the prostate cancer cohort.}
\label{fig:prostate_rfg}
\end{figure}

For the prostate cancer cohort, we observe that LLM-suggested domain-informed preprocessing consistently improves the performance of BoW methods, both BoW Classic and BoW TF-IDF, relative to standard preprocessing (Figure \ref{fig:prostate_rfg}). Because progress notes in this dataset are raw and unstandardized, removing non-medical content (e.g., URLs, email signatures) and normalizing measurement formats effectively reduced noise, thereby amplifying the signal for frequency-based models. This effect was most pronounced for BoW TF-IDF, which emerged as the best-performing RFG method. BoW TF-IDF already benefits from downweighting common n-grams; when combined with clinically grounded preprocessing, it more effectively highlights discriminative phrases associated with cancer recurrence risk, such as descriptions of tumor involvement or Gleason patterns.

In contrast, LLM-informed preprocessing produces no performance gains for the Gemini Embedding model. This is expected; as a transformer-based embedding model trained on noisy, large-scale text corpora, Gemini embeddings are designed to be robust to raw natural language. Light preprocessing may inadvertently remove contextual signals (e.g., punctuation or specific sentence structures) that such models implicitly leverage.

Overall, LLM-informed BoW TF-IDF achieves the strongest performance among all RFG methods for the prostate cancer cohort. A plausible explanation is that recurrence-associated concepts in prostate cancer (e.g., “pattern 4,” “extracapsular extension,” “core involvement”) appear in relatively fixed textual forms. Therefore, sparse term-weighted representations such as TF-IDF can make these high-value tokens especially salient after noise reduction.

A related consideration is vocabulary specificity. BoW Classic and BoW TF-IDF constructs its entire vocabulary directly from the prostate cancer notes, yielding a corpus-specific representation tuned to local documentation patterns. Given the highly idiosyncratic structure of biopsy reports, this tailored vocabulary may offer an advantage over general-purpose embedding models like Gemini Embedding model, whose training corpus spans diverse domains.

\begin{figure}[H]
\centering
\includegraphics[width=0.9\linewidth]{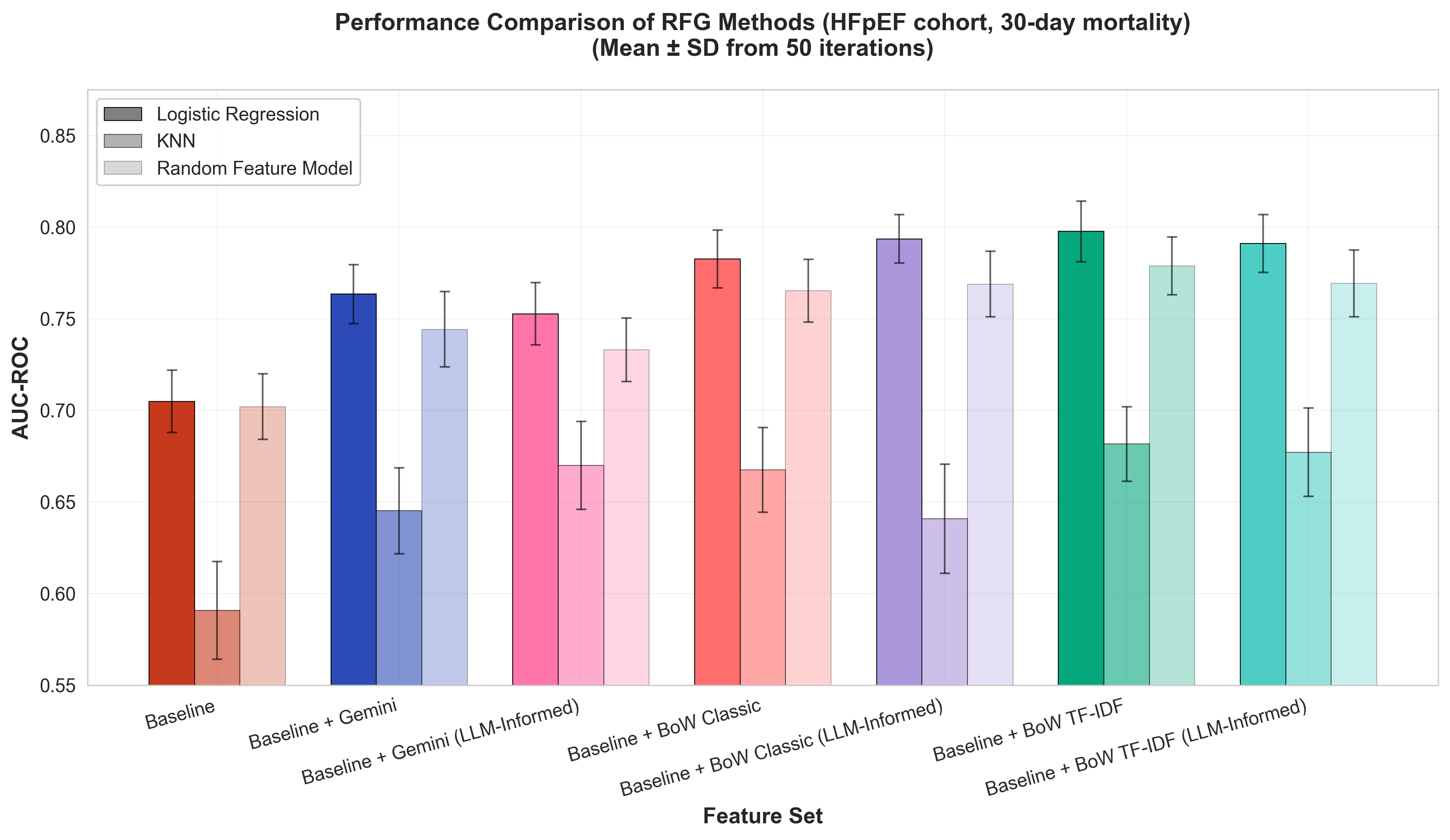}
\caption{Comparison of AUC-ROC distributions across RFG methods on the HFpEF cohort for 30-day mortality prediction.}
\label{fig:hfpef_30days_rfg}
\end{figure}

\begin{figure}[H]
\centering
\includegraphics[width=0.9\linewidth]{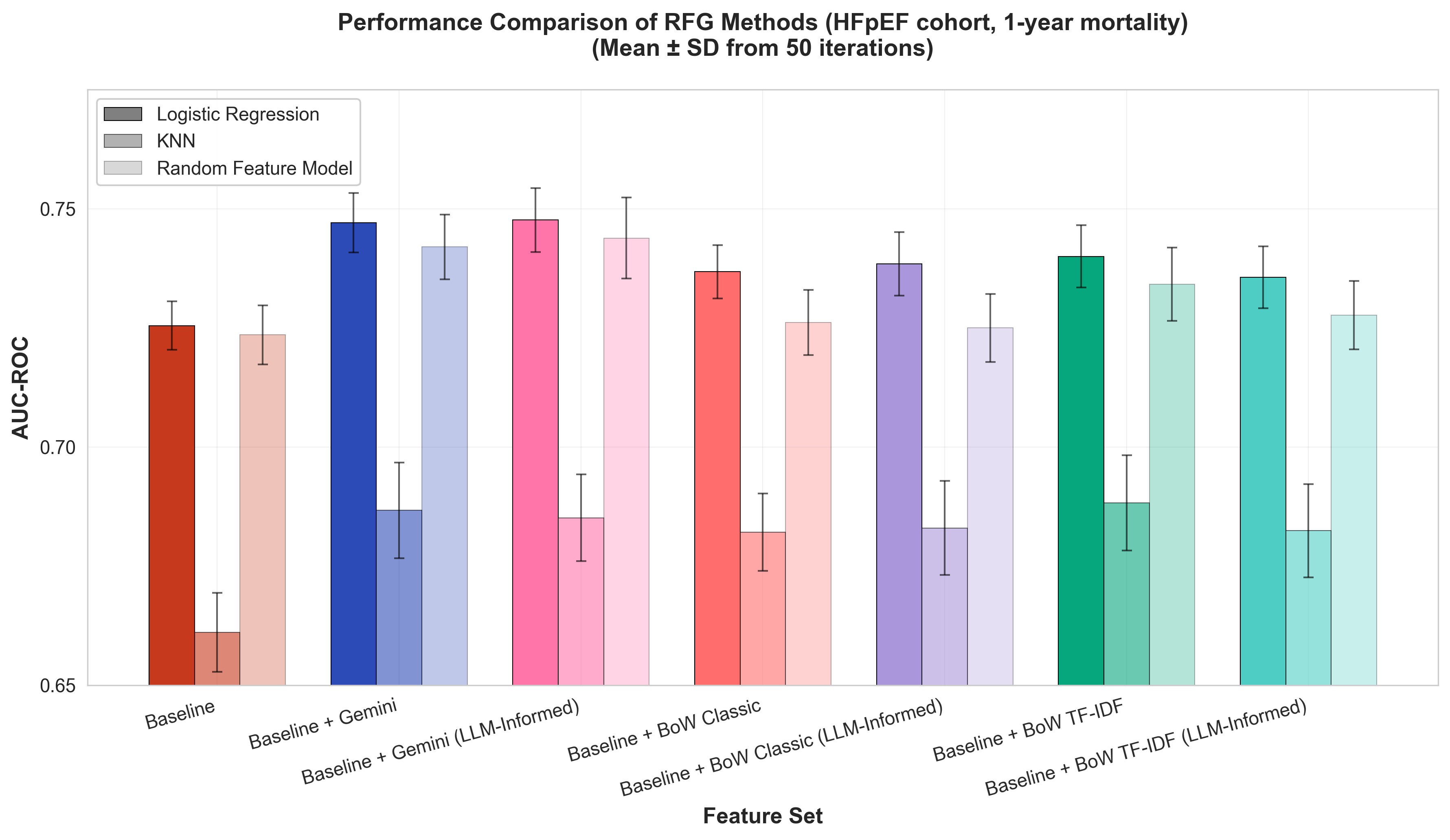}
\caption{Comparison of AUC-ROC distributions across RFG methods on the HFpEF cohort for 1-year mortality prediction.}
\label{fig:hfpef_1year_rfg}
\end{figure}

For the HFpEF cohort, the benefit of LLM-suggested preprocessing is substantially diminished (Figures \ref{fig:hfpef_30days_rfg} and \ref{fig:hfpef_1year_rfg}). Unlike the raw notes in the prostate cohort, MIMIC-IV discharge summaries have already undergone extensive de-identification and formatting normalization. As a result, additional LLM-informed preprocessing offers limited marginal noise reduction, and performance differences between standard and LLM-informed BoW features are small.

For 30-day mortality prediction, BoW TF-IDF remains the strongest RFG method, mirroring results from the prostate cohort. This may reflect that short-term mortality signals, such as acute physiological deterioration, comorbidity burden, and hospitalization complications, tend to appear in standardized textual patterns that are effectively captured by sparse lexical features.

For 1-year mortality prediction, however, all RFG methods perform only slightly above baseline, suggesting discharge summaries contain limited information about long-term outcomes in HFpEF. In this setting, Gemini embeddings achieve the highest RFG performance. Dense contextual embeddings may better encode subtle indicators of chronic disease progression or physiological frailty that are not explicitly reflected in discrete n-grams.

An additional factor shaping these results is the use of singular value decomposition (SVD) for dimensionality reduction across all RFG methods. While SVD helps mitigate overfitting in the high-dimensional feature space, the variance captured by the leading components of the BoW TF-IDF matrix appears to align well with clinically meaningful distinctions present in the notes. In contrast, Gemini embeddings are already dense, semantically rich representations, and projecting them into a lower-dimensional subspace may compress or distort subtle contextual relationships. This may explain why Gemini performs relatively worse in the small prostate cancer cohort (reduced to $\leq 147$ dimensions) but comparatively stronger in the larger HFpEF cohort (reduced to $\leq 2084$ dimensions), where the SVD projection retains more of the embedding’s intrinsic structure.

%% file: appendix/feature_analysis.tex
Given the enhanced performance of the 'Baseline + CFG' and 'Baseline + SNOW' models, we further explored which features were most frequently selected by them in predicting cancer recurrence. For each feature set and for each of 50 random seeds, we fit a logistic regression using nested cross-validation to select the L1/L2 penalty type, penalty parameter $C$, and solver that maximized AUC-ROC in each inner fold. After training each model on the full outer training set with its optimal hyperparameters, we marked a feature as ``selected'' if the absolute value of its coefficient was $\geq 0.05$. We repeated this procedure across all 50 seeds and 3 outer folds, computing the frequency with which each feature exceeded this threshold across repetitions.

These selection frequencies show a fundamental difference in how the models prioritize information. The Baseline model relies disproportionately on sociodemographic proxies, such as `marital status' (selected in 36\% of models), `language' (30\%), and `ethnicity' (27\%), likely to compensate for the absence of detailed tumor pathology. In contrast, incorporating granular features via CFG or SNOW dramatically reduces this reliance on sociodemographic factors; for instance, the selection of marital status drops to 4\% in the SNOW model. Instead, the augmented models shift their focus to direct biological measures of tumor burden. While CFG drives the model toward established metrics like `Percent involvement' and `Gleason grade', SNOW identifies additional high-signal features such as `Cancer length of any core' (selected in 46\% of models) and `PSA at diagnosis' (39\%), suggesting that the agentic workflow successfully recovers different, highly prognostic signals buried in the narrative text that are otherwise inaccessible to the Baseline model.

\begin{center}
\scriptsize
\begin{longtable}{
    l
    S[table-format=1.2]
    S[table-format=1.2]
    S[table-format=1.2]
}
\caption{Selection frequency (fraction of fitted models in which the absolute value of the feature’s coefficient is $\geq$ 0.05) for each feature in ‘Baseline’, ‘Baseline + CFG’, and ‘Baseline + SNOW’ models. A dash (--) indicates that the feature is not used in that model.}
\label{tab:feature_importance_prostate} \\
\toprule
\textbf{Feature} & \textbf{Baseline} & \textbf{Baseline + CFG} & \textbf{Baseline + SNOW} \\
\midrule
\endfirsthead

\toprule
\textbf{Feature} & \textbf{Baseline} & \textbf{Baseline + CFG} & \textbf{Baseline + SNOW} \\
\midrule
\endhead

\bottomrule
\endfoot

\multicolumn{4}{l}{\textit{Baseline Features}} \\
Maximum pre-treatment PSA                          & 0.50 & 0.20 & 0.04 \\
Charlson Comorbidity Index       & 0.43 & 0.28 & 0.17 \\
Marital status = with partner               & 0.36 & 0.12 & 0.04 \\
Language = other                        & 0.30 & 0.35 & 0.33 \\
Ethnicity = non-Hispanic        & 0.27 & 0.15 & 0.07 \\
Race = white                  & 0.17 & 0.11 & 0.04 \\
Age at treatment                         & 0.16 & 0.13 & 0.04 \\
\midrule
\multicolumn{4}{l}{\textit{CFG Features}} \\
Percent involvement of any core              & \na & 0.36 & 0.22 \\
Gleason grade group of any core                 & \na & 0.35 & \na \\
Percent Gleason pattern 4/5 of any core     & \na & 0.31 & \na \\
Clinical T stage                         & \na & 0.20 & 0.08 \\
Maximum Gleason secondary                      & \na & 0.19 & 0.05 \\
Maximum Gleason primary                        & \na & 0.15 & 0.04 \\
Clinical N stage                                      & \na & 0.10 & \na \\
Maximum percent Gleason pattern 4/5                      & \na & 0.09 & \na \\
Percent positive regions                   & \na & 0.06 & 0.02 \\
Mean percent core involvement                      & \na & 0.06 & \na \\
Maximum Gleason grade group                           & \na  & 0.03 & \na \\
Mean percent Gleason pattern 4/5                     & \na & 0.02 & \na \\
\midrule
\multicolumn{4}{l}{\textit{Additional SNOW Features}} \\
Cancer length of any core            & \na & \na & 0.46 \\
PSA at diagnosis                           & \na & \na & 0.39 \\
Gleason score of any core                 & \na & \na & 0.35 \\
Prior prostate treatment                   & \na & \na & 0.33 \\
DRE findings                               & \na & \na & 0.32 \\
PSA density                                & \na & \na & 0.21 \\
Nodal status                               & \na & \na & 0.17 \\
Length of any core                   & \na & \na & 0.16 \\
Number previous negative biopsies          & \na & \na & 0.16 \\
Anterior apex sampled                      & \na & \na & 0.15 \\
PSA doubling time                          & \na & \na & 0.14 \\
Smoking status                             & \na & \na & 0.13 \\
Maximum cancer core length            & \na & \na & 0.10 \\
On 5$\alpha$-reductase inhibitor              & \na & \na & 0.09 \\
Androgen deprivation therapy               & \na & \na & 0.09 \\
Seminal vesicle invasion                   & \na & \na & 0.09 \\
PSA velocity                               & \na & \na & 0.08 \\
Active surveillance history                & \na & \na & 0.07 \\
Atypical small acinar proliferation        & \na & \na & 0.07 \\
Hypoechoic lesion present                  & \na & \na & 0.07 \\
Multifocal disease                         & \na & \na & 0.06 \\
Maximum Gleason score in anterior apex        & \na & \na & 0.06 \\
Radiation dose                             & \na & \na & 0.06 \\
Extracapsular extension           & \na & \na & 0.05 \\
High-grade PIN                             & \na & \na & 0.05 \\
Prior negative biopsy                      & \na & \na & 0.05 \\
Extracapsular extension                    & \na & \na & 0.05 \\
Any pattern 5 present                             & \na & \na & 0.05 \\
Karnofsky performance status               & \na & \na & 0.05 \\
Treatment type                             & \na & \na & 0.04 \\
Urinary symptoms                           & \na & \na & 0.04 \\
Nerve sparing procedure                    & \na & \na & 0.03 \\
Tertiary Gleason pattern                   & \na & \na & 0.03 \\
Bilateral disease                          & \na & \na & 0.03 \\
Transition zone involvement                & \na & \na & 0.03 \\
Total tumor length                    & \na & \na & 0.03 \\
Perineural invasion                        & \na & \na & 0.03 \\
Time on active surveillance                & \na & \na & 0.03 \\
Inflammation                               & \na & \na & 0.03 \\
Anterior apex cancer present               & \na & \na & 0.03 \\
Anterior apex cancer present left          & \na & \na & 0.02 \\
Family history of prostate cancer             & \na & \na & 0.02 \\
Maximum percent core involvement               & \na & \na & 0.01 \\
Prostate volume                            & \na & \na & 0.01 \\
Age at diagnosis                               & \na & \na & 0.01 \\
Total number of cores                                & \na & \na & 0.01 \\
Number of positive cores                             & \na & \na & 0.01 \\
Maximum Gleason score                      & \na & \na & 0.01 \\
Cribriform pattern present                 & \na & \na & 0.01 \\
Intraductal carcinoma                      & \na & \na & 0.00 \\
Lymphovascular invasion                    & \na & \na & 0.00 \\
\end{longtable}
\end{center}

%% file: appendix/168_patients.tex
\begin{figure}[H]
\centering
\includegraphics[width=0.9\linewidth]{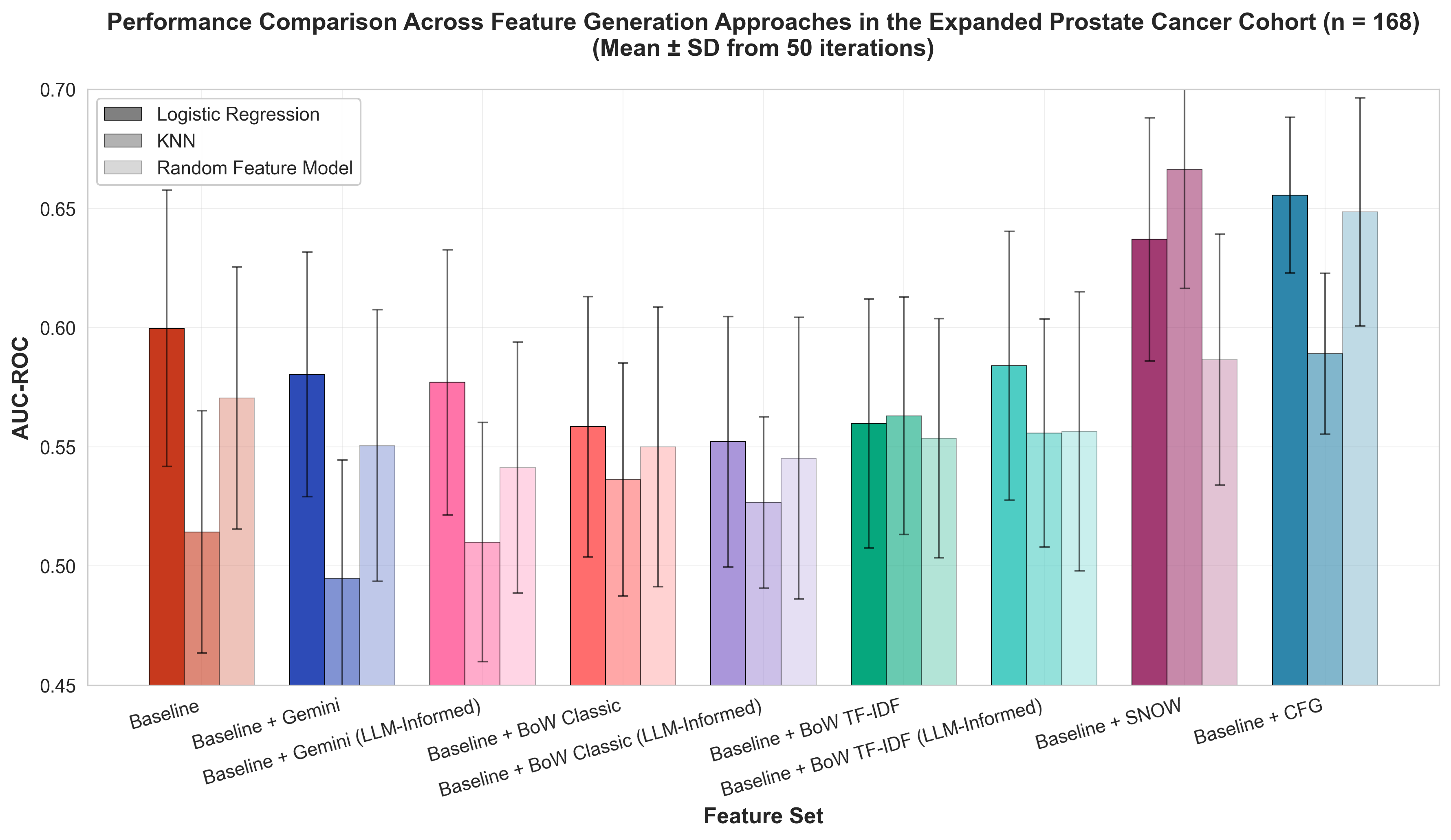}
\caption{Comparison of AUC--ROC distributions across feature generation approaches in the expanded cohort including patients receiving post-prostatectomy radiotherapy.}
\label{fig:168-main}
\end{figure}

In the primary analysis, we excluded patients who underwent both prostatectomy and radiotherapy within five years of initial treatment. However, this exclusion criterion may introduce selection bias, especially if a substantial fraction of patients with BF during the study period received salvage radiotherapy.

To assess the impact of this exclusion and evaluate the robustness of our findings, we conducted a sensitivity analysis that explicitly includes the 21 patients who underwent prostatectomy followed by radiotherapy within five years, expanding the cohort from 147 to 168 patients. For each of these patients, post-prostatectomy radiotherapy was classified as either adjuvant or salvage using clinical information available in the medical record. When PSA measurements were available between prostatectomy and initiation of radiotherapy, radiotherapy was classified as salvage if the intervening PSA value was $\geq 0.1$ ng/mL, and as adjuvant otherwise. When no intervening PSA measurement was documented, we applied a timing-based proxy: radiotherapy initiated more than six months after prostatectomy was classified as salvage, whereas radiotherapy initiated within six months was classified as adjuvant.

The resulting breakdown of adjuvant versus salvage radiotherapy among these 21 patients is reported in Table~\ref{tab:rt_breakdown}. Using this expanded cohort, we repeated the full end-to-end modeling pipeline exactly as in the primary analysis, including feature generation, preprocessing, imputation, nested cross-validation, and performance evaluation. No changes were made to feature definitions, outcome definitions, or modeling hyperparameters for this sensitivity analysis. For SNOW, the feature definition phase was skipped to ensure that the exact same feature set as in the primary analysis was used; only the extract--validate loop and feature aggregation stages were re-executed.

\begin{table}[htbp]
\centering
\caption{Breakdown of post-prostatectomy radiotherapy within five years}
\label{tab:rt_breakdown}
\begin{tabular}{lcc}
\hline
Treatment category & Number of patients & Percentage (\%) \\
\hline
Total cohort & 168 & 100.0 \\
Prostatectomy + radiotherapy within 5 years & 21 & 12.5 \\
\hspace{5mm}Adjuvant radiotherapy & 6 & 3.6 \\
\hspace{5mm}Salvage radiotherapy & 15 & 8.9 \\
\hline
\end{tabular}
\end{table}

We observed a noticeable decline in absolute predictive performance across all models in this expanded cohort compared to the primary analysis (e.g., AUC-ROC dropped by approximately 10\% for the best-performing models; Figure~\ref{fig:168-main}). This reduction likely reflects the difficulty of accurately distinguishing between adjuvant radiotherapy (part of the primary curative plan) and salvage radiotherapy (signaling recurrence) using retrospective EHR data. Our classification relied on PSA thresholds and timing proxies, which may inevitably misclassify ambiguous cases, thereby introducing label noise that degrades overall model discrimination. Crucially, however, the relative performance ordering remained consistent: SNOW and CFG continued to perform comparably to each other while consistently outperforming structured baselines and RFG approaches. This stability confirms that despite the increased label noise and cohort complexity, the agentic workflow's advantage in extracting prognostic signal persists.

%% file: appendix/alternative_BF.tex
\begin{figure}[H]
\centering
\includegraphics[width=0.9\linewidth]{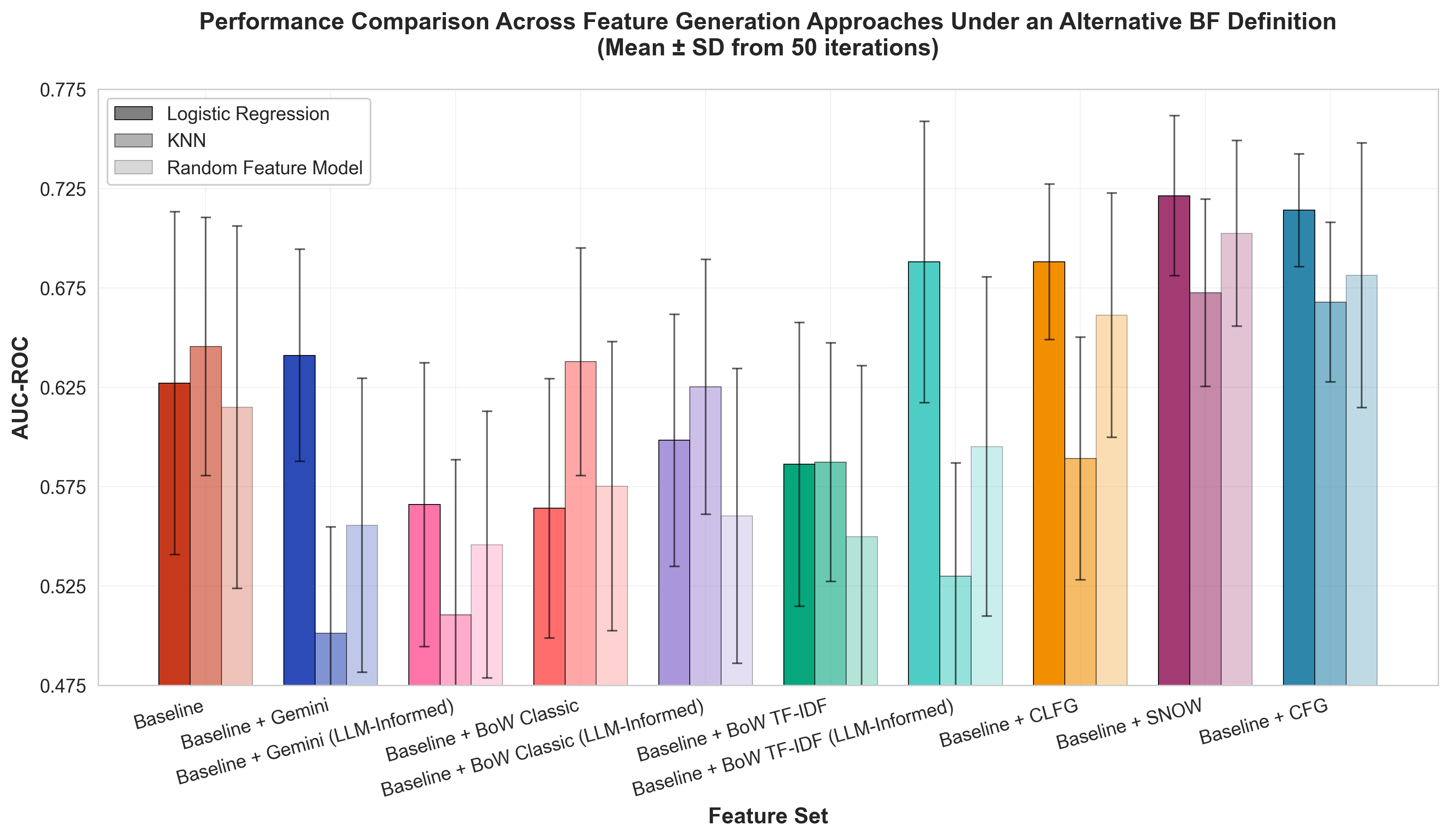}
\caption{Comparison of AUC--ROC distributions across feature generation approaches in the expanded cohort including patients receiving post-prostatectomy radiotherapy.}
\label{fig:168-main}
\end{figure}

In the main analyses, we defined BF following prostatectomy using a PSA threshold of $\geq 0.4$ ng/mL with a subsequent rising PSA, consistent with commonly used definitions in prior clinical studies. However, the optimal PSA threshold for defining post-prostatectomy biochemical failure is not uniform across the literature, and lower thresholds have been proposed to improve sensitivity for earlier recurrence detection.

To assess the robustness of our findings to this choice, we conducted a sensitivity analysis using an alternative BF definition for patients who underwent prostatectomy. Under this alternative definition, a patient was classified as having biochemical failure if they had a post-prostatectomy PSA value $\geq 0.2$ ng/mL, with the subsequent PSA measurement also $\geq 0.2$ ng/mL, indicating a confirmed elevation rather than a transient fluctuation. This definition reflects more stringent criteria that are commonly used in clinical practice and guideline discussions for early biochemical recurrence.

The BF definition for patients treated with radiation therapy was unchanged: patients were classified as having biochemical failure if their PSA level increased by $\geq 0.2$ ng/mL above the post-treatment nadir.

Using this alternative outcome definition, we regenerated BF labels for the prostatectomy subgroup and repeated the full modeling pipeline, including feature generation, nested cross-validation, and performance evaluation across all feature generation approaches (Baseline, RFG variants, CLFG, SNOW, and manual CFG). All modeling choices, hyperparameter tuning procedures, and evaluation metrics were identical to those used in the primary analyses.

We observed a decrease in absolute AUC-ROC across feature sets under the alternative definition. This decline is expected because a lower PSA threshold incorporates a higher proportion of earlier and more borderline biochemical events, which can be less separable from non-failure in retrospective cohorts due to variation in surveillance timing and low-level PSA fluctuations.

Overall, results under the 0.2 ng/mL threshold definition were qualitatively unchanged compared to the primary analyses using the 0.4 ng/mL threshold. The relative performance ordering of feature generation approaches remained stable, with clinician-informed methods (CFG, CLFG, and SNOW) consistently outperforming RFG approaches and baseline features alone. SNOW continued to achieve performance comparable to manual CFG under this alternative definition, supporting the robustness of our conclusions to reasonable variations in the definition of biochemical failure after prostatectomy.

These findings suggest that the observed performance differences between feature generation strategies are not driven by a specific PSA threshold choice for defining post-prostatectomy biochemical failure, but instead reflect systematic differences in the clinical signal captured from unstructured notes.

%% file: appendix/sociodemographic_analysis.tex
To evaluate the impact of removing sociodemographic features, we repeated the same 50 iterations of nested cross-validation described in Section \ref{sec:patient-level-methods-ml-models} for every feature generation approach, using the prostate cancer cohort and the default SVD imputation method, but excluded race, ethnicity, and language from the baseline feature set. Thus, none of the feature sets had access to these three variables. 

We find that removing race, ethnicity, and language improves the performance of the 'Baseline' and all 'Baseline + RFG' models, but has nearly no effect on 'Baseline + CLFG', 'Baseline + SNOW', or 'Baseline + CFG'. One plausible explanation is that the machine learning models operating on weaker feature sets (Baseline and RFG) tend to over-rely on sociodemographic features, which provide noise rather than clinically meaningful signal for predicting prostate cancer recurrence. In contrast, feature generation methods that incorporate clinical reasoning, i.e. CLFG, SNOW, and CFG, produce higher-quality features, reducing reliance on sociodemographic information. Importantly, the removal of race, ethnicity, and language does not qualitatively change our conclusions: 'Baseline + SNOW' and 'Baseline + CFG' remain the superior-performing feature sets.

\begin{figure}[H]
\centering
\includegraphics[width=0.9\linewidth]{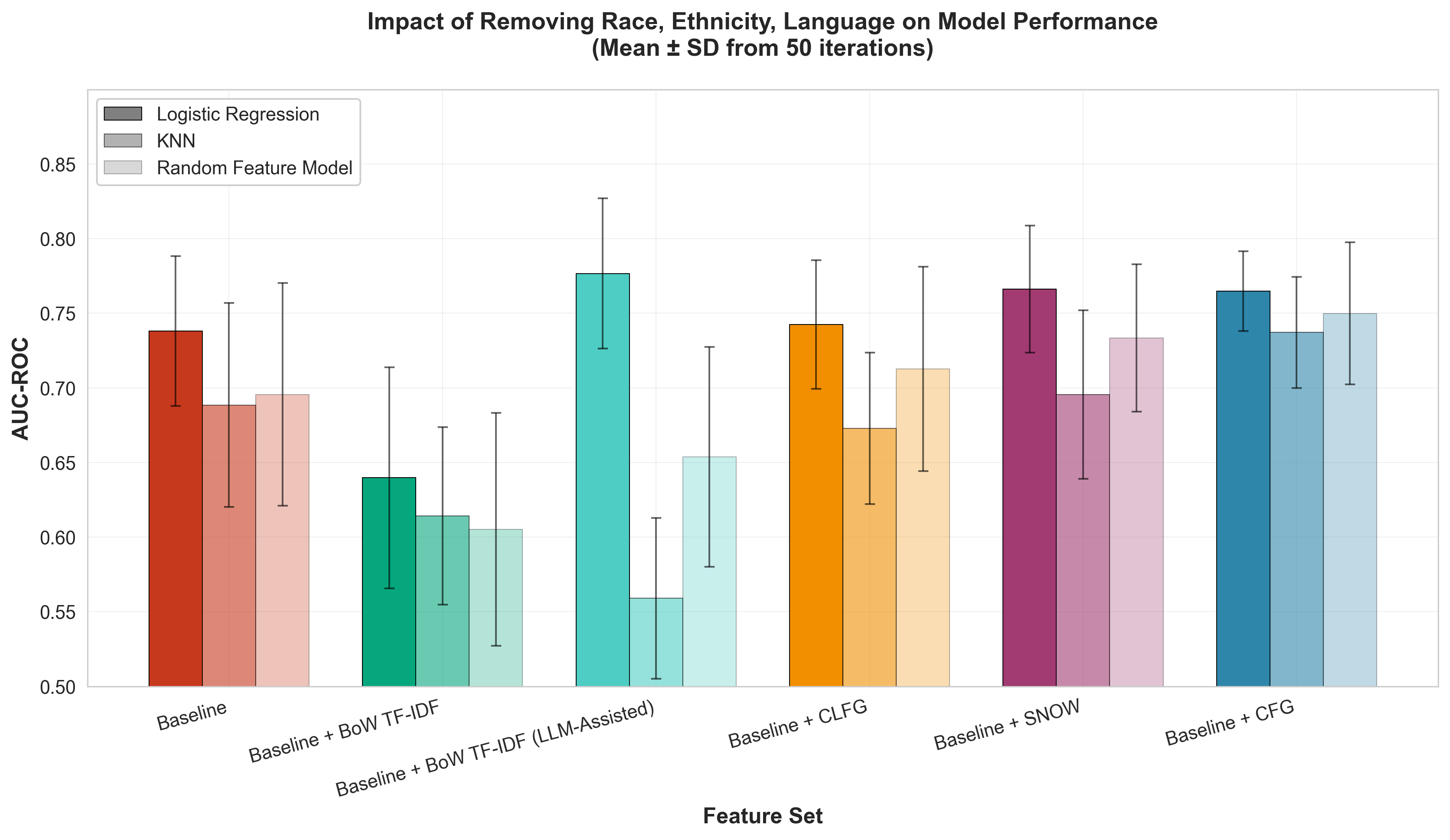}
\caption{Sensitivity analysis showing AUC-ROC distributions across feature generation approaches after excluding sociodemographic features (race, ethnicity, language) from all feature sets. Among RFG methods, BoW TF-IDF is the best-performing non–LLM assisted
variant and BoW TF-IDF (LLM assisted) is the best-performing LLM assisted RFG variant.}
\label{fig:removing_race}
\end{figure}

%% file: appendix/aupr_f1.tex
\begin{figure}[H]
\centering
\includegraphics[width=0.9\linewidth]{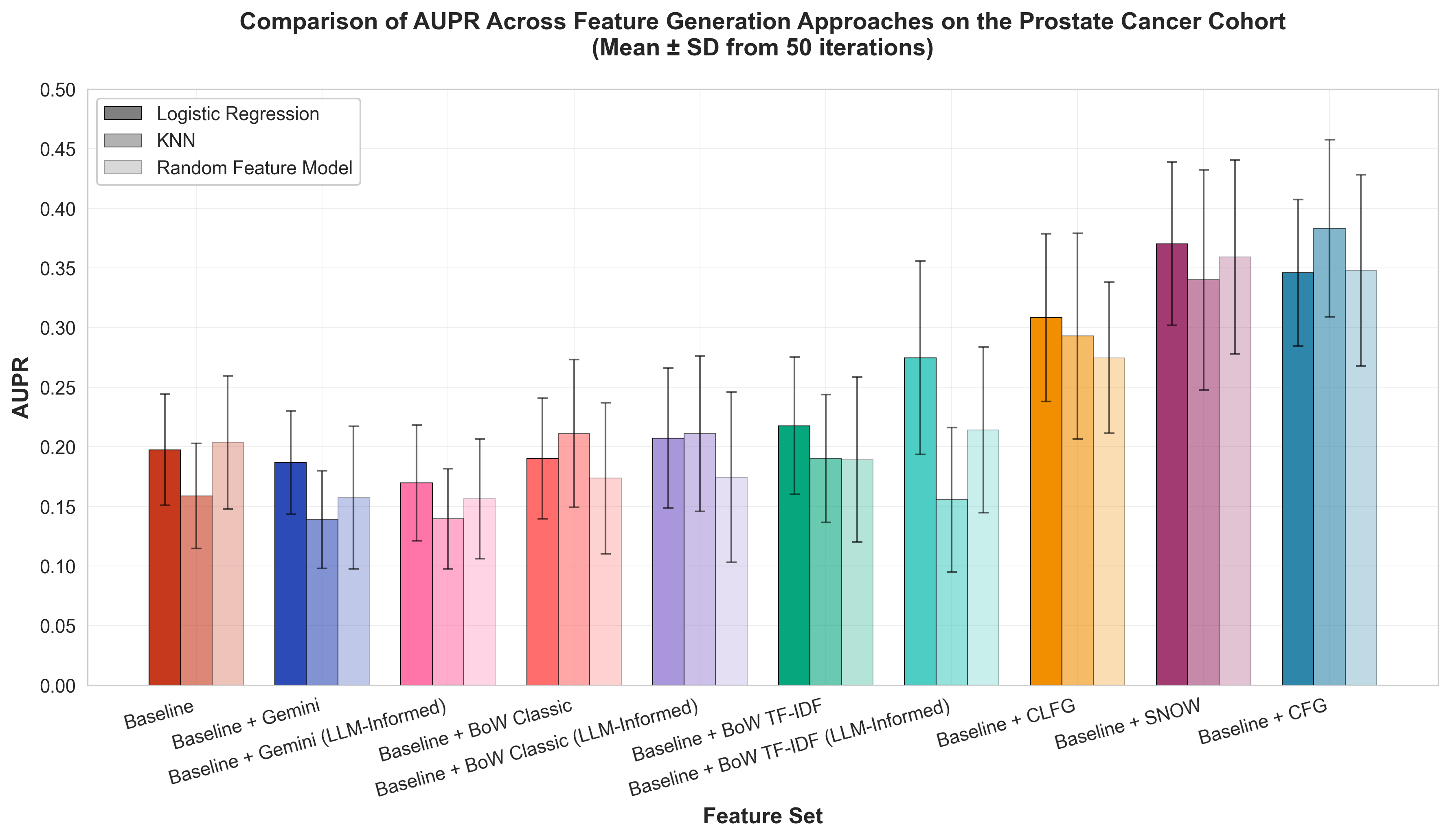}
\caption{Comparison of AUPR across feature generation approaches on the prostate cancer cohort.}
\label{fig:aupr-main}
\end{figure}

\begin{figure}[H]
\centering
\includegraphics[width=0.9\linewidth]{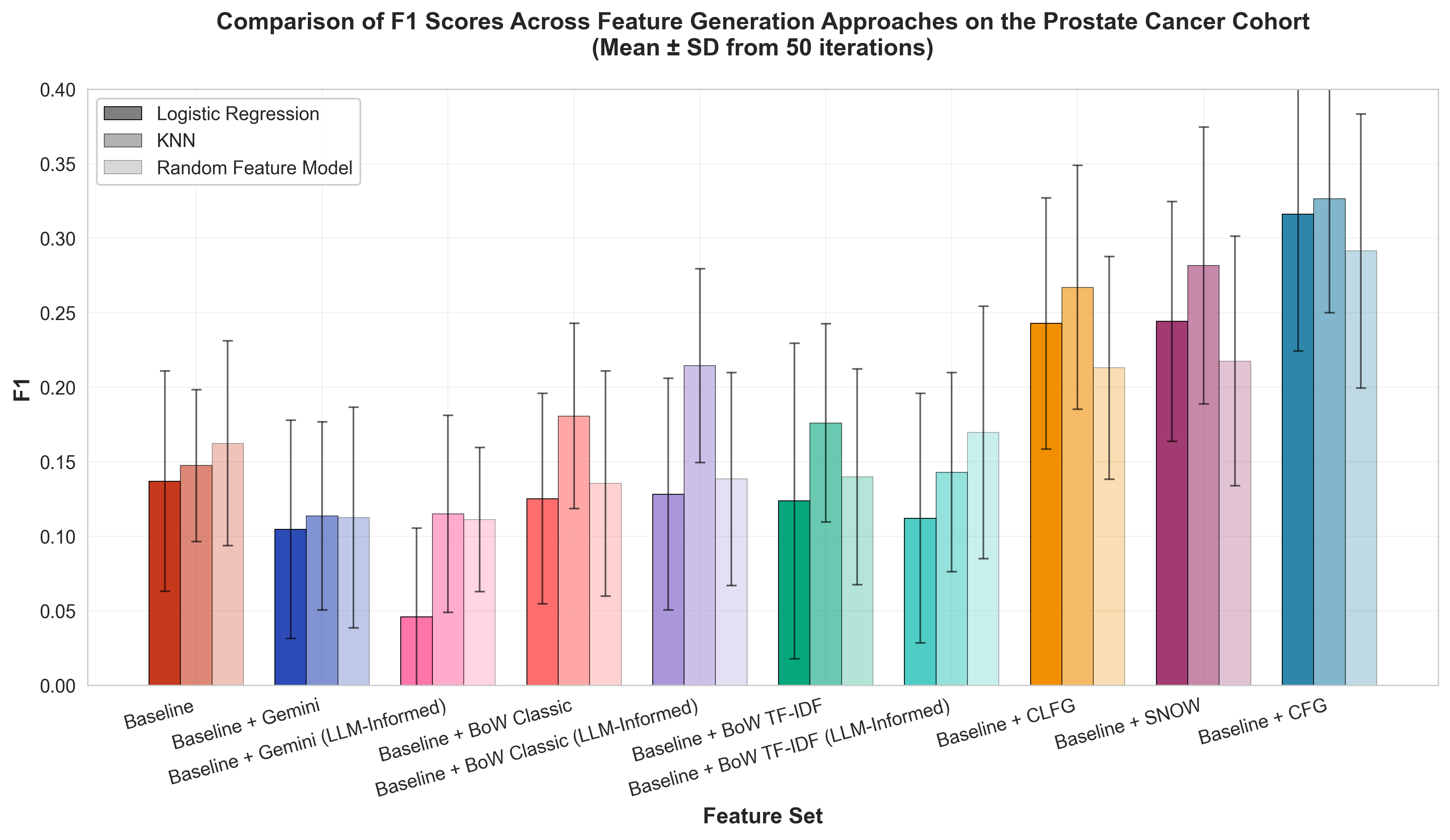}
\caption{Comparison of F1 Scores across feature generation approaches on the prostate cancer cohort.}
\label{fig:f1-main}
\end{figure}

Because BF is a rare outcome in the prostate cancer cohort, we supplement the primary AUC-ROC analysis with additional evaluation metrics that better characterize performance under class imbalance. Specifically, we report the area under the precision–recall curve (AUPR) as a ranking-based metric emphasizing the positive class and F1 score as a threshold-dependent measure of classification performance.

All metrics are computed using the same repeated nested cross-validation framework employed in the AUC-ROC analysis to ensure methodological consistency and to avoid information leakage. We perform 50 repetitions of nested cross-validation, with a 3-fold outer loop for performance evaluation and a 3-fold inner loop for model and threshold selection.

AUPR is computed on the held-out outer folds using predicted probabilities, analogous to the computation of AUC-ROC. For F1 score, the decision threshold is selected entirely within the inner cross-validation loop. For each outer fold, the threshold that maximizes the F1 score on the inner folds is chosen and then applied to the corresponding held-out outer fold. This procedure avoids optimistic bias that can arise from selecting thresholds on test data. F1 is then evaluated on the outer fold and aggregated across repetitions.

Across all models and feature generation approaches, absolute AUPR and F1 values are modest, consistent with the small sample size and low BF prevalence. Nevertheless, the relative ordering is consistent across both ranking-based (AUC-ROC, AUPR) and threshold-based (F1) metrics: adding CFG or SNOW features yields higher AUPR and F1 than baseline features alone and the RFG approaches, with CLFG intermediate. 

These results indicate that the main conclusions of the study are robust to the choice of evaluation metric.

%% file: appendix/imputation_analysis.tex
To assess whether our results were sensitive to the choice of imputation strategy, we repeated the same 50 iterations of nested cross validation described in Section \ref{sec:patient-level-methods-ml-models} for each feature generation approach, using the prostate cancer cohort and replacing the default SVD imputation with two alternatives: median imputation and multivariate imputation by chained equations (MICE). For each imputation method, we evaluated the three machine learning models (Logistic Regression, KNN, and Random Feature Model), resulting in three corresponding comparison plots. These plots allow for a direct visual assessment of how imputation affects the distribution of AUC-ROC. 

\begin{figure}[H]
\centering
\includegraphics[width=0.9\linewidth]{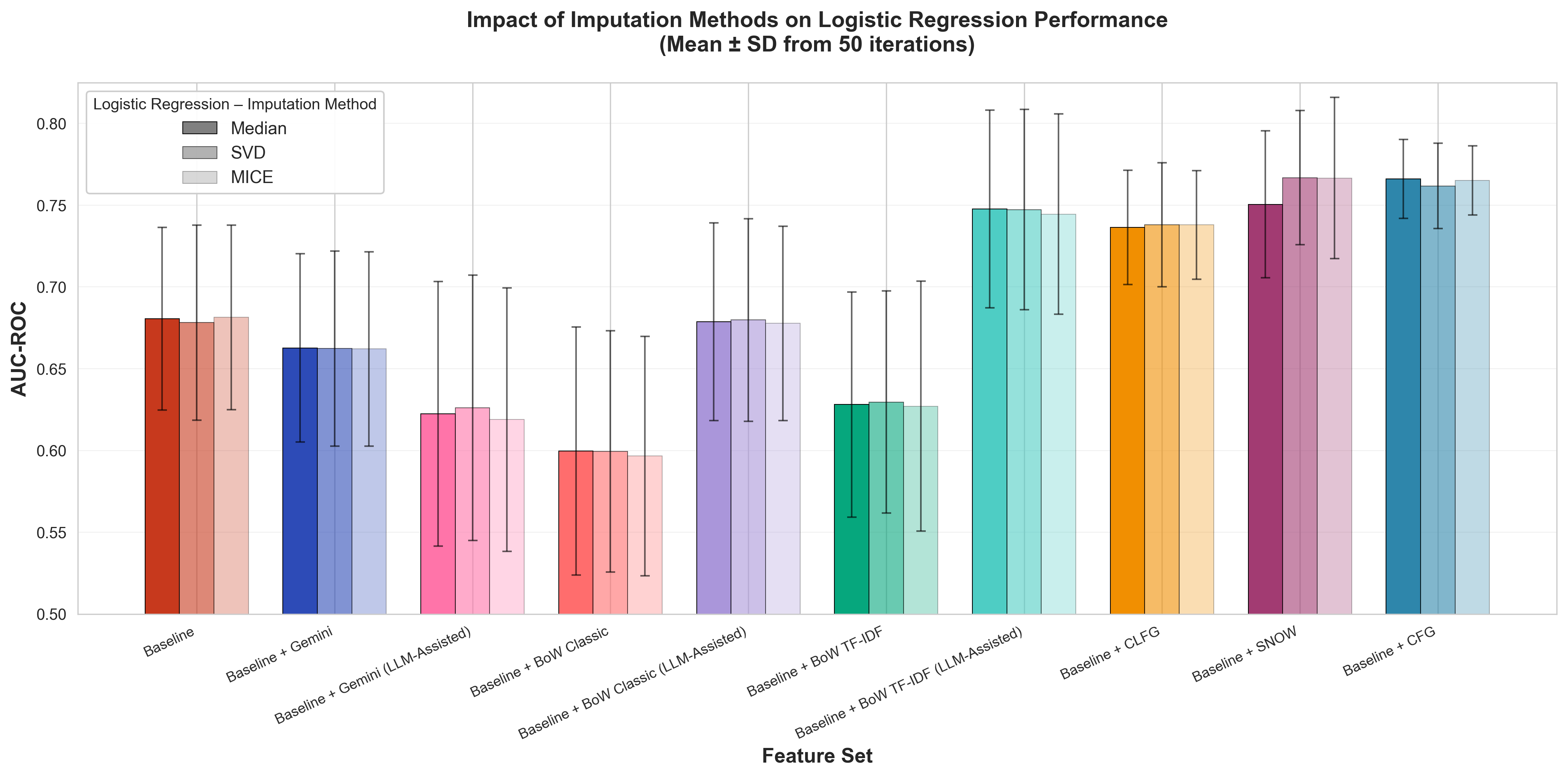}
\caption{Comparison of AUC-ROC distributions across feature generation approaches under different imputation methods using Logistic Regression on the prostate cancer cohort.}
\label{fig:imputation-lr}
\end{figure}

\begin{figure}[H]
\centering
\includegraphics[width=0.9\linewidth]{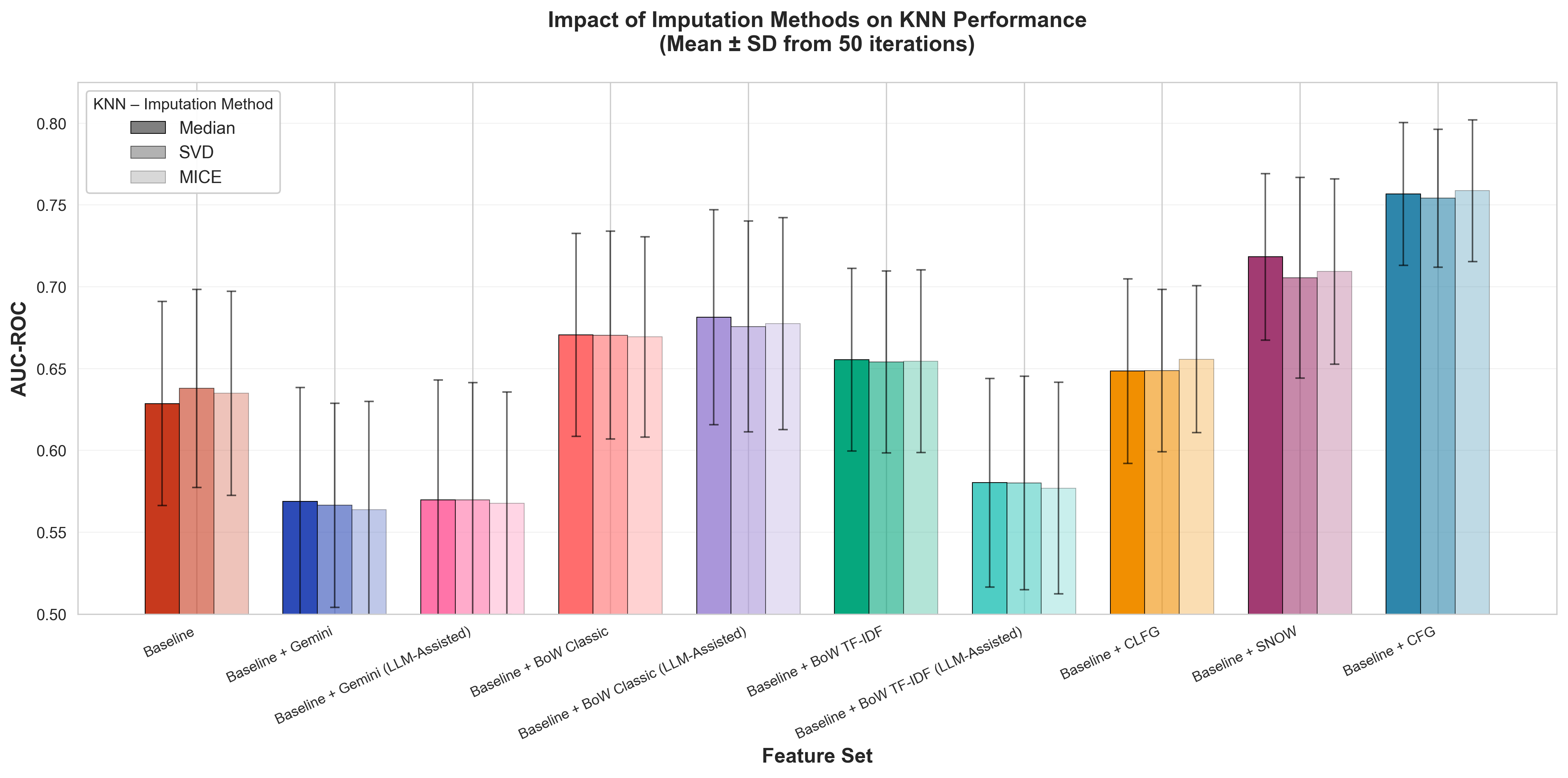}
\caption{Comparison of AUC-ROC distributions across feature generation approaches under different imputation methods using KNN on the prostate cancer cohort.}
\label{fig:imputation-knn}
\end{figure}

\begin{figure}[H]
\centering
\includegraphics[width=0.9\linewidth]{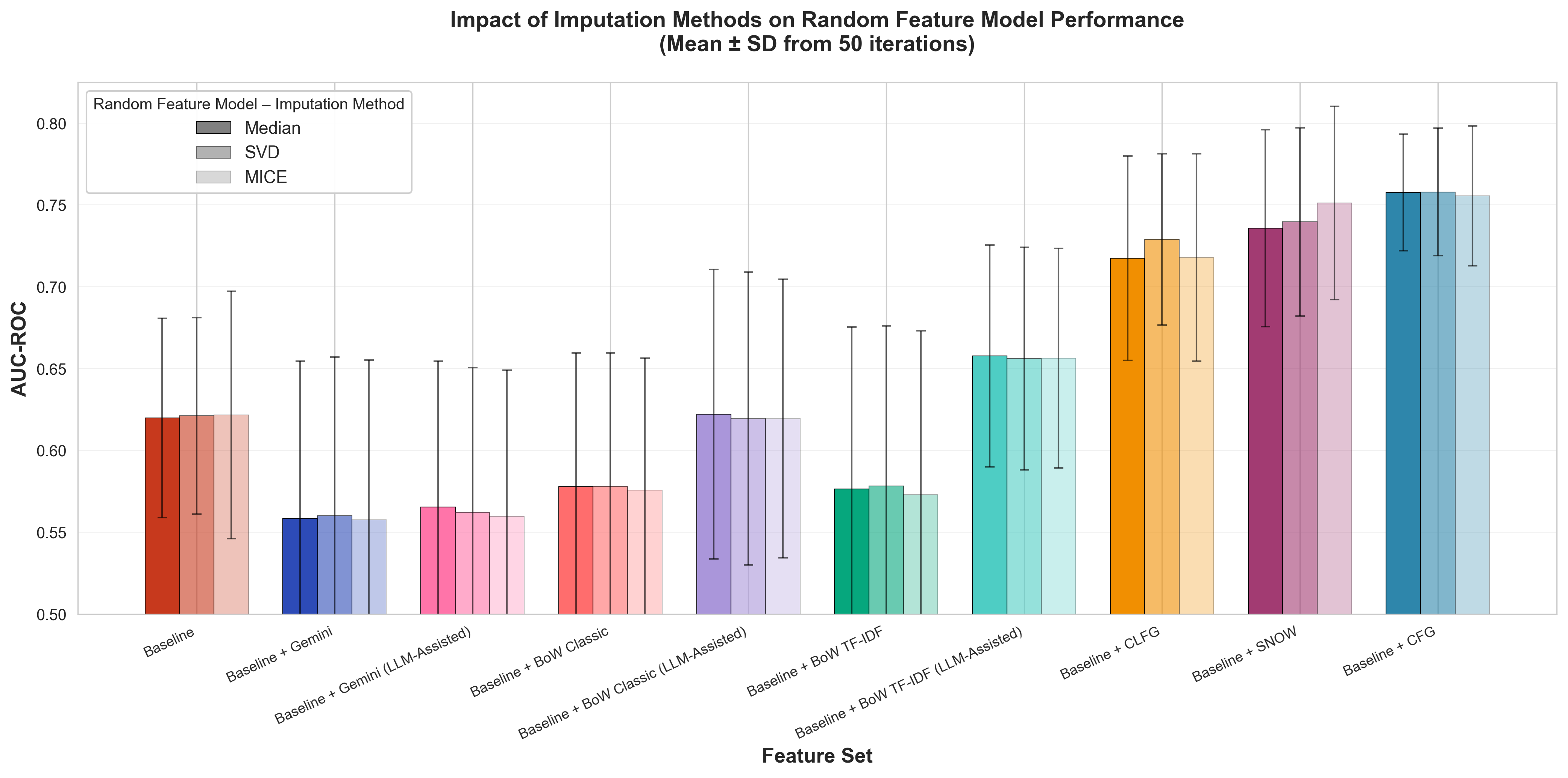}
\caption{Comparison of AUC-ROC distributions across feature generation approaches under different imputation methods using Random Feature Model on the prostate cancer cohort.}
\label{fig:imputation-rf}
\end{figure}

Across all three models, the bars representing the three imputation methods for any given feature generation approach are nearly identical in height (Figures \ref{fig:imputation-lr}, \ref{fig:imputation-knn}, \ref{fig:imputation-rf}). The largest observed difference is 0.016, between the 'Baseline + SNOW' models fit with Logistic Regression under median versus SVD imputation (Figure \ref{fig:imputation-lr}). Importantly, the relative performance ordering of feature generation approaches remains stable across all imputation strategies, indicating that our conclusions are robust to the choice of imputation method.

%% file: appendix/LLM_specs.tex
\subsection{Specifications and Configuration}

For the \textbf{prostate cancer cohort} ($n=147$), agents were provided with a task-specific context describing the input text as \emph{pathology reports of prostate cancer patients}, and the prediction target was defined as \emph{biological failure following prostate cancer treatment (e.g., radical prostatectomy or radiation therapy)}. To prevent feature duplication, the system was explicitly informed of the seven structured baseline features already available in the EHR structured data. Clinical notes were partitioned into three chunks for review by the Feature Alignment Agent. To improve computational throughput, execution of the extract--validate loop was distributed across ten parallel workers.

All agents for the prostate cancer cohort were implemented using \textbf{Anthropic Claude 3.7}. The Feature Proposal Agent and the feature list merging prompt were executed with reasoning enabled (\emph{thinking mode}), using a thinking budget of 10{,}000 tokens. All remaining agents in the pipeline operated with reasoning disabled to reduce latency and cost.

For the \textbf{HFpEF external validation cohort} ($n=2{,}084$), input texts were described as \emph{discharge summaries documenting the reason for admission, hospital course, and discharge plans}, and the prediction target was defined as \emph{death within 1 year/30 days of hospital discharge}. Given the larger corpus size, notes were partitioned into ten chunks for the Feature Alignment Agent. As with the prostate cancer cohort, the extract--validate loop was executed using ten parallel workers.

For the HFpEF cohort, the system relied on \textbf{Google’s Gemini 2.5 model family}. The Feature Proposal Agent and the feature list merging prompt were executed using \textbf{Gemini 2.5 Pro} with reasoning capabilities enabled. All other agents employed the lightweight \textbf{Gemini 2.5 Flash} model.

Across both cohorts, all models were run using default temperature settings.

\subsection{Prompts}

This section contains the full system prompts used by SNOW. 

\subsection*{Feature Proposal Agent}
\textit{This agent analyzes raw clinical notes to identify potential structured features for prognosis modeling.}

\begin{Prompt}
You are a clinical machine learning scientist developing a prognosis model.
The clinical notes you will receive are: \{notes_description\}
Your prediction target is: \{outcome_description\}

Identify the most clinically relevant structured features—either categorical or 
numerical—commonly found in such reports.
You have access to each clinical note (indexed 0 to \{MAX_NOTE_INDEX\}) via the 
get_note tool. You are expected to identify patterns, structures, and field 
repetitions (e.g., per-region or per-sample reporting) by reading and analyzing 
these notes directly.
Please read at least 6 notes.

Do not include the following features, as they are already available:
\{structured_feature_names\}

Important context:
Clinical notes may contain multiple anatomical regions, repeated measurements, or 
temporal sequences of observations. It is your responsibility to identify such 
structures from the notes and propose features that are specific to these 
subgroups or aggregated across them.

Feature requirements:
  1.  Each feature must be directly usable in a regression model: clearly defined 
      and numerically or categorically encoded.
  2.  Features can be:
  • Specific_subgroups: If the notes report values per anatomical region, time 
    point, or other subgroup, identify all distinct subgroups and propose one 
    feature per subgroup.
  • Aggregated: If multiple features are summarized using functions like mean, 
    max, or count, clearly indicate this.

If a feature varies across subgroups (e.g., different values per region or 
timepoint), then:
1. Return only one feature object, with the specific_subgroups field listing all 
   relevant subgroups.
2. We would like to extract the feature for each subgroup separately.

If a feature is aggregated from multiple fields, for example, *max*, *mean*, 
*count*, *sum*, you must:
1. Set "is_aggregated": True
2. In "aggregated_from", include a list of JSON objects, each containing:
   - "feature_name": the base feature name
   - (if applicable) "specific_subgroups": a list of the associated subgroup names
3. **Make sure that each of those base features is also included as a feature 
   object in the JSON output**

If the feature is not aggregated, set "is_aggregated": False and leave out 
"aggregated_from".

For each feature, include:
1. Feature Name
2. Description and clinical rationale for inclusion
3. A sample extraction rule as instructions (if the feature is categorical, 
   explicitly list all possible categories and assign a numeric code for each 
   category)
4. Whether the feature is aggregated ('True' or 'False')
5. If aggregated, which features it was aggregated from

Return your output as a JSON array of objects, where each object has the 
following keys:
- "feature_name"
- *(optional)* "specific_subgroups" (a list of subgroup names if values vary 
  across subgroups)
- "description" (include rationale for why this feature relates to the 
  prediction target)
- "instructions" (if the feature is categorical, explicitly list all possible 
  categories and their numeric codes)
- "is_aggregated" (boolean)
- *(optional)* "aggregated_from" (JSON object array, required only if 
  is_aggregated is True)

Return the JSON array between <JSON> and </JSON> tags.
\end{Prompt}

\subsection*{Feature Alignment Agent}
\textit{This agent validates the initial feature set against the full corpus of notes to identify new features based on available information.}

\begin{Prompt}
You are a clinical machine learning scientist performing **feature validation**.  
You previously generated a set of candidate features from clinical notes.  
The clinical notes you will receive are: \{notes_description\}  
Your prediction target is: \{outcome_description\}
Now, you must carefully review each feature recommendation against 10 of the 
actual notes given.

Each feature has the following fields:
- "feature_name" 
- *(optional)* "specific_subgroups" (a list of subgroup names if values vary 
  across subgroups)
- "description" (include rationale for why this feature relates to the 
  prediction target)
- "instructions" (if the feature is categorical, explicitly list all possible 
  categories and their numeric codes)
- "is_aggregated" (boolean)
- *(optional)* "aggregated_from" (JSON object array, required only if 
  is_aggregated is True)

Your tasks:  
1. **Confirm relevance**: For each feature, check whether the concept is 
   consistently present and clinically meaningful in these notes. Mark features 
   that are too rare, ambiguous, redundant with existing structured features, 
   or unlikely to be extractable.  
2. **Suggest edits**: If a feature is useful but poorly defined, suggest better 
   subgroup definitions, sharper extraction rules, or (for categorical features) 
   a better list of categories with explicit numeric codes.  
3. **Identify gaps**: While reviewing the notes, look for clinically important 
   elements that were not included in the current candidate feature list. 
   Propose additional features following the same schema.  
4. **Aggregation consistency**: For aggregated features, confirm that the 
   underlying base features exist, are valid, and are not double-counted.  

Feature requirements:
  1. Each feature must be directly usable in a regression model: clearly defined 
     and numerically or categorically encoded.
  2. Features can have:
  • Specific_subgroups: If the notes report values per anatomical region, time 
    point, or other subgroup, identify all distinct subgroups and propose one 
    feature per subgroup.
  • Aggregated: If multiple features are summarized using functions like mean, 
    max, or count, clearly indicate this.

If a feature varies across subgroups (e.g., different values per region or 
timepoint), then:
1. Return only one feature object, with the specific_subgroups field listing all 
   relevant subgroups.
2. We would like to extract the feature for each subgroup separately.

If a feature is aggregated from multiple fields, for example, *max*, *mean*, 
*count*, *sum*, you must:
1. Set "is_aggregated": True
2. In "aggregated_from", include a list of JSON objects, each containing:
   - "feature_name": the base feature name
   - (if applicable) "specific_subgroups": a list of the associated subgroup 
     names
3. **Make sure that each of those base features is also included as a feature 
   object in the JSON output**

If the feature is not aggregated, set "is_aggregated": False and leave out 
"aggregated_from".

### Final output rules:  
Return an updated JSON array of feature objects between `<JSON>` and `</JSON>` 
tags, where:  
- **New features ("status": "new")** → Return the **full feature object** with 
  all required fields.
- **Edited features ("status": "edited")** → Return only the **updated fields** (not the full object) along with "feature_name" and "status".  
- **Confirmed features ("status": "confirmed")** → Return only "feature_name" 
  and "status".  
- **Dropped features ("status": "drop")** → Return only "feature_name", 
  "status", and a short "rationale".

Features that you recommend for **dropping must still be included in the output 
with "status": "drop" and a rationale. These drops are not final—they may be 
reversed later if evidence from other notes supports keeping the feature.

### Constraints:  
- Do not include the following features, as they are already available:  
\{structured_feature_names\}  
- Ensure each feature is well-defined for direct use in a regression model.  
- All feature values must be strictly numerical (integers or floats) with no 
  text, units, or symbols.

Note #1: \{note_1\}
...
Note #10: \{note_10\}

Candidate features:
\{features\}
\end{Prompt}

\subsection*{Prompt for Merging Feature Lists}
\textit{This prompt consolidates feature lists from multiple parallel feature alignment runs.}

\begin{Prompt}
You are a clinical machine learning scientist performing **feature validation**.
You previously generated \{num_chunks\} sets of candidate features from reviewing 
\{num_chunks\} subsets of clinical notes.
The clinical notes you received were: \{notes_description\}
Your prediction target is: \{outcome_description\}

### Your task:
1. Take the **union** of the \{num_chunks\} sets of candidate features and return a 
   **single consolidated set** of features.
2. If multiple features are very similar or redundant, keep only one of them 
   (the clearest and most clinically relevant).

### Final output:
Return an updated JSON array of feature objects between `<JSON>` and `</JSON>` 
tags, where:
Each feature must include:
- "feature_name"
- *(optional)* "specific_subgroups"
- "description"
- "instructions"
- "is_aggregated"
- *(optional)* "aggregated_from"

### Inputs:
\{feature_sets\}
\end{Prompt}

\subsection*{Feature Extraction Agent}
\textit{This agent performs the extraction of specific feature values from individual notes.}

\begin{Prompt}
You are tasked with extracting specific feature from a patient's clinical note. 
The clinical notes you will receive are \{notes_description\}.
The note is provided below:

<clinical_note>
\{note\}
</clinical_note>

Your task is to extract the following features from the clinical note:

<features_detail>
\{features_detail\}
</features_detail>

1. Consider the different ways each feature might be described or implied in 
   clinical language.
2. Apply different strategies to identify, infer, or calculate each feature 
   based on the available information in the note.
3. If a feature is truly not found or not extractable, set "value" to null.
4. Make sure that all extracted values are returned strictly as numerical values 
   (integers or floats) without any text, units, or symbols.

After your analysis, provide the extracted data in JSON format. The JSON object 
should contain the following keys:
\{feature_list\}

Return the JSON object between <JSON> and </JSON> tags.
\end{Prompt}

\subsection*{Feature Validation Agent}
\textit{This agent audits the extraction quality and decides whether to proceed, remove, or re-extract.}

\begin{Prompt}
You are a senior clinical data scientist. Your job is to review a feature that 
was extracted from clinical notes and decide whether to:
1. proceed - if the feature is consistently and correctly extracted.
2. remove - if the feature cannot be reliably extracted due to too many absences 
   from the notes.
3. reextract - if the feature could be consistently extracted but extraction 
   logic needs revision or the values require post-processing (e.g., formatting, 
   grouping, normalization).

The clinical notes you will receive are: \{notes_description\}
Your prediction target is: \{outcome_description\}

The current feature is:
<feature_detail>
\{feature_detail\}
</feature_detail>

The **extracted values**, keyed by clinical note index (0–\{MAX_NOTE_INDEX\}):
<extracted_values>
\{extracted_values\}
</extracted_values>

The extracted values will be used in a regression model. Please analyze the 
consistency, accuracy, and usefulness of this feature extraction. You have 
access to each of the clinical notes (indexed from 0 to \{MAX_NOTE_INDEX\}) via 
the 'get_note' tool. Use it to validate extracted values against the original 
notes as needed.

For your reference, this feature has been extracted \{extraction_count\} times and 
validated \{validation_count\} times.
Features that are already included in the model are: \{all_feature_names\}

For missing values:
1. \{missing_percent\} percent of the values are currently missing.
2. Validate against at least 4 of the original notes to see whether these are 
   true absences or extraction mistakes.
3. Apply different strategies to identify, infer, or calculate the feature based 
   on the available information in the notes.
4. If a new strategy works, update the instructions and re-extract the feature.
5. Try reduce the missing values as much as possible through better extraction.
6. If the missing values are due to true absences and the missing rate remains 
   too high to support meaningful analysis, remove the feature.

Special Instructions for Categorical Features:

If the feature is **categorical**, additionally evaluate:
- Are the **categories consistent** in terminology, format, and meaning?
- Do the **categories make clinical and modeling sense**?
- Are there semantically redundant or overlapping categories?
- Is the **number of categories appropriate** (e.g., not overly granular)?

If the current categories are unclear or inconsistent:
- Define a clear set of **numerical categories**
- Recommend reextract if the extraction logic must be updated to match these 
  categories or the extracted values are correct, but can be transformed or 
  mapped into numerical categories

### Additional Features/Columns/Indicators:
If you believe new features (e.g. binary indicator or count) should be derived 
from or replace the current feature, propose them in a JSON array. Each object 
should include:
- "feature_name"
- *(optional)* "specific_subgroups"
- "description"
- "instructions"

New features will be reextracted from the notes. If the newly suggested features 
can effectively replace the current feature, your final decision must be 
**remove**, meaning removing the current feature and keeping the new features.

### Formatting Requirement:
Make sure that all extracted values are returned strictly as **numerical values 
(integers or floats) without any text, units, or symbols**. If a value cannot be 
determined, return `null`.

### Final Output:
Return your decision as a JSON object with this structure:
\{\{
    "decision": proceed|remove|reextract,
    "add_additional_feature": null or JSON array of the suggested additional 
    features
    "reasoning": Brief explanation of your decision,
    "current_feature_instructions": If decision is "reextract", provide 
    concrete guidance for improving category mapping or extraction logic. 
    Reference specific input examples where possible. **IMPORTANT**: DON'T 
    mention additional features/columns/indicators here.
\}\}

Return the JSON object between <JSON> and </JSON> tags.
\end{Prompt}

\subsection*{Aggregation Code Generator}
\textit{This agent generates Python code to handle complex aggregations (e.g., mean, max) across extracted values.}

\begin{Prompt}
You are an expert Python data scientist.
Below is ONE aggregated clinical feature:

<aggregated_feature>
\{aggregated_feature\}
</aggregated_feature>

Write a pure-Python function:

def aggregate_<feature_name>(features: dict[str, Any]) -> Any:
    """Return the aggregated value or None."""

Rules
------
* "features" contains the base feature keys listed in "aggregated_from".
* Treat None as missing.
* If the aggregation is an average/mean/min/max/sum, ignore None values in the 
  mean/min/max/sum. Return None if *all* constituents are None.
* Use only the Python stdlib and numpy as np.
* If unable to write the function given the provided base features, return 
  <ERROR> followed by a string explaining why.

Below are descriptions of the base feature keys listed in "aggregated_from":
\{aggregated_from_features\}

ONLY output valid Python code — no markdown or extra text.
\end{Prompt}

%% file: appendix/SNOW_examples.tex
The following tables detail the step-by-step validation process for three specific features. Each iteration includes a review of specific notes, the reasoning applied by the validation agent, and the subsequent decisions.

\subsection*{Feature: \texttt{surgical\_margin\_status}}
\textbf{Definition:} Status of surgical margins for patients who underwent radical prostatectomy (0 = negative, 1 = positive). \\
\textbf{Final Outcome:} Feature removed.

\subsubsection*{Iteration 1}
\textbf{Dataset-wide Result:} All values missing (null).

\begin{longtable}{p{0.08\textwidth} p{0.12\textwidth} p{0.35\textwidth} p{0.35\textwidth}}
\toprule
\textbf{Index} & \textbf{Extracted Value} & \textbf{Content Summary} & \textbf{Validation Reasoning} \\
\midrule
\endhead

0 & nan & Biopsy pathology and RRP operative report; no post-op pathology text found. & Margin status requires final pathology report. Null is expected given text. \\
25 & nan & Biopsy pathology (aggressive); radiation planning note. & No prostatectomy pathology; margins not applicable. Null appropriate. \\
50 & nan & Biopsy pathology; radiation treatment progress note. & No surgery specimen report; margins absent. Null appropriate. \\
100 & nan & Biopsy pathology; pre-op clinic note planning RRP. & Margin status absent (would be in later report). Null appropriate. \\
75 & nan & Biopsy pathology; anesthesia/pre-op H\&P. & No post-op pathology margins. Null appropriate. \\
10 & nan & Biopsy pathology; clinical follow-up (hormonal/radiation). & No surgical margins reported. Null appropriate. \\
15 & nan & Biopsy pathology with IHC (PIN4); no surgical specimen. & Margins only reported post-prostatectomy. Null appropriate. \\
35 & nan & Biopsy pathology with immunostains. & No surgical margins. Null appropriate. \\
\bottomrule
\end{longtable}

\iterSummary{Re-extract}{Specifically target post-surgical pathology reports (e.g., ``prostatectomy specimen''); use proximity searches for terms like ``inked margin''; map explicit language (``margins negative'' = 0).}{The feature is clinically important, but 100\% missing suggested the logic wasn’t targeting the right document sections. Margin status is generally only available in the final surgical pathology.}

\subsubsection*{Iteration 2 (Final)}
\textbf{Dataset-wide Result:} All values missing (\texttt{None}).

\begin{longtable}{p{0.08\textwidth} p{0.12\textwidth} p{0.35\textwidth} p{0.35\textwidth}}
\toprule
\textbf{Index} & \textbf{Extracted Value} & \textbf{Content Summary} & \textbf{Validation Reasoning} \\
\midrule
\endhead

0 & None & Same mixed content (biopsy/op note). & Margin status absent. Null expected. \\
10 & None & Biopsy; outpatient follow-up; radiation discussion. & No surgical pathology margins. Null appropriate. \\
20 & None & Biopsy pathology; pre-op H\&P. & No surgical specimen margins. Null appropriate. \\
30 & None & Biopsy pathology; radiation treatment note. & Margins absent. Null appropriate. \\
50 & None & Biopsy pathology; radiation treatment note. & Margins absent. Null appropriate. \\
75 & None & Biopsy pathology; pre-op evaluations. & No post-op pathology. Null appropriate. \\
100 & None & Biopsy pathology; pre-op planning note. & Margin status absent. Null appropriate. \\
\bottomrule
\end{longtable}

\iterSummary{Remove feature}{None (Process stopped).}{Despite targeted re-extraction logic, all values remained null. The necessary post-prostatectomy pathology reports are structurally absent from this corpus. The feature cannot be reliably populated.}

\subsection*{Feature: \texttt{psa\_velocity}}
\textbf{Definition:} Rate of PSA change over time pre-treatment (ng/mL/year). \\
\textbf{Final Outcome:} Proceed (Calculated feature).

\subsubsection*{Iteration 1}
\textbf{Status:} High variance and implausible extremes observed.

\begin{longtable}{p{0.08\textwidth} p{0.12\textwidth} p{0.35\textwidth} p{0.35\textwidth}}
\toprule
\textbf{Index} & \textbf{Extracted Value} & \textbf{Content Summary} & \textbf{Validation Reasoning} \\
\midrule
\endhead

5 & 637.9 & PSA 9.6 $\rightarrow$ 21.9 (1 week). & Math correct but clinically unrealistic. Needs normalization/capping. \\
1 & -60.4 & PSA 17.6 $\rightarrow$ 2.5 (3 months). & Magnitude implausible for modeling. Needs capping. \\
97 & 31.2 & Short-interval changes (9.8 $\rightarrow$ 11.4). & High velocity plausible but needs consistent normalization rules. \\
27 & -1.2 & Oscillating values (7.1, 5.9, 6.3, 5.9). & Small negative slope plausible. \\
10 & 1.55 & Long-term rise (5.9 $\rightarrow$ 9.0 over 2 years). & Moderate rise consistent; acceptable. \\
8 & nan & Single PSA value (7.0). & Correctly marked missing due to insufficient data. \\
\bottomrule
\end{longtable}

\iterSummary{Re-extract}{Normalize to annual rates using exact time differences; use linear regression slope when $\ge$3 values exist; cap velocities to a clinically meaningful range (-10 to +40 ng/mL/year).}{Values were mathematically correct but clinically implausible due to short intervals; standardization and capping required.}

\subsubsection*{Iteration 2}
\textbf{Status:} Capping inconsistently applied.

\begin{longtable}{p{0.08\textwidth} p{0.12\textwidth} p{0.35\textwidth} p{0.35\textwidth}}
\toprule
\textbf{Index} & \textbf{Extracted Value} & \textbf{Content Summary} & \textbf{Validation Reasoning} \\
\midrule
\endhead

1 & -60.4 & Same as above. & Still uncapped; violation of instructions. \\
5 & 638.4 & Same as above. & Weekly change correctly annualized but exceeds cap; violation. \\
40 & 22.8 & PSA 9.6 (2008) $\rightarrow$ 32.4 (2010). & Using longer interval yields valid range. Acceptable. \\
16 & 32.3 & Series on Proscar (2.9 $\rightarrow$ 65.2). & Linear regression slope high but under cap. Acceptable. \\
8 & null & Single PSA value. & Correctly missing. \\
\bottomrule
\end{longtable}

\iterSummary{Re-extract}{Strictly enforce caps: values outside -10 to +40 must be clipped; return null if only a single PSA value exists.}{Normalization improved, but cap enforcement was inconsistent. Extraction logic needs tighter QA.}

\subsubsection*{Iteration 3}
\textbf{Status:} Improved, but treatment effects noted.

\begin{longtable}{p{0.08\textwidth} p{0.12\textwidth} p{0.35\textwidth} p{0.35\textwidth}}
\toprule
\textbf{Index} & \textbf{Extracted Value} & \textbf{Content Summary} & \textbf{Validation Reasoning} \\
\midrule
\endhead

1 & -10.0 & (Capped value). & Matches instructions; acceptable. \\
5 & 40.0 & (Capped value). & Matches instructions; acceptable. \\
41 & -88.4 & PSA 4.2 $\rightarrow$ 0.3 after therapy start. & Velocity influenced by treatment; should be excluded/null. \\
40 & 22.8 & Longer interval used. & Within range; acceptable. \\
76 & -21.84 & PSA 19.6 $\rightarrow$ 17.8 (1 month). & Should be capped at -10; inconsistent enforcement. \\
\bottomrule
\end{longtable}

\iterSummary{Re-extract}{Apply caps consistently to every computed value; exclude treatment-affected PSA changes (set to null if pre-treatment trend cannot be isolated).}{Material improvement in normalization, but inconsistent cap enforcement and treatment effects need explicit handling to avoid bias.}

\subsubsection*{Iteration 4 (Final)}
\textbf{Status:} Consistent caps and exclusions.

\begin{longtable}{p{0.08\textwidth} p{0.12\textwidth} p{0.35\textwidth} p{0.35\textwidth}}
\toprule
\textbf{Index} & \textbf{Extracted Value} & \textbf{Content Summary} & \textbf{Validation Reasoning} \\
\midrule
\endhead

1 & -10.0 & (Capped). & Correctly capped; acceptable. \\
5 & 40.0 & (Capped). & Correctly capped; acceptable. \\
41 & null & Post-treatment decline. & Appropriate exclusion due to treatment effect. \\
76 & -10.0 & (Capped). & Correctly capped; acceptable. \\
42 & 24.0 & PSA 5 $\rightarrow$ 7 (1 month). & Annualized rate correct; within range. \\
\bottomrule
\end{longtable}

\iterSummary{Proceed}{No further updates.}{Extraction stabilized with correct annualization, cap enforcement, and exclusion of treatment-driven changes.}

\subsection*{Feature: \texttt{percent\_core\_involvement\_right\_apex\_medial}}
\textbf{Definition:} Percentage of the biopsy core involved by tumor in the right apex medial region. \\
\textbf{Final Outcome:} Proceed (Calculated feature).

\subsubsection*{Iteration 1}
\textbf{Status:} Calculation gaps and lack of capping.

\begin{longtable}{p{0.08\textwidth} p{0.12\textwidth} p{0.35\textwidth} p{0.35\textwidth}}
\toprule
\textbf{Index} & \textbf{Extracted Value} & \textbf{Content Summary} & \textbf{Validation Reasoning} \\
\midrule
\endhead

1 & 90.0 & ``Involving 90\% of the core''. & Correct (direct percentage). \\
38 & 166.67 & Tumor 1.5 mm / Core 9 mm. & Incorrect calculation (unit/decimal error). Should be $(1.5/9)\times100 = 16.67\%$. \\
12 & null & Tumor 0.4 cm / Core 1.5 cm. & Error: missing. Should be computable ($(0.4/1.5)\times100 = 26.67\%$). \\
76 & null & Tumor 2 mm / Core 5 mm. & Error: missing. Should be computable ($(2/5)\times100 = 40\%$). \\
41 & 116.67 & Tumor 0.7 cm / Core 0.6 cm. & Computation correct but violates cap rule; should be $100\%$. \\
32 & 43.75 & ``Involving 3.5 mm of a single 8 mm core''. & Correct length-based calculation ($(3.5/8)\times100 = 43.75\%$). \\
3 & 0.0 & ``Benign prostatic glands and stroma''. & Correct; benign coded as $0\%$ per rules. \\
\bottomrule
\end{longtable}

\iterSummary{Re-extract}{Use gross-description core length; ensure unit consistency (mm vs cm); cap at $100\%$; treat ``$<X\%$'' appropriately; set null when no measurable data.}{The feature is extractable, but logic failed to compute from available lengths consistently, introduced unit errors, and did not cap values $>100\%$.}

\subsubsection*{Iteration 2}
\textbf{Status:} Improved coverage, but capping and some calculations still inconsistent.

\begin{longtable}{p{0.08\textwidth} p{0.12\textwidth} p{0.35\textwidth} p{0.35\textwidth}}
\toprule
\textbf{Index} & \textbf{Extracted Value} & \textbf{Content Summary} & \textbf{Validation Reasoning} \\
\midrule
\endhead

1 & 90.0 & Direct percentage. & Correct. \\
32 & 43.75 & Tumor 3.5 mm / Core 8 mm. & Correct calculation ($43.75\%$). \\
12 & 26.67 & Tumor 0.4 cm / Core 1.5 cm. & Corrected from missing; unit-consistent calculation. \\
38 & 16.67 & Tumor 1.5 mm / Core 9 mm. & Corrected prior decimal error. \\
76 & 2.0 & Tumor 2 mm / Core 5 mm. & Incorrect; tumor length treated as raw percentage. Should be $40\%$. \\
3 & 0.0 & Benign prostatic glands. & Correct; benign coded as $0\%$. \\
89 & null & ASAP, highly suspicious for carcinoma. & Correct (ASAP is not confirmed carcinoma; set null). \\
41 & 116.67 & Tumor 0.7 cm / Core 0.6 cm. & Still uncapped; violation of instructions (should be capped at $100\%$). \\
\bottomrule
\end{longtable}

\iterSummary{Re-extract}{Strictly enforce capping $>100\%$; fix logic treating length as raw percentage; verify right-side core length from gross description when only tumor length is provided in diagnosis.}{Accuracy improved, but failure to consistently cap values and occasional confusion between length and percentage values remained.}

\subsubsection*{Iteration 3 (Final)}
\textbf{Status:} Capping enforced; Benign vs. Missing logic refined.

\begin{longtable}{p{0.08\textwidth} p{0.12\textwidth} p{0.35\textwidth} p{0.35\textwidth}}
\toprule
\textbf{Index} & \textbf{Extracted Value} & \textbf{Content Summary} & \textbf{Validation Reasoning} \\
\midrule
\endhead

1 & 90.0 & Direct percentage. & Correct (matches ``involving 90\% of the core''). \\
12 & 26.67 & Tumor 0.4 cm / Core 1.5 cm. & Correct calculation with unit conversion and rounding. \\
13 & 100.0 & ``Involving 100\% of biopsy core''. & Correct (explicit percentage). \\
35 & null & ASAP diagnosis (suspicious). & Correct (not confirmed carcinoma; null). \\
44 & null & ``No glands present''. & Correct (unevaluable tissue; null rather than $0\%$). \\
76 & 40.0 & Tumor 2 mm / Core 5 mm. & Corrected calculation ($(2/5)\times100$); unit-consistent. \\
41 & 100.0 & Tumor 0.7 cm / Core 0.6 cm. & Correctly capped at $100\%$ (computed $>100\%$). \\
146 & 100.0 & Tumor 1.0 cm / Core 1.0 cm. & Correct (full-core involvement; computed and capped). \\
3 & 0.0 & Benign prostatic glands and stroma. & Correct (benign explicitly coded as $0\%$). \\
89 & null & ASAP at right apex medial. & Correct (retain null; no percent or length). \\
\bottomrule
\end{longtable}

\iterSummary{Proceed}{No further updates.}{Extraction stabilized with correct length-based calculations, cap enforcement, and appropriate handling of benign/unevaluable cases. Unit normalization (mm/cm), color-ink mapping to laterality (green = right), and use of gross-description core length were consistently applied.}

%% file: appendix/hfpef_missing_rate.tex
\begin{table}[ht]
\caption{Missing Values (count and \%) of HFpEF baseline features.}
\centering
\begin{tabular}{lrrr}
\toprule
Feature & Count (n) & Percentage (\%) \\
\midrule
ntprobnp           & 1612 & 77.35 \\
troponin           & 1240 & 59.50 \\
temperature        &  861 & 41.31 \\
oxygen\_saturation &  858 & 41.17 \\
heart\_rate        &  856 & 41.07 \\
systolic\_bp       &  856 & 41.07 \\
bmi                &  791 & 37.96 \\
inr                &  530 & 25.43 \\
platelet\_count    &   15 &  0.72 \\
hemoglobin         &   13 &  0.62 \\
wbc\_count         &   13 &  0.62 \\
bicarbonate        &    4 &  0.19 \\
creatinine         &    3 &  0.14 \\
potassium          &    3 &  0.14 \\
sodium             &    3 &  0.14 \\
\bottomrule
\end{tabular}
\end{table}